\documentclass[review]{elsarticle} 
\usepackage{subfigure}
\usepackage{bm}
\usepackage{algorithmic}
\usepackage{algorithm}
\usepackage{amsmath}
\usepackage{amsfonts}
\usepackage{amssymb}
\usepackage{multirow}
\usepackage[normalem]{ulem}
\useunder{\uline}{\ul}{}
\usepackage{bm}
\usepackage{float} 
\usepackage{textcomp}
\usepackage{subfigure}
\usepackage{geometry}
\usepackage{stmaryrd}
\usepackage{geometry}
\usepackage{amssymb}
\usepackage{amsmath}
\usepackage{amsfonts}
\usepackage{array}
\usepackage{balance}       % to better equalize the last page
\usepackage[T1]{fontenc} 
\biboptions{sort&compress}

\journal{Journal of \LaTeX\ Templates}

\begin{document}

\begin{frontmatter}

\title{Fast Hypergraph Regularized Nonnegative Tensor Ring Factorization Based on Low-Rank Approximation}

%\author{Xinhai Zhao\fnref{myfootnote}} 
%\author{Yuyuan Yu\fnref{myfootnote}}  
%\address{School of Automation, Guangdong University of Technology, Guangzhou, China}
 
% or include affiliations in footnotes:
\author[mymainaddress]{Xinhai Zhao}
\author[mymainaddress]{Yuyuan Yu}

\author[mymainaddress]{Guoxu Zhou\corref{mycorrespondingauthor}}

\cortext[mycorrespondingauthor]{Corresponding author}
\ead{gx.zhou@gdut.edu.cn}

\author[mymainaddress,mysecondaddress]{Qibin Zhao\corref{mycorrespondingauthor}}
\cortext[mycorrespondingauthor]{Corresponding author}
\ead{qibin.zhao@riken.jp}

\author[mymainaddress,mythirdaddress]{Weijun Sun}

\address[mymainaddress]{School of Automation, Guangdong University of Technology, Guangzhou, China}
\address[mysecondaddress]{The Center for Advanced Intelligence Project (AIP), RIKEN , Tokyo, Japan}
\address[mythirdaddress]{Guangdong Key Laboratory of IoT Information Technology, Guangzhou, China }
%\address[mysecondaryaddress]{Guangdong-Hong Kong-Macao Joint Laboratory for Smart Manufacturing, Guangzhou, China}

\begin{abstract}
For the high dimensional data representation, nonnegative tensor ring (NTR) decomposition equipped with manifold learning has become a promising model to exploit the multi-dimensional structure and extract the feature from tensor data.  However, the existing methods such as graph regularized tensor ring decomposition (GNTR) only models the pair-wise similarities of objects. For tensor data with complex manifold structure, the graph can not exactly construct similarity relationships. In this paper, in order to effectively utilize the higher-dimensional and complicated similarities  
among objects, we introduce hypergraph to the framework of NTR to further enhance the feature extraction, upon which a hypergraph regularized nonnegative tensor ring decomposition (HGNTR) method is developed. To reduce the computational complexity and suppress the noise, we apply the low-rank approximation trick to accelerate HGNTR (called LraHGNTR). Our experimental results show that compared with other state-of-the-art algorithms, the proposed HGNTR and LraHGNTR can achieve higher performance in clustering tasks, in addition, LraHGNTR can greatly reduce running time  without decreasing accuracy.
\end{abstract}


\end{frontmatter}
%
%\linenumbers

\section{Introduction}
\label{intro}

Nowadays, extracting meaningful features from high-dimensional data has always been an important topic in pattern recognition and machine learning.
Nonnegative matrix factorization (NMF) has been received plenty of attention \cite{lee1999learning}. 
However, with the development of data mining or acquisition and multi-sensor technology, more and more high-dimensional and large data are generated. They also emphasize that the limitation of the flat-view matrix models \cite{cichocki2015tensor}. 
Tensors are the multi-dimensional extension of matrices, which can preserve the structural information of high-order data.  
In recent years, tensor ring decomposition (TR) has been proposed and demonstrated great potentials in image completion and clustering \cite{zhao2019learning,yuan2019tensor}. 
The tensor ring decomposition essentially expresses the information of each dimension in tensors in the form of a three-order core tensor. 
Moreover, the TR complexity is increasing linearly with the dimension increasing and it can alleviate the curse of dimension. In other words,  TR is highly suitable for higher-order tensor analysis. 
To learn the localized parts of tensor objects, Yu \emph{et al.} proposed the non-negative tensor ring decomposition (NTR) \cite{yu2020graph}.  

Nonnegative tensor models have the abilities to discover local-parts based information and take full advantages of the multi-linear structure of data. Many researchers recently have found that data can be thought of as a low-dimensional non-linear manifold embedded in high-dimensional data space \cite{belkin2006manifold,belkin2001laplacian}. Because manifold learning can learn the intrinsic manifold structure of data, nonnegative tensor decomposition methods that combine with manifold learning showed better performance in many tasks and hence gained great popularity in recent years. Cai \emph{et al.} explored the geometrical information by constructing the nearest neighbours graph and proposed the graph regularized nonnegative matrix factorization (GNMF) \cite{cai2010graph}. However, the GNMF can only perform feature extraction from vector data. Yu \emph{et al.} proposed the graph regularized non-negative tensor ring decomposition (GNTR) to learn the localized parts of tensor objects and capture the manifold information\cite{yu2020graph}. But the learned graph from GNTR only models the pairs of similarity relationships between two objects and  the manifold information of data can not be sufficiently learned especially in the high-dimension.

There are two main problems in the above-mentioned methods: (1) they do not take the character of higher-order manifold similarity relations of tensor data into consideration, for example, in  graphs, we can easily obtain the similarity relations between two samples, but we can not predict if there exist more than two samples having similarity. Therefore, the operation that complex relationships among objects are simply squeezed as pair-wise forms inevitably results in manifold information loss. (2)  most of the manifold methods adopt the Gaussian kernel to measure the distance between two samples. It has been proved that the normal two-order graph may be easily affected by the radius parameter $\sigma$ of the Gaussian kernel \cite{yu2012adaptive}. These two major issues can largely be overcome by using hypergraph. The hypergraph is not limited to learning only one connection relationship of both data samples, but also the multi-level connection relationship among the multiple data samples \cite{bretto2013hypergraph}. A graph is a special case of hypergraph with each edge containing two vertices \cite{zhou2006learning}.  Zeng \emph{et al.} incorporated the manifold regularized hypergraph with the standard NMF framework. They proposed the hypergraph regularized nonnegative matrix factorization (HNMF) \cite{zeng2014image}. But it is also limited in its ability to discover the multi-linear structure of data. Yin \emph{et al.} pointed that contrasting the normal graph method, the hypergraph has the better ability to extract the local and high-order geometry structure information. So they combined the nonnegative Canonical Polyadic (CP) decomposition with hypergraph and proposed the HyperNTF \cite{yin2021hyperntf}. However, this model is difficult to be optimized. And it also has a slow convergence rate.

When large volumes of high-dimensional data become increasingly common, it is time-consuming obviously to analyze these data due to their high storage requirements and high computational complexity. Fortunately, the meaningful features usually also have relatively low-rank properties \cite{tenenbaum2000global,martinsson2011randomized}. A much smaller number of potential variables and components can contain most of the meaningful features of data. Therefore, low-rank approximation methods can be used to preprocess the obtained data.  Zhou \emph{et al.} aiming at the problems of slow convergence and excessive computational complexity, they introduced the low-rank approximation to NMF and nonnegative Tucker decomposition (NTD) \cite{zhou2012fast,zhou2015efficient}.

In this paper, 
we propose a novel non-negative tensor ring decomposition method to learn the local-parted features and capture the high-dimensional manifold structure information simultaneously. The combination can learn the manifold information completely and is significantly beneficial to improve the clustering performance. This proposed method is called hypergraph regularized non-negative tensor ring decomposition (HGNTR). 
Considering that the computation of HGNTR has a large complexity, we introduce the method of low-rank approximations into the HGNTR, then propose the LraHGNTR to reduce running time and suppress noise. The main contributions of this article include the following aspects:

\begin{itemize}
	
	\item [(1)]  HGNTR integrates the novel method of hypergraph with nonnegative tensor ring decomposition, can effectively learn the local-parted attributes of tensor data, and discover more high-dimensional manifold information to enhance the efficiency of feature extraction in the clustering tasks.
	
	\item [(2)] We develop an effective iterative algorithm based on multiplicative updating rules (MUR) to optimize the HGNTR model. Meanwhile, we apply the low-rank approximation trick to significantly reduce the computational complexity of HGNTR, which is called LraHGNTR. 
	
	\item [(3)]  Our experimental results show that both proposed algorithms can achieve higher performance metrics in clustering tasks, in addition, LraHGNTR can greatly reduce running time and even filter the noise from the original data. 
	%	We conduct comprehensive experiments to empirically analyze our algorithm.
\end{itemize}
The rest of this article is as follows: In Part 2, we firstly review the concepts and basic knowledge. In Part 3, we propose the HGNTR method and develop the MUR to update the model, and then develop the LraHGNTR. In Part 4, we take the different kinds of  experiments to valid our method. The experimental results on the database prove the effectiveness of our theory. Part 5 is followed by the conclusion.

\section{Notations and Preliminaries}
\label{sec:1}
We simply review the related notations and the definitions of nonnegative tensor decomposition as follows, and the basic notations are briefly reviewed in Table 1.

\begin{table}[htb]
	\caption{The notations}
	\centering
	\label{notations}       % Give a unique label
	\scalebox{1}{
		\begin{tabular}{ll|ll}
			\hline
			Notations & Descriptions & Notations & Descriptions  \\
			\hline 
			$x$ & A scalar & $ \textbf{x}$ & A vector \\
			$\mathbf{X}$ & A matrix & $\mathcal{X}$ & A tensor \\ 
			$\|\cdot\|_{F}$ & Frobenius norm& $\mathbf{I}$ & Identity matrix\\
			$\operatorname{tr}\{\cdot\}$ & Trace operation & $\otimes	$& Kronecker product \\
			$\times_n $ & Mode-$n$ product & $\overline{\times}^{1}	$& Multi-linear product \\
			\hline
	\end{tabular}}
\end{table}

\subsection{Definition}
\label{sec:2.1}
\textbf{Definition 1} (Mode-$n$ unfolding) Given a tensor $\mathcal{X} \in \mathbb{R}^{I_{1} \times I_{2} \times \cdots \times I_{N}}$, its standard mode-$n$ unfolding matrix can be expressed as $\mathbf{X}_{(n)} \in \mathbb{R}^{I_{n} \times I_{1} \cdots I_{n-1} I_{n+1} \cdots I_{N}}$, by fixing all the indices except $I_{n}$ \cite{kolda2009tensor}. 
%The unfolding is as follow element-wisely:
%\begin{equation}
%	\label{mode-n unfolding}
%	\left [ \mathbf{A}_{(n)}\right]_{i_{n},j} = a_{i_{n},j}^{(I_{n} \times I_{1} \cdots I_{n-1} \times I_{n+1} \cdots I_{N} )} = a_{i_{1}i_{2} \cdots i_{N}},
%\end{equation}  
%where $i_{n}=1,\cdots ,I_{n}$, and
%\begin{equation*}
%	\label{model-n unfolding details}
%	j = \sum_{p=1}^{N-2}\left( (i_{N+n-p}-1) \prod_{q=n+1}^{N+n-p-1} I_{q} \right ) + i_{n+1},\quad n=1,\cdots ,N,
%\end{equation*}
%where $I_{N+m}=I_{m}, i_{N+m} = i_{m}, ( m > 0)$, \cite{kolda2009tensor} had shown more details for the unfolding. 
If we transforms the unfolding matrix to the tensor, it is the inversion of the above operation and can be denoted as follow:
\begin{equation*}
	\label{folding}
	\mathcal{X} \leftarrow folding\left({\mathbf{X}_{(n)}}\right).
\end{equation*}
The another mode-$n$ unfolding matrix of tensor $\mathcal{X}$ is denoted by $\mathbf{X}_{[n]}\in \mathbb{R}^{I_{n} \times I_{n+1} \cdots I_{N} I_{1} \cdots I_{n-1}}$, , by fixing all the indices except $I_{n}$, which is often used in TR operations \cite{zhao2016tensor}. The distinction between two kinds of unfolding is the sequence of the $N-1$ indices.
%\textbf{Definition 2} (Inner product) Given the tensors $\mathcal{A}$ and $\mathcal{B}$ of the same size $I_{1} \times \cdots \times I_{M}$, the inner product of $\mathcal{A}$ and $\mathcal{B}$ is the sum of the products of their entries, it is a scalar and can be written as
%\begin{equation}
%	\label{inner product}
%	\left \langle \mathcal{A},\mathcal{B} \right \rangle = \sum_{i_{1}=1}^{I_{1}} \cdots \sum_{i_{M}=1}^{I_{M}} {a}_{i_{1} \cdots i_{M}} {b}_{i_{1}\cdots i_{M}}.
%\end{equation}

%\textbf{Definition 3} (Output product)
%Given tensors $\mathcal{A} \in \mathbb{R}^{I_{1} \times I_{2} \times \cdots \times I_{P}}$ and $\mathcal{B} \in \mathbb{R}^{J_{1} \times J_{2} \times \cdots \times J_{Q}}$, $I_{P}=J_{Q}$. The output product of two tensors is still a tensor, can be denoted as follow:
%\begin{equation}
%	\label{output product}
%	\left(\mathcal{A}\circ\mathcal{B}\right)_{i_{1}i_{2}\cdots i_{P}j_{1}j_{2}, \ldots j_{Q}} = a_{i_{1}i_{2}\cdots i_{P}}b_{j_{1}j_{2}\cdots j_{Q}}.
%\end{equation}

\textbf{Definition 2} (Mode-$n$ product) The mode-$n$ product of a tensor $\mathcal{X} \in \mathbb{R}^{I_{1} \times I_{2} \times \cdots \times I_{N}}$ with a matrix $\mathbf{U} \in \mathbb{R}^{J_{n} \times I_{n}}$ is denoted by $\mathcal{X} \times_n \mathbf{U} \in \mathbb {R}^{I_{1} \times \cdots \times I_{n-1} \times J_{n} \times I_{n+1} \times \cdots \times I_{N}}$. See the \cite{kolda2009tensor} for the details.
Element-wisely, we have the following formula:
\begin{equation}
	\label{mode-n product}
	\left(\mathcal{X} \times_{n} \mathbf{U}\right) _{i_{1} \cdots i_{n-1} j i_{n+1} \cdots i_{N}}= \sum_{i_{n}=1}^{I_{n}} x_{i_{1} i_{2} \ldots i_{N}} u_{j i_{n}}.
\end{equation}

\textbf{Definition 3} (Multi-linear product) Given the tensor $\mathcal{G}^{(n)}\in \mathbb{R}^{R_{n}\times I_{n} \times R_{n+1}}$ and $\mathcal{G}^{(n+1)}\in \mathbb{R}^{R_{n+1}\times I_{n+1} \times R_{n+2}}$, since the tensors $\mathcal{G}^{(n)}$ and $\mathcal{G}^{(n+1)}$ have an equivalent mode size $R_{n+1}$, they can be integrated into a single tensor by applying multi-linear product. And the single tensor can be denoted as $\mathcal{G}^{(n, n+1)}\in \mathbb{R}^{R_{n}\times I_{n}I_{n+1} \times R_{n+2}}$ \cite{chen2019nonlocal}. For multi-linear product, we have the following formula element-wisely:
\begin{equation}
	\label{Multi-liner product}
	\mathcal{G}^{\left(n,n+1\right)}_{r_{n},\left(i_{n}-1\right)I_{n+1}+i_{n+1},r_{n+2}} = \sum_{r_{n+1}=1}^{R_{n+1}}\mathcal{G}^{\left(n\right)}_{r_{n},i_{n},r_{n+1}}\mathcal{G}^{(n+1)}_{r_{n+1},i_{n+1},r_{n+2}},
\end{equation}
where $i_{n} = 1,2,\cdots ,I_{n}$ and $i_{n+1} = 1,2,\cdots ,I_{n+1}$. For simplicity, it can be rewritten as:
\begin{equation}
	\label{Multi-liner product_simple}
	\mathcal{G}^{\left(n,n+1\right)} = \mathcal{G}^{(n)}\overline{\times}^{1} \mathcal{G}^{(n+1)}.
\end{equation}
\subsection{Nonnegative Tucker Decomposition}
\label{sec:2.2} 
The nonnegative Tucker decomposition (NTD) make the tensor into the multiplication of nonnegative core tensor and factor matrices in tensor's each mode \cite{kim2007nonnegative}. It can be written as follow:
\begin{equation}
	\label{tucker}
	\begin{aligned}
		\mathcal{A}=\mathcal{G} \times_{1} \mathbf{A}^{(1)} \times_{2} \mathbf{A}^{(2)} \cdots \times_{n} \mathbf{A}^{(N)},
	\end{aligned}
\end{equation}
where $\mathcal{G} \in \mathbb{R}_{+}^{R_{1} \times R_{2} \times \cdots \times R_{N}}$ denotes the nonnegative core tensor, $\mathbf{A}^{(n)} \in \mathbb{R}_{+}^{I_{n} \times R_{n}}, n=1,2, \cdots, N$ denote the nonnegative factor matrices, $\mathbf{A}^{(n)} \geq 0$, and $\mathcal{G} \geq 0$. The $\bm{R} = \left[R_{1}, R_{2}, \cdots, R_{N}\right]$ is the nonnegative rank of NTD. 
However, the existence of the core tensor also increases the computation complexity of the model and limits the ability to represent higher-dimensional tensors.
\subsection{Graph Regularized Nonnegative Tensor Ring Decomposition}
\label{sec:2.3}
Tensor ring (TR) decomposition  is a more fundamental and general decomposition model than the Canonical Polyadic (CP) decomposition \cite{hazan2005sparse} and the Tucker decomposition (TD) \cite{zhao2016tensor}. The research results also proved the model's effectiveness in various applications, e.g, tensor-based image completion \cite{wang2017efficient,yuan2019randomized}, hyperspectral image denoising \cite{chen2019nonlocal} , neural network compression \cite{bengua2017matrix,pan2019compressing,zhao2019learning}, etc. Through the TR model, a high-order tensor $\mathcal{X} \in \mathbb{R}^{I_{1} \times I_{2} \times \cdots \times I_{N}}$ can be expressed as the circular contract products over a series of 3rd-order core tensors. Each element of the tensor $\mathcal{X}$ can be written the as follow:
\begin{equation}
	\label{tr}
	\mathcal{X}\left(i_{1}, i_{2}, \ldots, i_{N}\right)=\operatorname{Tr}\left\{\prod_{n=1}^{N} \mathbf{G}_{n}(i_{n})\right\}, 
\end{equation}
where $\mathbf{G}_{n}(i_{n}) \in \mathbb {R}^{R_{n} \times  R_{n+1}}$ is the $i_{n}$-th lateral slice matrix of the core tensor $\mathcal{G}^{(n)} \in \mathbb{R}^{R_{n} \times I_{n} \times R_{n+1}}$.  $\bm{R} = \left[R_{1}, R_{2}, \cdots, R_{N}\right]$ is called TR-rank. The TR restricts the rank of the border core tensors to be equal, such as $R_{1}=R_{N+1}$, not must be value-1, in other words, both border core tensors are 3$rd$-order tensors. Comparing with the Tensor Train decomposition (TT) \cite{oseledets2011tensor} where  the rank of edge core tensors $R_{1} = R_{N+1}$ must be restricted to value-1 and it can only keep relatively limited connection and interaction with between edge core tensors, the border core tensors of TR $\mathcal{G}^{(1)}$ and $\mathcal{G}^{(N)}$ can fully connect or interact each other directly enough. So TR can construct into a ring-like structure. Meanwhile, it is well known that the trace operations of matrices have cyclic invariance, such as $\operatorname{Tr}\left(ABC\right)=\operatorname{Tr}\left(CAB\right)$, and this property also holds for tensor ring decomposition. So the TR has the circular dimensional permutation invariance. These special attributes can make the TR model have some important numeric properties in the clustering task. 

To further sufficiently learn local-parted information,  Yu \emph{et al.} proposed the Nonnegative Tensor Ring decomposition (NTR) \cite{yu2020graph}. The NTR is based on the TR model, it can be written as follow:

\begin{equation}
	\begin{aligned}
		\label{ntr}
		\min &\frac{1}{2}\left\|\mathcal{X}-\operatorname{NTR}\left(\mathcal{G}^{(1)}, \mathcal{G}^{(2)}, \cdots, \mathcal{G}^{(N)}\right)\right\|^{2}_{F} \\
		&\text{s.t.}\ \mathcal{G}^{(n)} \geq 0,\quad n=1,2, \cdots, N, 
	\end{aligned}
\end{equation}
where $\mathcal{G}^{(n)} \in \mathbb{R}_{+}^{R_{n} \times I_{n}\times R_{n+1}}$ denotes the $n$-th nonnegative core tensor. $\operatorname{NTR}\left(\mathcal{G}^{(1)}, \mathcal{G}^{(2)}, \cdots, \mathcal{G}^{(N)}\right) $ can be thought of as the reconstruction of the core tensors. The Frobenius norm is adopted to measure the similarity of them. 
According to the theorem 3.5 in \cite{zhao2016tensor}, when we use alternating least squares (ALS) method, for the TR decomposition of $\mathcal{X}$, its matrix-form can be written as $\mathbf{X}_{[n]}=\mathbf{G}_{(2)}^{(n)}\left(\mathbf{G}_{\left[ 2 \right]}^{\neq n}\right)^{\top}$. The central idea of ALS is to optimize a core tensor when the other core tensors are fixed, which has been extensively implemented in the CP decomposition and TD. Thus, the NTR problem (\ref{ntr}) can be rewritten as the sub-problem framework:
\begin{equation}
	\begin{aligned}
		\label{ntr-subproblem}
		\min&\mathcal{F}_{NTR}^{(n)}= \frac{1}{2}\left\|\mathbf{X}_{[n]}-\mathbf{G}_{(2)}^{(n)}\left(\mathbf{G}_{\left[ 2 \right]}^{\neq n}\right)^{\top}\right\|_{F}^{2} 
		\\ 
		&\text {s.t. } \mathbf{G}_{(2)}^{(n)} \geq 0, \quad n=1,2, \cdots, N, 
	\end{aligned}
\end{equation}
where $\mathbf{G}_{(2)}^{(n)} \in \mathbb{R}_{+}^{I_{n} \times R_{n} R_{n+1}}$ denotes the classical mode-$2$ unfolding matrix of $\mathcal{G}^{(n)}$. $\bm{R} = \left[R_{1}, R_{2}, \cdots, R_{N}\right]$ denotes nonnegative rank of NTR and $\mathbf{G}^{\neq n}_{\left[ 2 \right]} \in \mathbb{R}_{+}^{I_{n+1} \cdots I_{N}I_{1} \cdots I_{n-1} \times R_{n}R_{n+1}}$ denotes the another type of mode-2 unfolding matrix of sub-chain tensor $\mathcal{G}^{\neq n} \in \mathbb{R}_{+}^{R_{n+1} \times I_{n+1} \cdots I_{N}I_{1} \cdots I_{n} \times R_{n}}$. The sub-chain tensor $\mathcal{G}^{\neq n}$ denotes the multi-linear product of all core tensors except the $n$-th core tensor and can be written as follow:
\begin{equation*}\label{sub-chain}
	\mathcal{G}^{\neq n} = \mathcal{G}^{(n+1)} \overline{\times}^{1} \cdots \overline{\times}^{1} \mathcal{G}^{(N)} \overline{\times}^{1} \mathcal{G}^{(1)} \overline{\times}^{1}  \cdots \overline{\times}^{1} \mathcal{G}^{(n-1)}.
\end{equation*}
Detailed instructions about $\mathcal{G}^{\neq n}$ can be found in the Definition 3.3 \cite{zhao2016tensor}. 
To enable NTR to observe the manifold geometrical structure of high-dimensional data,  Yu \emph{et al.} propose a graph regularized nonnegative tensor ring decomposition (GNTR) \cite{yu2020graph}. 
%GNTR not only inherits the advantages of NTR, but also additionally learns the manifold geometric information of tensor data to enhance the effectiveness in data representation in clustering and classification tasks. 
%The objective function of GNTR is as follow:
%
%\begin{equation}
%	\begin{aligned}
%		\label{GNTR}
%		\min_{\mathbf{G}_{(2)}^{(n)}} \mathcal{F}_{GNTR}^{(n)}=& \frac{1}{2}\left\|\mathbf{X}_{[n]}-\mathbf{G}_{(2)}^{(n)}\left(\mathbf{G}_{\left[ 2 \right]}^{\neq n}\right)^{\top}\right\|_{F}^{2}
%		+ \alpha \operatorname{tr}\left(\left(\mathbf{G}_{(2)}^{(N)}\right)^{\top} \mathbf{H}_{g} \mathbf{G}_{(2)}^{(N)}\right)
%		\\ 
%		&\text {s.t. } \mathbf{G}^{(n)}_{(2)} \geq 0, \mathbf{G}_{[2]}^{\neq n} \geq 0, \mathbf{H}_{g} \geq 0, n=1,2, \cdots, N, 
%	\end{aligned}
%\end{equation}
%where $\alpha \geq 0$ is the parameter of graph regularization term. $\mathbf{H}_{g}=\mathbf{D}-\mathbf{W} \in \mathbb{R}^{I_{N} \times I_{N}}$ is the graph Laplacian matrix, where $\mathbf{D}_{i i}=\sum_{j} \mathbf{W}_{i j}$. 
However, the learned graph form GNTR only models the single relationship between two objects, and therefore does not accurately indicate the high-dimensional manifold information.
%By joint minimizing the graph regularization term, we hope if two tensor objects are close in the manifold geometric structure, their low-dimensional feature after decomposition is also close to each other. 
%As the $N$-th dimension of $\mathcal{X}$ has defaulted the number of tensor objects, the manifold geometric structure information of the tensor objects can naturally integrate into the mode-2 unfolding matrix of the $N$-th core tensor through the Laplacian matrix
% $\mathbf{H}_{g}=\mathbf{D}-\mathbf{W} \in \mathbb{R}^{I_{N} \times I_{N}}$, where $\mathbf{D}_{i i}=\sum_{j} \mathbf{W}_{i j}$.

\subsection{Hypergraph}
\label{sec:2.4}
\begin{figure}[htp]  
	\centering  
	\subfigure[ ]{
		\includegraphics[width=0.33\textwidth]{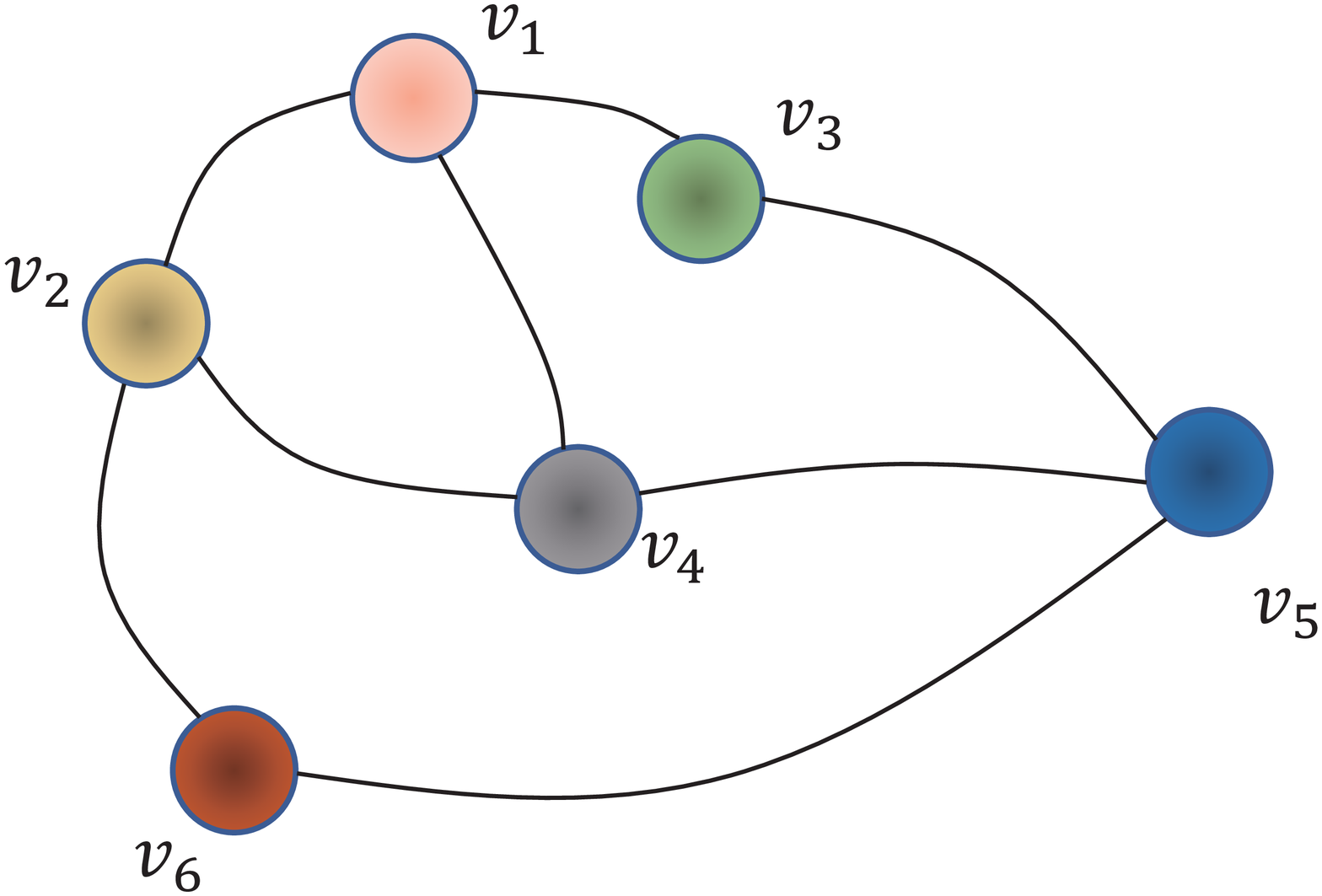}}
	\subfigure[ ]{
		\includegraphics[width=0.33\textwidth]{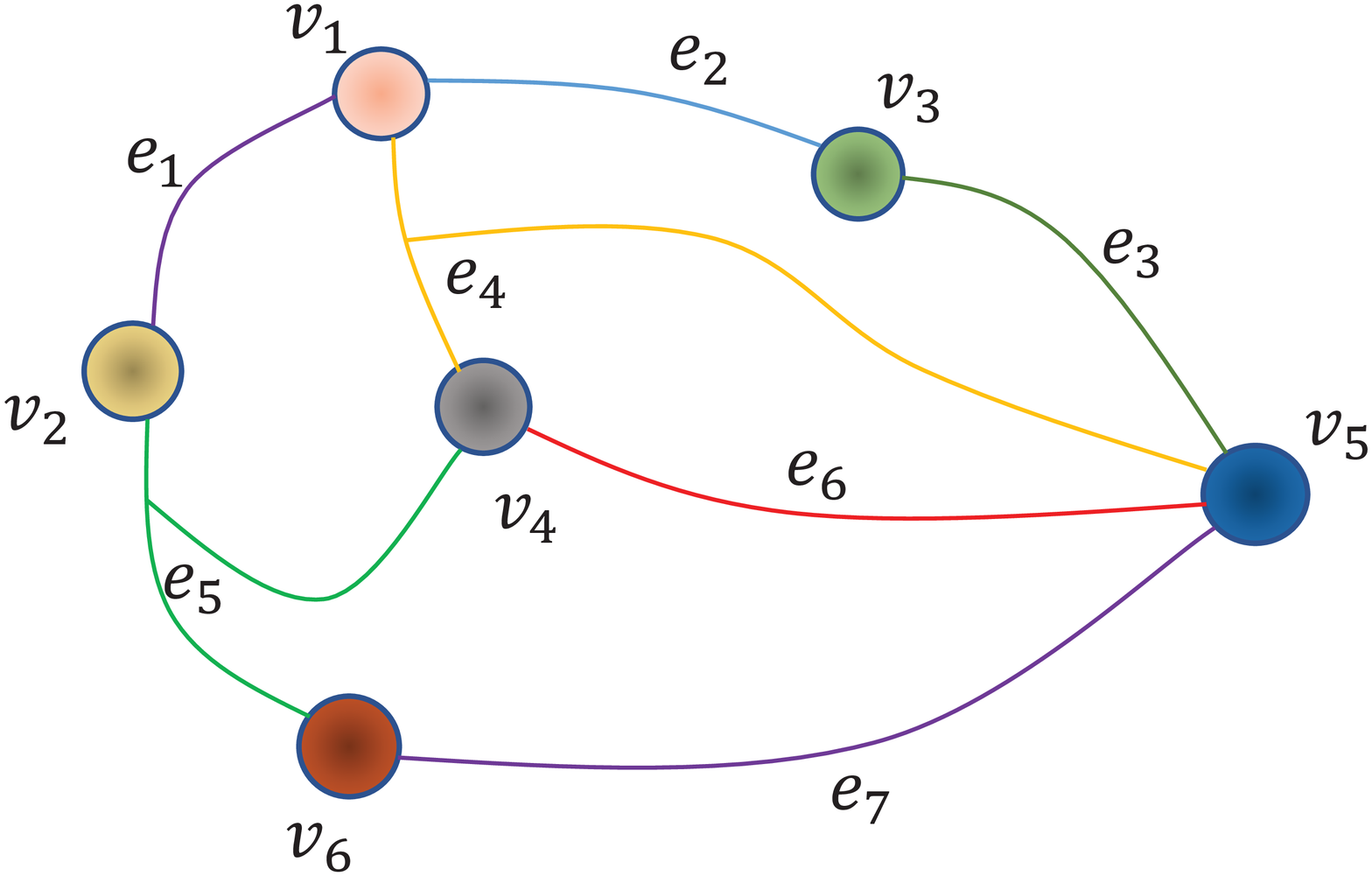}}
	\subfigure{
		\includegraphics[width=0.28\textwidth]{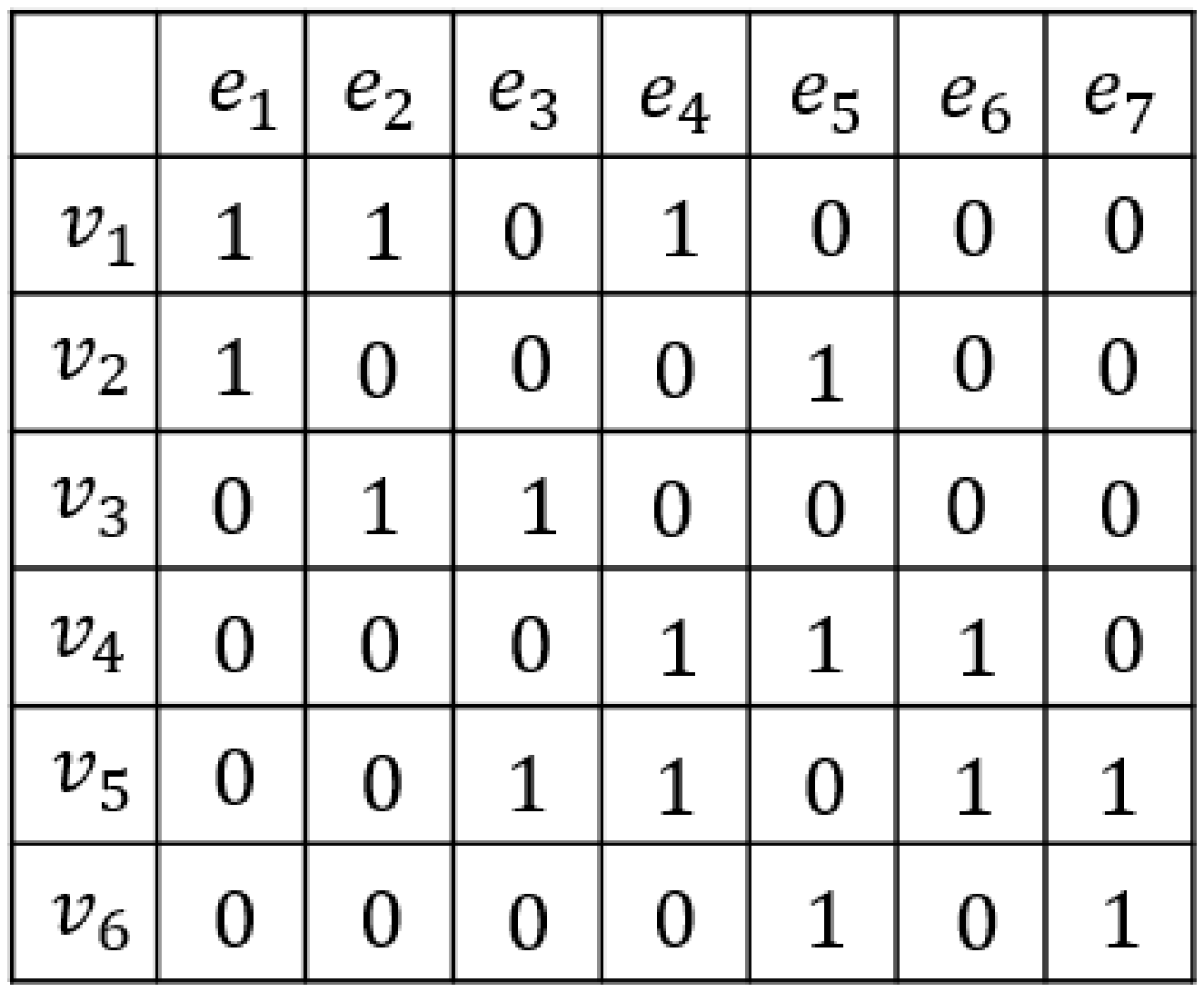}}
	\caption{The difference between the hypergraph and normal graph. (a) A normal graph. (b) A hypergraph consists of the nodes and lines that denotes their relationships. (c) 
		The incidence matrix $\mathbf{H}$ of the hypergraph.}
	\label{fig:hypergraph}
\end{figure}

The idea of hypergraph manifold learning is inspired by the theory of  simple two-dimensional graph \cite{yin2021hyperntf,zeng2014image}. In a normal graph, an edge is only connected with two vertices and the weight of the edge merely indicates the relationship between two connected vertices. However, in reality, the relationship of objects is more complex, there is a great possibility that there is a close connection between three vertices or more than. In this situation, the normal graph can not deal with this complex data well. The main distinction between hypergraph and normal two-order graph is that the hypergraph uses a subset of the vertices as an edge and each edge of a hypergraph usually connects more than two vertices. So the method of hypergraph manifold learning can promote the clustering performance \cite{hong2013multi,wang2013high,agarwal2006higher,huang2011unsupervised,tian2009hypergraph}. $Fig.\ref{fig:hypergraph}$  is an example that shows the difference between hypergraph and normal graph.

We assume that $ \mathbf{G}=\left ( V,E,\mathbf{W} \right ) $ is a weighted hypergraph. $ V $ is a finite set of vertices in the hypergraph. $ E $ is the set of hyperedges. And each hyperedge $ e $ is a subset of $ V $, then $\cup_{e\in E}=V$. $\mathbf{W}$ is a diagonal matrix that represents the weights of hyperedge in which its elements are $w$.
%\begin{equation}\label{weight W}
%	w\left ( e \right )=\frac{1}{C_{2}^{d_E\left ( e \right )}}\sum _{u,v\in e}exp\left ( -\left \| u-v \right \|^2/\sigma  \right ) 
%	\right.
%\end{equation}

The relationship of nodes and hyperedges in the hypergraph can be expressed by an indicator matrix $ \mathbf{H} $ with the size of $\left|V\right|\times\left|E\right|$. And the incidence matrix $\mathbf{H}$ of hypergraph $\mathbf{G}$ is defined as follows element-wisely \cite{yin2021hyperntf}: 
\begin{equation}\label{matrix H}
	h\left ( v,e \right )=\left\{\begin{matrix}
		1 & if\: \: \: v\in e\\ 
		0 & if\: \: \: v\notin e .\\  
	\end{matrix}\right.
\end{equation}

%\begin{figure}[htb]  %?????????????????????
%	\centering  %??
%	\includegraphics[width=0.3\linewidth]{bibtex/image/12.eps} %photo???????????????????????????photo.png????????
%	\caption{ The incidence matrix $\mathbf{H}$ of the hypergraph in the $Figure~\ref{fig:matrix H}(b)$. %\wg{Please use a better example to show hypergr (edge is a subset of vertexes, this example shows very clearly.)}
%	}
%	\label{fig:matrix H} %?????????
%\end{figure}
%$Fig~\ref{fig:matrix H}$ is the incidence matrix $\mathbf{H}$ of the hypergraph in the $Figure~\ref{fig:matrix H}(b)$.
The affinity matrix $\mathbf{A}$ of hypergraph is calculated as follow:

\begin{equation}\label{matrix A}
	\mathbf{A}_{ij} = \mathbf{exp}\left(-\frac{\left\|v_{i}-v_{j}\right\|^{2}}{\sigma^{2}}\right).
\end{equation}
where $\sigma$ is the average distance, $v_{i}$ and$v_{j}$ denotes two vertices in the hypergraph. It denotes the probabilistic hypergraph \cite{huang2010image}. It is important to note that since hypergraph can establish multiple connections among multiple vertices, they can learn a more stable and compact representation than normal graph. This special characteristic causes the hypergraph to be insensitive to hyper-parameters $\sigma$. And the weight matrix $\mathbf{W}$ of hypergraph is calculated as follow:
\begin{equation}\label{matrix W}
	\mathbf{W}_{i} = \sum_{v_{j}\in e_{i}} A \left(i,j\right).
\end{equation}

The degree of hyperedge $ e $ with a symbol of $d_E\left ( e \right )$, measures the amount of vertices or nodes $ v\in V $, is defined as $d_E\left ( e \right )=\begin{matrix}\sum _{v\in V}\end{matrix}h\left ( v,e \right )$.
The degree of each vertex  $ v\in V $ is the sum of weights for the hyperedges, is denoted as $d_V\left ( v \right )=\begin{matrix}\sum _{e\in E}\end{matrix}w\left ( e \right )h\left ( v,e \right )$. Here we denote $\mathbf{D}_{V}$ as a diagonal matrix where its elements is $d_V\left ( v \right )$. And $\mathbf{D}_{E}$ can be denoted by a diagonal matrix where its elements is $d_E\left ( e \right )$.

After obtaining a hypergraph that encodes the relationships among the samples, the data analysis tasks can be fulfilled by grouping the vertices in a way that measures the degree of similarity between samples, also known as hypergraph cutting missions from a graph learning perspective \cite{zhou2006learning}.
%According to this paper \cite{huang2018improved}, 
Then we can have the following hypergraph structure loss function:

\begin{equation}\label{hypergraph_loss_function}
	\begin{aligned}
	\Omega =\frac{1}{2}	\sum_{e\in E} \sum_{v_{i}, v_{j}\in e_{i}}\frac{w(e)h(v_{i},e)h(v_{j},e)}{d_{E}(e)}\|\frac{\mathbf{F}(v_{i})}{\sqrt{d(v_{i})}}-\frac{\mathbf{F}(v_{j})}{\sqrt{d(v_{j})}}\|_{F}^{2},
	\end{aligned}
\end{equation}
where $\mathbf{F} = \left[f_{1},\cdots,f_{m}\right]$ denotes the $n\times m $ matrix. And $f_{i}$ is a $n$-dimensional vector where each element indicates the probability that a vertex is a part of sub-graph.
The Eq.(\ref{hypergraph_loss_function}) is designed to learn a set of hypergraph cuts that retain as many hyperedges as possible when  partitioning. Notably, the sub-graphs are clusters of data after grouping. Thus, the complex data geometrical information can be preserved in the space of learned hypergraph cuts with much lower dimensionality than one of the sample spaces \cite{huang2018improved}.
After a few steps of calculation \cite{zhou2006learning}, the Eq.(\ref{hypergraph_loss_function}) can be converted into a matrix form, as follows:
\begin{equation}\label{hypergraph_loss_function_plus}
	\begin{aligned}
	\Omega = & \operatorname{tr}\left\{\mathbf{F}^{T}(\mathbf{D}_{v}-\mathbf{H}\mathbf{W}\mathbf{D}_{E}^{-1}\mathbf{H}^T)\mathbf{F}\right\} \\ 
	 = & \operatorname{tr}\left\{\mathbf{F}^{T}\mathbf{L}_{hyper}\mathbf{F}\right\},	
	\end{aligned}
\end{equation}
where $\mathbf{L}_{hyper} = \mathbf{D}_{v} -\mathbf{S}$, and $\mathbf{S} = \mathbf{H}\mathbf{W}\mathbf{D}_{E}^{-1}\mathbf{H}^T$.
%\begin{equation}\label{laplacian}
%	\begin{aligned}
%		\mathbf{L}_{hyper} = \mathbf{D}_{v} -\mathbf{S},
%	\end{aligned}
%\end{equation}
%where $\mathbf{S} = \mathbf{H}\mathbf{W}\mathbf{D}_{E}^{-1}\mathbf{H}^T $.  

\section{Hypergraph Regularized Nonnegative Tensor Ring Factorization}
\label{sec:3}%Text with citations \cite{RefB} and
In this part, in order to capture the multi-linear structural information and high-dimensional geometric manifold information of the data simultaneously, we innovatively add a hypergraph regularization term to the NTR model and propose the HGNTR model based on NTR decomposition. This model can express the original tensor data as a series of non-negative tensor circular contract products and completely capture the high-dimensional and complex manifold structural information.  And we develop an effective algorithm to optimize the HGNTR model. At the same time, considering that the meaningful part of the data usually has a certain low-rank property, to reduce the computational complexity significantly, then we propose a low-rank approximation method (LraHGNTR). 
\subsection{Objective Function of HGNTR}
\label{sec:3.1}
\begin{figure}[htb] 
	\centering  %??
	\includegraphics[width=0.55\linewidth]{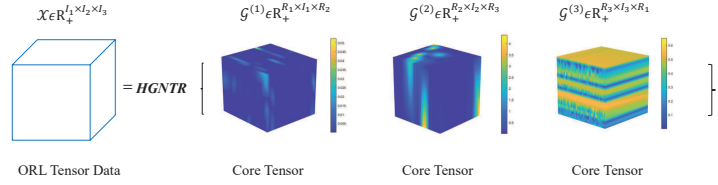}
	\caption{ Illustration picture of HGNTR on three-order database. 
		HGNTR decomposes the tensor into three core tensor, and expresses the information of each dimension as corresponding core tensor.}
	\label{fig:tensor_orl} %?????????
\end{figure}

\begin{figure}[htb]  %?????????????????????
	\centering  %??
	\includegraphics[width=0.75\linewidth]{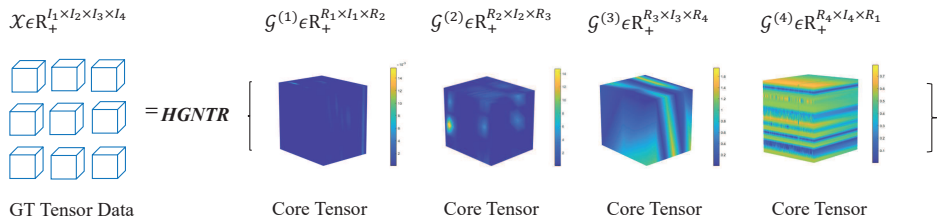} %photo???????????????????????????photo.png????????
	\caption{ Illustration picture of HGNTR on four-order database. 
		HGNTR decomposes the tensor into four core tensor, and expresses the information of each dimension as corresponding core tensor.
	}
	\label{fig:tensor_gt} %?????????
\end{figure}

Here we propose the hypergraph regularized NTR model, called HGNTR, to factorize the nonnegative tensors. And our goal is to use the hypergraph to learn the manifold information of the last dimension. It is also be said that to learn the manifold structure of the last dimensional unfolding matrix if we write the HGNTR in the sub-problem framework. The reason is that in tensor data of our experimental database, the last dimension of data usually denotes the samples and defaults the number of objects, and the hypergraph regularized term is naturally imposed on the $N$-th core tensor. Given a tensor $\mathcal{X} \in \mathbb{R}^{I_{1} \times I_{2} \times \cdots \times I_{N}}$, the objective function of HGNTR is as follows:

\begin{equation}
	\begin{aligned}
		\label{hgntr-function}
		\min_{\mathbf{G}_{(2)}^{(n)}} \mathcal{F}_{HGNTR}^{(n)}=& \frac{1}{2}\left\|\mathbf{X}_{[n]}-\mathbf{G}_{(2)}^{(n)}\left(\mathbf{G}_{\left[ 2 \right]}^{\neq n}\right)^{\top}\right\|_{F}^{2}
		+ \frac{\beta}{2} \operatorname{tr}\left(\left(\mathbf{G}_{(2)}^{(N)}\right)^{\top} \mathbf{L}_{hyper}   \mathbf{G}_{(2)}^{(N)}\right)
		\\
		&\text {s.t. }  \mathbf{G}^{(n)}_{(2)} \geq 0,\quad n=1,2, \cdots, N,
	\end{aligned}
\end{equation}
in which the $ \mathbf{L}_{hyper}\in \mathbb{R}^{I_{N}\times I_{N}} $ is the hypergraph Laplacian matrices; $ \lambda \geq 0 $ is a trade-off parameter to control the punishment of regularization term. $Fig.\ref{fig:tensor_orl}$ and $Fig.\ref{fig:tensor_gt}$ explain the principle and mechanism of HGNTR decomposition on ORL and GT database. The brief introduction of the databases is presented in Section 4.2. The last core tensor can be thought of as the feature tensor, and the other $N-1$ core tensors can be thought of as the basic tensors. By applying hypergraph regularization to the last core tensor, the higher-order similarity relations are preserved in the feature tensors that are decomposed by HGNTR.


\subsection{Optimization of HGNTR}
\label{sec:3.2}
To solve the learned objective function in Eq.(\ref{hgntr-function}), we adopt the method of Multiplicative Updating Rules (MUR). 
%Since $\mathcal{F}_{HGNTR}^{(n)}$ is not convex with respect to $\mathbf{G}_{(2)}^{(n)}$ and $\mathbf{G}_{\left[ 2 \right]}^{\neq n}$  jointly, instead, we can find a local minimum by iteratively updating $\mathbf{G}_{(2)}^{(n)}$ and $\mathbf{G}_{\left[ 2 \right]}^{\neq n}$ in a similar way [23]. 
Noticing that the equation $\left\|\mathbf{A}\right\|_{F}^{2} = \operatorname{Tr}(\mathbf{A}\mathbf{A}^{\top})$,
%\begin{equation}\label{hgntr-extend}
%	\begin{aligned}
%		\mathcal{F}_{HGNTR}^{(n)} & =\frac{1}{2} \operatorname{tr}\left(\left(\mathbf{X}_{[n]}-\mathbf{G}_{(2)}^{(n)}\left(\mathbf{G}_{\left[ 2 \right]}^{\neq n}\right)^{\top}\right)\left(\mathbf{X}_{[n]}-\mathbf{G}_{(2)}^{(n)}\left(\mathbf{G}_{\left[ 2 \right]}^{\neq n}\right)^{\top}\right)^{\top}\right) \\ 
%		&+\frac{\beta}{2}\operatorname{tr}\left(\left(\mathbf{G}_{(2)}^{(n)}\right)^{\top} \mathbf{L}_{hyper}   \mathbf{G}_{(2)}^{(n)}\right) \\ 
%		& =\frac{1}{2} \operatorname{tr}\left(\mathbf{X}_{[n]}\left(\mathbf{X}_{[n]}\right)^{\top} - 2\mathbf{G}_{(2)}^{(n)}\left(\mathbf{G}_{\left[ 2 \right]}^{\neq n}\right)^{\top}\left(\mathbf{X}_{[n]}\right)^{\top} + \mathbf{G}_{(2)}^{(n)}\left(\mathbf{G}_{\left[ 2 \right]}^{\neq n}\right)^{\top}\mathbf{G}_{\left[2\right]}^{\neq n}\left(\mathbf{G}_{\left( 2 \right)}^{\left(n\right)}\right)^{\top}\right)\\
%		&+\frac{\beta}{2}\operatorname{tr}\left(\left(\mathbf{G}_{(2)}^{(n)}\right)^{\top} \mathbf{L}_{hyper}\mathbf{G}_{(2)}^{(n)}\right). \\ 			
%	\end{aligned}
%\end{equation}
we define $\mathbf{a}_{ik}$ as the Langrange multiplier for constraints $\mathbf{G}^{(n)}_{(2)} \geq 0$.
And $\mathbf{A} =\left[\mathbf{a}_{ik}\right]$.
%and $\mathbf{B} =\left[\mathbf{b}_{jk}\right]$. 
The Langrange function $\mathcal{L}_{HGNTR}^{(n)}$ is as follow:
\begin{equation}\label{langrange function}
	\begin{aligned}
		\mathcal{L}_{HGNTR}^{(n)}& =\frac{1}{2} \operatorname{tr}\left(\mathbf{X}_{[n]}\left(\mathbf{X}_{[n]}\right)^{\top} - 2\mathbf{G}_{(2)}^{(n)}\left(\mathbf{G}_{\left[ 2 \right]}^{\neq n}\right)^{\top}\left(\mathbf{X}_{[n]}\right)^{\top} + \mathbf{G}_{(2)}^{(n)}\left(\mathbf{G}_{\left[ 2 \right]}^{\neq n}\right)^{\top}\mathbf{G}_{\left[2\right]}^{\neq n}\left(\mathbf{G}_{\left( 2 \right)}^{\left(n\right)}\right)^{\top}\right)\\	
		&+\frac{\beta}{2}\operatorname{tr}\left(\left(\mathbf{G}_{(2)}^{(n)}\right)^{\top} \mathbf{L}_{hyper}\mathbf{G}_{(2)}^{(n)}\right)+\operatorname{Tr}\left(\mathbf{A}\left(\mathbf{G}_{(2)}^{(n)}\right)^{T}\right).
	\end{aligned}
\end{equation}

In the case of $n=N$, the derivatives of $\mathcal{L}_{HGNTR}^{(n)}$ with respect to $\mathbf{G}^{(n)}_{(2)}$ 
% $ \mathbf{G}_{[2]}^{\neq n} $ 
is as follow :
\begin{equation}\label{derivatives_11}
	\begin{aligned}
		\frac{\partial\mathcal{L}_{HGNTR}^{(n)}}{\partial\mathbf{G}_{(2)}^{(n)}}= -\mathbf{X}_{[n]}\mathbf{G}_{\left[ 2 \right]}^{\neq n} + \mathbf{G}_{(2)}^{(n)}\left(\mathbf{G}_{\left[ 2 \right]}^{\neq n}\right)^{\top}\mathbf{G}_{\left[ 2 \right]}^{\neq n}+\beta\left(\mathbf{G}_{(2)}^{(n)}\right)^{\top} \mathbf{L}_{hyper}+\mathbf{A}.	
	\end{aligned}
\end{equation}

%\begin{equation}\label{derivatives_12}
%	\begin{aligned}
%		\frac{\partial\mathcal{L}_{HGNTR}^{(n)}}{\partial\left(\mathbf{G}_{\left[ 2 \right]}^{\neq n}\right)^{\top}}= -\left(\mathbf{G}_{\left( 2 \right)}^{(n) }\right)^{\top}\mathbf{X}_{[n]}+\left(\mathbf{G}_{\left( 2 \right)}^{(n) }\right)^{\top}\mathbf{G}_{(2)}^{(n)}\left(\mathbf{G}_{\left[2\right]}^{\neq n}\right)^{\top} +\mathbf{B}.
%	\end{aligned}
%\end{equation}
%At the other times such as $n=1,2, \cdots, N-1$, the derivatives formula can ignore the parts of regularization term, is rewritten as:
%\begin{equation}\label{derivatives_21}
%	\begin{aligned}
%		\frac{\partial\mathcal{L}_{HGNTR}^{(n)}}{\partial\mathbf{G}_{(2)}^{(n)}}= -\mathbf{X}_{[n]}\mathbf{G}_{\left[ 2 \right]}^{\neq n} + \mathbf{G}_{(2)}^{(n)}\left(\mathbf{G}_{\left[ 2 \right]}^{\neq n}\right)^{\top}\mathbf{G}_{\left[ 2 \right]}^{\neq n}+\mathbf{A}.	
%	\end{aligned}
%\end{equation}

%\begin{equation}\label{derivatives_22}
%	\begin{aligned}
%		\frac{\partial\mathcal{L}_{HGNTR}^{(n)}}{\partial\left(\mathbf{G}_{\left[ 2 \right]}^{\neq n}\right)^{\top}}= -\left(\mathbf{G}_{\left( 2 \right)}^{(n) }\right)^{\top}\mathbf{X}_{[n]}+\left(\mathbf{G}_{\left( 2 \right)}^{(n) }\right)^{\top}\mathbf{G}_{(2)}^{(n)}\left(\mathbf{G}_{\left[2\right]}^{\neq n}\right)^{\top}+\mathbf{B}.	
%	\end{aligned}
%\end{equation}
By 
%calculating the derivative of Eq.(\ref{langrange function}) with respect to $\mathbf{G}^{(n)}_{(2)}$ and 
using Karush-Kuhn-Tucker (KKT) condition, element-wisely $\mathbf{a}_{ik}\left(\mathbf{G}_{(2)}^{(n)}\right)_{ik} = 0$, we can get the follow update rules for each element in case of $n =N$:
%and
%$\mathbf{b}_{kj}\left(\mathbf{G}_{\left[ 2 \right]}^{\neq n}\right)^{\top}_{kj} = 0$ , 
%,we get the following equations with respect to $\left(\mathbf{G}_{(2)}^{(n)}\right)_{ik}$ :

%\begin{equation}\label{equ1}
%	\begin{aligned}
%	-\left(\mathbf{X}_{[n]}\mathbf{G}_{\left[ 2 \right]}^{\neq n}\right)_{ik}\left(\mathbf{G}_{(2)}^{(n)}\right)_{ik} + \left(\mathbf{G}_{(2)}^{(n)}\left(\mathbf{G}_{\left[ 2 \right]}^{\neq n}\right)^{\top}\mathbf{G}_{\left[2\right]}^{\neq n}\right)_{ik}\left(\mathbf{G}_{(2)}^{(n)}\right)_{ik} =0.
%	\end{aligned}
%\end{equation}
% and the equations on $\left(\mathbf{G}_{\left[ 2 \right]}^{\neq n}\right)^{\top}_{kj}$ can be derived in a similar way.
%The Eq.(\ref{equ1}) 

\begin{equation}\label{update_11}
	\begin{aligned}
		\left(\mathbf{G}_{(2)}^{(n)}\right)_{ik}\gets\left(\mathbf{G}_{(2)}^{(n)}\right)_{ik}\frac{\left(\mathbf{X}_{[n]}\mathbf{G}_{\left[ 2 \right]}^{\neq n}+\beta\left(\mathbf{G}_{\left( 2 \right)}^{( n)}\right)^{\top}\mathbf{S}\right)_{ik}}{\left(\mathbf{G}_{(2)}^{(n)}\left(\mathbf{G}_{\left[ 2 \right]}^{\neq n}\right)^{\top}\mathbf{G}_{\left[2\right]}^{\neq n}+\beta\left(\mathbf{G}_{\left( 2 \right)}^{(n)}\right)^{\top}\mathbf{D}_{v}\right)_{ik}} .
	\end{aligned}
\end{equation}

While $n=1,2, \cdots, N-1$, we just need to ignore the hypergraph regularized term. %The iterative rule of $\left(\mathbf{G}_{(2)}^{(n)}\right)_{ik}$ is the same as the above. And the rule of $\left(\mathbf{G}_{\left[ 2 \right]}^{\neq n}\right)^{\top}_{kj}$ 
It is rewritten as follow:

\begin{equation}\label{update_13}
	\begin{aligned}
		\left(\mathbf{G}_{(2)}^{(n)}\right)_{ik}\gets\left(\mathbf{G}_{(2)}^{(n)}\right)_{ik}\frac{\left(\mathbf{X}_{[n]}\mathbf{G}_{\left[ 2 \right]}^{\neq n}\right)_{ik}}{\left(\mathbf{G}_{(2)}^{(n)}\left(\mathbf{G}_{\left[ 2 \right]}^{\neq n}\right)^{\top}\mathbf{G}_{\left[2\right]}^{\neq n}\right)_{ik}} .
	\end{aligned}
\end{equation}

The whole algorithm procedure is given in Algorithm \ref{HGNTR_algorithm}. Convergence can be proved by constructing auxiliary functions. The method is the same as the convergence proof in \cite{cai2010graph}. The valid experiment is designed to show the convergence property of HGNTR and LraHGNTR in section 4.7.
\begin{algorithm}[tb]
	\small 
	\caption{HGNTR based on MU method} 
	\label{HGNTR_algorithm} 
	\begin{algorithmic}[1]
		\REQUIRE Tensor $\mathcal{X} \in \mathbb{R}^{I_{1} \times I_{2} \times \cdots \times I_{N}}$, nonnegative multi-way rank $r_{n}$ for $ n=1, 2, \cdots, N$, maximum number of iterations of $t_{max}$, balance parameter $\beta$, the number of nearest neighbours $k$. \\ 
		\ENSURE Core tensors $\mathcal{G}_{n} $ for $ n=1, 2, \cdots, N$.\\
		\STATE{Initialize $ \mathcal{G}_{n} \in \mathbb{R}^{R_{n} \times I_{n} \times R_{n+1}}$ for $n=1, 2, \cdots, N$ as random tensors from the uniform distribution between 0 and 1}.
		\STATE Calculate hypergraph Laplacian matrix $\mathbf{L}_{hyper} $ by $ \mathbf{L}_{hyper}=\mathbf{D}_V-\mathbf{S} $.
		\REPEAT 
		\FOR{$n$ = 1 to $N$}
		\STATE Obtain mode-2 unfolding matrix of sub-chain $\mathbf{G}_{[2]}^{ \neq n}$.
		\FOR{$t$ = 1 to $t_{max}$}
		\IF{$n=N$}
		\STATE Updating $\mathbf{G}_{(2)}^{(n)}$ and by Eq.(\ref{update_11}).
		\ELSE
		\STATE Updating $\mathbf{G}_{(2)}^{(n)}$ and by Eq.(\ref{update_13}) .
		\ENDIF
		\ENDFOR
		\STATE $\mathcal{G}_{n} \leftarrow$ folding$\left({\mathbf{G}_{(2)}^{(n)}}^{t_{max}}\right)$.
		\ENDFOR
		\UNTIL{convergence}
	\end{algorithmic}
\end{algorithm}
\subsection{Fast HGNTR Algorithm Based on Low-rank Approximation}
\label{sec:3.3}
In Eq.(\ref{update_11}) and Eq.(\ref{update_13}), $\mathbf{X}_{[n]}\mathbf{G}_{\left[ 2 \right]}^{\neq n}$ results in a large amount of computational complexity, especially in the condition of large-scale data. If $\mathbf{X}_{[n]}$ can be replaced by a smaller term, it is possible to reduce the running time by a significant amount. Considering the well ability of Tucker decomposition (TD) \cite{kolda2009tensor} to approximate tensor data, and the fact that it also can decompose high-dimensional tensor into a series of low-rank factors, we choose the Tucker decomposition (TD) for the low-rank approximation. 

The low-rank approximation has been successfully used in the field of matrix decomposition \cite{drineas2006fast,halko2011finding,mahoney2011randomized}.
In the field of tensor, the fast tensor decomposition methods are proposed in \cite{caiafa2010generalizing} , which are based on fiber sampling. Sequence extraction based on tSVD is also proposed in \cite{vannieuwenhoven2011truncated}, which can achieve a balance between accuracy and efficiency. 
In addition, the theory of low-rank approximation has been applied to the NMF and NTF \cite{zhou2012fast,zhou2015efficient}, and the experiment results have demonstrated the effectiveness in reducing complexity and denoising. 
These methods can be used for dimensionality reduction of tensor data.

In this paper, we approximately change the original tensor $\mathcal{X}$ into $\tilde{\mathcal{X}}$ by the TD. Then the low-rank approximate term is used in iterative processes of HGNTR. We call this fast HGNTR algorithm the low-rank HGNTR (LraHGNTR).Details about denoising can be found in the section 4.5. The LraHGNTR can be explained as follows in two steps:

The low-rank approximation (LRA) step: obtain the LRA result of $\mathcal{X}$ by normal Tucker decomposition (TD). It is denoted as follow:
\begin{equation}\label{low-rank approximation 1}
	\begin{aligned}
		\mathcal{X}\approx\mathcal{\tilde{X}} &= \left \llbracket  \mathcal{C};\mathbf{U}^{(1)},\mathbf{U}^{(2)},\cdots,\mathbf{U}^{(N)}\right \rrbracket \\
		& = \mathcal{C}\times_{1}\mathbf{U}^{(1)}\times_{2}\mathbf{U}^{(2)}\cdots\times_{N}\mathbf{U}^{(N)} .
	\end{aligned}
\end{equation}
the $\mathcal{C}_{+}\in \mathbb{R}_{+}^{\tilde{R_{1}} \times \tilde{R_{2}} \times \cdots \times \tilde{R_{N}}}$  denotes Tucker core tensor, and $\mathbf{U}^{(n)}\in \mathbb{R}_{+}^{{I_{n} \times \tilde{R_{n}}}}, n = 1, 2, \cdots , N$ denote factor matrices. Here, $\left(\mathbf{U}^{n}\right)^{\top}\mathbf{U}^{n} = \mathbf{I}$. And $\bm{\tilde{R}} = \left[\tilde{R}_{1}, \tilde{R}_{2},\cdots,\tilde{R}_{N}\right]$ is used to control the degree and error of approximation. The effect of rank $\bm{\tilde{R}}$ on  of the approximation accuracy will be described in detail in Section 4.4.

The NTR step: we need to perform NTR to minimize $\frac{1}{2}\left \|\tilde{\mathcal{X}}-\operatorname{NTR} \left(\mathcal{G}^{(1)},\mathcal{G}^{(2)},\cdots \mathcal{G}^{(N)}\right) \right \|_{F}^{2} $. Now, we show how the derivatives computing of $\mathcal{L}_{HGNTR}^{(n)}$ on $\mathbf{G}^{(n)}_{(2)}$ can be accelerated after $\mathcal{X} $ is replaced with its low-rank approximation $\tilde{\mathcal{X}}$. 
If the precondition exists, which tensor $\mathcal{X}$ has not only the structure of Tucker, but also the structure of NTR, we can have the following formula:
\begin{equation}\label{low-rank approximation 8}
	\begin{aligned}
		\tilde{\mathcal{X}} &= ~\mathcal{C}\times_{1}\mathbf{U}^{(1)}\times_{2}\mathbf{U}^{(2)}\cdots\times_{N}\mathbf{U}^{(N)}\\
		& \approx \operatorname{NTR} \left(\mathcal{G}^{(1)},\mathcal{G}^{(2)},\cdots \mathcal{G}^{(N)}\right), 
	\end{aligned}
\end{equation}
where $\mathcal{G}^{(n)} \in \mathbb{R}^{R_{n} \times I_{n}\times R_{n+1}}$. Considering $\left(\mathbf{U}^{n}\right)^{\top}\mathbf{U}^{n} = \mathbf{I}$ ,we can get the following:
\begin{equation}\label{low-rank approximation 2}
	\begin{aligned}
		\mathcal{C} \approx \operatorname{NTR} \left(\mathcal{G}^{(1)},\mathcal{G}^{(2)},\cdots \mathcal{G}^{(N)}\right) \times_{1} \left(\mathbf{U}^{(1)}\right)^{T}\times_{2} \left(\mathbf{U}^{(2)}\right)^{T} \times_{3}\cdots\times_{N}\left(\mathbf{U}^{(N)}\right)^{T}.
	\end{aligned}
\end{equation}
By storing $\mathcal{C}$, $\mathcal{G}$ and $\mathbf{U}$ to reconstruct the approximation $\tilde{\mathcal{X}}$, we can consume much less computational complexity and storage especially when $\tilde{R_{n}} \ll I_{n}$. By the Theorem 4.2 in the \cite{zhao2016tensor}, Eq.(\ref{low-rank approximation 2}) can be rewritten as follow:
\begin{equation}
	\label{low-rank approximation 3}
	\begin{aligned}
		&\mathcal{C} \times_{n}  \mathbf{U} ^{(n)} \\
		\approx &\operatorname{NTR}\left(\mathcal{G}^{(1)} \times_{2}\left(\mathbf{U}^{(1)}\right)^{T},\cdots,\mathcal{G}^{(n-1)} \times_{2}\left(\mathbf{U}^{(n-1)}\right)^{T},\mathcal{G}^{n},\mathcal{G}^{(n+1)} \times_{2}\left(\mathbf{U}^{(n+1)}\right)^{T},\cdots \mathcal{G}^{(N)} \times_{2}\left(\mathbf{U}^{(N)}\right)^{T}\right)\\
		= & \operatorname{NTR}\left(\mathcal{Z}^{(1)},\cdots,\mathcal{Z}^{(n-1)},\mathcal{G}^{(n)},\mathcal{Z}^{(n+1)},\cdots,\mathcal{Z}^{N}\right),
	\end{aligned}
\end{equation}
%Or we can write it as follow in matrix form:
%\begin{equation}\label{low-rank approximation 4}
%	\begin{aligned}
%		\left(\mathbf{U}^{(n)}\right)\mathbf{C}_{(n)} \approx \mathbf{G}_{(2)}^{(n)}\left(\mathbf{Z}_{[2]}^{\neq n}	\right)^{T},
%	\end{aligned}
%\end{equation}
where $\mathcal{Z}^{(n)} = \mathcal{G}^{(n)} \times_{2}\left(\mathbf{U}^{(n)}\right)^{T}\in \mathbb{R}^{R_{n} \times \tilde{R}_{n}\times R_{n+1}}, ~ n = 1,\cdots, n-1,n+1,\cdots, N$. The term $\mathbf{X}_{[n]}\mathbf{G}_{\left[ 2 \right]}^{\neq n}$ can be written as follow:
\begin{equation}\label{low-rank approximation 4}
	\begin{aligned}
		&~\mathbf{X}_{[n]}\mathbf{G}_{[2]}^{\neq n} \\ \approx &~\mathbf{U}^{(n)}\mathbf{C}_{(n)}\left(\mathbf{U}^{\neq n}\right)^{T}\mathbf{G}^{\neq n}_{[2]}\\ =	
		&~\mathbf{U}^{(n)}\mathbf{C}_{(n)} \mathbf{Z}^{\neq n}_{[2]}.
	\end{aligned}
\end{equation}
where $\mathbf{C}_{(n)}\in \mathbb{R}^{\tilde{R}_{n} \times \tilde{R}_{1} \cdots \tilde{R}_{n-1} \tilde{R}_{n+1} \cdots \tilde{R}_{N}} $ is the mode-n unfolding matrix of $\mathcal{C}$ 
and $\mathbf{Z}^{\neq n}_{\left[ 2 \right]} \in \mathbb{R}^{\tilde{R}_{n+1} \cdots \tilde{R}_{N}\tilde{R}_{1} \cdots \tilde{R}_{n-1} \times R_{n}R_{n+1}}$ denotes the mode-2 unfolding matrix of sub-chain tensor $\mathcal{Z}^{\neq n}$.  $\mathcal{Z}^{\neq n}$ and $\mathbf{U}^{\neq n}$ is defined as follow:
\begin{equation*}\label{low-rank approximation 10}
	\mathcal{Z}^{\neq n} = \mathcal{Z}^{(n+1)} \overline{\times}^{1} \cdots \overline{\times}^{1}\mathcal{Z}^{(N)} \overline{\times}^{1} \mathcal{Z}^{(1)} \overline{\times}^{1} \cdots \overline{\times}^{1} \mathcal{Z}^{(n-1)},
\end{equation*}
\begin{equation*}\label{low-rank approximation 12}
	\mathbf{U}^{\neq n} = \left[\mathbf{U}^{(N)}\otimes \cdots \otimes \mathbf{U}^{(n+1)} \otimes \mathbf{U}^{(n-1)} \otimes \cdots \otimes  \mathbf{U}^{(1)} \right]^{T}.
\end{equation*}
%The sub-chain tensor $\mathcal{Z}^{\neq n}$ denotes the multi-liner product of all core tensors except the $n$-th core tensor. 
%By the equation $\mathbf{X}_{[n]} = \mathbf{G}_{(2)}^{(n)}\left(\mathbf{G}_{[2]}^{\neq n}\right)^{T}$, we can get the final result:
%\begin{equation}\label{low-rank approximation 5}
%	\begin{aligned}
%	\mathbf{X}_{[n]}\mathbf{G}_{[2]}^{\neq n} \approx \left(\mathbf{U}^{(n)}\right) \mathbf{C}_{(n)}\mathbf{Z}_{[2]}^{\neq n}.	
%	\end{aligned}
%\end{equation}
%\begin{equation}\label{low-rank approximation 6}
%	\begin{aligned}
%		\left(\mathbf{G}_{\left( 2 \right)}^{(n) }\right)^{\top}\mathbf{X}_{[n]} \approx \left(
%	\mathbf{Z}_{(2)}^{(n)}\right)^{\top}\mathbf{U}^{(n)} \mathbf{C}_{(n)}.	
%	\end{aligned}
%\end{equation}

Finally, we need to change the $\mathbf{X}_{[n]}\mathbf{G}_{[2]}^{\neq n}$ 
%and $\left(\mathbf{G}_{\left( 2 \right)}^{(n) }\right)^{\top}\mathbf{X}_{[n]}$ 
by Eq.(\ref{low-rank approximation 4}).
And we can get the following update rules for each element in case of $n =N$:
\begin{equation}\label{low-rank approximation 6}
	\begin{aligned}
		\left(\mathbf{G}_{(2)}^{(n)}\right)_{ik}\gets\left(\mathbf{G}_{(2)}^{(n)}\right)_{ik}\frac{\left(\mathbf{U}^{(n)} \mathbf{C}_{(n)}\mathbf{Z}_{[2]}^{\neq n}+\beta\left(\mathbf{G}_{\left( 2 \right)}^{( n)}\right)^{\top}\mathbf{S}\right)_{ik}}{\left(\mathbf{G}_{(2)}^{(n)}\left(\mathbf{G}_{\left[ 2 \right]}^{\neq n}\right)^{\top}\mathbf{G}_{\left[2\right]}^{\neq n}+\beta\left(\mathbf{G}_{\left( 2 \right)}^{(n)}\right)^{\top}\mathbf{D}_{v}\right)_{ik}} .
	\end{aligned}
\end{equation}

While $n=1,2, \cdots, N-1$, the update rules are rewritten as follow:

\begin{equation}\label{low-rank approximation 7}
	\begin{aligned}
		\left(\mathbf{G}_{(2)}^{(n)}\right)_{ik}\gets\left(\mathbf{G}_{(2)}^{(n)}\right)_{ik}\frac{\left(\mathbf{U}^{(n)} \mathbf{C}_{(n)}\mathbf{Z}_{[2]}^{\neq n}\right)_{ik}}{\left(\mathbf{G}_{(2)}^{(n)}\left(\mathbf{G}_{\left[ 2 \right]}^{\neq n}\right)^{\top}\mathbf{G}_{\left[2\right]}^{\neq n}\right)_{ik}} .
	\end{aligned}
\end{equation}

The whole LraHGNTR algorithm is given in Algorithm \ref{LraGNTR_algorithm}.

\begin{algorithm}[tb] 
	\small
	\caption{LraHGNTR based on MU method} 
	\label{LraGNTR_algorithm} 
	\begin{algorithmic}[1]
		\REQUIRE Tensor $\mathcal{X} \in \mathbb{R}^{I_{1} \times I_{2} \times \cdots \times I_{N}}$, nonnegative multi-way rank $r_{n}$ for $ n=1, 2, \cdots, N$, multi-linear rank $\tilde{\mathbf{R}}_{n}$, maximum number of iterations of $t_{max}$, balance parameter $\beta$, the number of nearest neighbours $k$.\\ 
		\ENSURE Core tensors $\mathcal{G}_{n} $ for $ n=1, 2, \cdots, N$.\\
		\STATE{Initialize $ \mathcal{G}_{n} \in \mathbb{R}^{R_{n} \times I_{n} \times R_{n+1}}$ for $n=1, 2, \cdots, N$ as random tensors from the uniform distribution between 0 and 1}.
		\STATE Calculate hypergraph Laplacian matrix $\mathbf{L}_{hyper} $ by $ \mathbf{L}_{hyper}=\mathbf{D}_V-\mathbf{S} $.
		\REPEAT 
		\FOR{$n$ = 1 to $N$}
		\STATE Obtain the sub-chain tensor $\mathcal{Z}^{ \neq n}$.
		\FOR{$t$ = 1 to $t_{max}$}
		\IF{$n=N$}
		\STATE Updating $\mathbf{G}_{(2)}^{(n)}$ by Eq.(\ref{low-rank approximation 6}).
		\ELSE
		\STATE Updating $\mathbf{G}_{(2)}^{(n)}$ by Eq.(\ref{low-rank approximation 7}).
		\ENDIF
		\ENDFOR
		\STATE $\mathcal{G}_{n} \leftarrow$ folding$\left({\mathbf{G}_{(2)}^{(n)}}^{t_{max}}\right)$.
		\ENDFOR
		\UNTIL{convergence}.
	\end{algorithmic}
\end{algorithm}
\subsection{Computational Complexity Analysis}
\label{sec:3.4}
The HGNTR is efficiently updated by the method of Multiplicative Updating Rule. To simplify, we assume that  $I=I_{n}$ and $R=R_{n}R_{n+1}\,$, $n=1, 2, \cdots, N$. For the each iteration of the HGNTR algorithm, the tensor $\mathcal{G}^{\neq n}$ is calculated and its computational cost is $\mathcal{O}_{\mathcal{G}^{\neq n}} = (I^{2}+I^{3}+\cdots+I^{N-1})R^{\frac{3}{2}}$. 
It is important to note that we do not need to compute $\left(\mathbf{G}_{\left[ 2 \right]}^{\neq n}\right)^{\top}\mathbf{G}_{\left[2\right]}^{\neq n}$ explicitly, instead, we just use the self-contraction operation for each small core tensors $\mathcal{G}^{(n)}, n = 1, 2,\cdots, n = N-1$ in sequential order, then compute their product. The similar approaches has been used for CP and Tucker decomposition. By this operation, the computational complexity of $(\mathbf{G}_{\left[ 2 \right]}^{\neq n})^{\top}\mathbf{G}_{\left[2\right]}^{\neq n}$  is changing from $(I^{N-1})R^{2}$ to $(N-1)IR^{2}+(N-2)R^{3}$. We can conclude that as the dimension increases, the computational complexity increases linearly, making this operation more useful to represent higher-order tensor data such as four-order or more.

%The Table 2 shows the floating-point counts of HGNTR in detail when updating. 
%Then the computational cost of the gradient $\frac{\partial \mathcal{L}_{HGNTR}^{(n)}}{\partial \mathbf{G}^{(n)}_{(2)}}$ is $\mathcal{O}\left(I R^{2}+I^{N-1} R^{2}+I^{N}R\right)$.
%We suppose that the MU method converges through $t_{max}$ times when updating $\mathbf{G}^{(n)}_{(2)} $ and $ \mathbf{Y}, n=1, 2, \cdots, N$. Hence, 
The total computational complexity of HGNTR in case of $n=1,2, \cdots, N-1$ is $\mathcal{O}_{\mathcal{G}^{\neq n}}+\mathcal{O}((I^{N}R+IR+IR^{2}+\left(N-1\right)IR^{2}+\left(N-2\right)R^{3}))$.
%\begin{equation}
%	\mathcal{O}\left(\left(I^{2}+I^{3}+\cdots+I^{N-1}\right)R^{\frac{3}{2}}+2\left(I^{N}R+IR+IR^{2}+I^{N-1}R^{2}\right)\right).
%\end{equation}
In the case of $n=N$, the HGNTR only adds the cost of calculating the graph regularization term. As a result of the $\mathbf{L}_{hyper}$ is usually sparse, the computational complexity of the hypergraph term is small. Then the computational cost of HGNTR  is $\mathcal{O}_{\mathcal{G}^{\neq n}}+\mathcal{O}(I^{N}R+IR+IR^{2}+\left(N-1\right)IR^{2}+\left(N-2\right)R^{3}+2I^{2}R)$.
%\begin{equation}
%	\mathcal{O}\left(\left(I^{2}+I^{3}+\cdots+I^{N-1}\right)R^{\frac{3}{2}}+2\left(I^{N}R+2IR+IR^{2}+I^{N-1}R^{2}+I^{N-1}R(k+1)\right)\right).
%\end{equation}

%\begin{equation}
%	\mathcal{O}\left(\left(I^{2}+I^{3}+\cdots+I^{N-1}\right)R^{\frac{3}{2}}+2\left(I\left(\tilde{R}^{N-1}\left(R+\tilde{R}\right)\right)+IR+IR^{2}+I^{N-1}R^{2}\right)\right).
%\end{equation}
For the LraHGNTR, we assume that $I = I_{n}$ and $\tilde{R}=\tilde{R}_{n}\tilde{R}_{n+1}, n=1,2,\cdots,N$. The computing complexity of the mode-2 product of core tensors and factor matrices in Eq.(\ref{low-rank approximation 3}) is $\mathcal{O}_{M}= \left(N-1\right)\tilde{R}RI$.  
The total computational complexity of LraHGNTR in case of $n = 1,2,\cdots, N-1$ is $\mathcal{O}_{M}+ \mathcal{O}(I(\tilde{R}^{N-1}(R+\tilde{R}))+IR^{2}+IR+(N-1)IR^{2}+(N-2)R^{3})$. In the case of $n=N$, the total computational cost of LraHGNTR is $\mathcal{O}_{M}+\mathcal{O}(I(\tilde{R}^{N-1}(R+\tilde{R}))+IR^{2}+2I^{2}R+IR+(N-1)IR^{2}+(N-2)R^{3})$. 

The Table 2 shows the total floating-point counts of HGNTR and LraHGNTR in detail when updating.
From the above complexity analysis, we can conclude that, when the data size is large, and $R=\tilde{R}_{n}\ll I$, the computational complexity  is greatly reduced and the speed of LraHGNTR is significantly ahead. 

\renewcommand\arraystretch{1.0}
\newcolumntype{Y}{>{\centering\arraybackslash}c}
\begin{table}[ht]
	\centering
	\setlength{\tabcolsep}{8mm}
	\begin{tabular}{c|l|l}
		\hline \hline
		Algorithm         & HGNTR         & LraHGNTR                                                \\ \hline
		\multirow{3}{*}{Addition} & $NI^{N}R+NIR^{2}+$        & $NI(\tilde{R}^{N-1}(R+\tilde{R}))+2IR$ \\ 
		& $N(N-1)IR^{2}+2IR$        & $NIR^{2}+N(N-1)IR^{2}$ \\
		&$2I^{2}R+N(N-2)R^{3}$    & $+N(N-2)R^{3}+2I^{2}R$ \\
		 \hline
		\multirow{3}{*}{Multiplication}    & $NI^{N}R+NIR^{2}+NIR+$     & $NI(\tilde{R}^{N-1}(R+\tilde{R}))+NIR+$  \\ 
		& $+N(N-1)IR^{2}$     & $NIR^{2}+N(N-1)IR^{2}$  \\ 
		& $+2I^{2}R+N(N-2)R^{3}$&$+N(N-2)R^{3}+2I^{2}R$ \\
		\hline
		Division          & $NIR$                             & $NIR$                                                   \\ 
		\hline
		Overall          & $\mathcal{O}\left(NI^{N}R\right)$                             & $\mathcal{O}\left(NI(\tilde{R}^{N-1}(R+\tilde{R}))\right)$                                                   \\ 
		\hline
		%		Overall           & $\mathcal{O}\left(I^{2}R\right)$ & $\mathcal{O}\left(I^{2}R\right)$                       \\ \hline
	\end{tabular}
	\caption{Computational operation  floating-point  counts of HGNTR and LraHGNTR for each iteration when updating. }~\label{computational complexity 1}
\end{table}
%\renewcommand\arraystretch{1.5}
%\newcolumntype{Y}{>{\centering\arraybackslash}c}

%\begin{table}[ht]
%	\centering	
%	\setlength{\tabcolsep}{8mm}
%	\begin{tabular}{c|l|l}
%		\hline \hline
%		Operation & $n = 1, 2, \cdots , N-1$& $n = N$                                                \\ \hline
%		\multirow{3}{*}{Addition} & $I(\tilde{R}^{N-1}(R+\tilde{R}))+$ & $I\tilde{R}^{N-1}(R+\tilde{R})+IR^{2}+$ \\ 
%		& $IR^{2}+(N-1)IR^{2}$ & $I^{N-1}(k+1)R+2IR$ \\ 
%		&$+(N-2)R^{3}$&$+(N-1)IR^{2}+(N-2)R^{3}$ \\ \hline
%		
%		
%		\multirow{3}{*}{Multiplication} & $I(\tilde{R}^{N-1}(R+\tilde{R}))+$    & $I(\tilde{R}^{N-1}(R+\tilde{R}))+$  \\ 
%		& $IR^{2}+IR+(N-1)IR^{2}$    & $IR^{2}+I^{N-1}(k+1)R+IR$  \\ 
%		&$+(N-2)R^{3}$&$+(N-1)IR^{2}+(N-2)R^{3}$ \\ \hline
%		
%		\hline
%		Division          & $IR$                             & $IR$                                                   \\ \hline
%		%		Overall           & $\mathcal{O}\left(I^{2}R\right)$ & $\mathcal{O}\left(I^{2}R\right)$                       \\ \hline
%		
%		
%		
%	\end{tabular}
%	\caption{Computational operation  floating-point  counts for each iteration of LraHGNTR  }~\label{computational complexity 2}
%\end{table}

\section{Experiment}
\label{sec:4}

In this section, we firstly introduce performance metrics such as accuracy (ACC), normalized mutual information (NMI), and purity (PUR). Then we simply review the databases that are used for our experiment. We select six proposed algorithms as the comparison in the experiments to prove our model effectiveness.  We do clustering experiments to count the performance metrics of our algorithm on different datasets. After that, we test the accuracy of the noisy-added database. Finally, we will do the parameter sensitivity experiment and convergence experiment to test the robustness and stability of HGNTR and LraHGNTR. And we make the basic visualization to verify our excellent performance of clustering again.
\subsection{Performance Metrics}
\label{sec:4.1}
To evaluate the effectiveness of each algorithm, we adopt three metrics such as ACC, NMI, and PUR. The definition of ACC is as follows:
\begin{equation*}
	\operatorname{ACC}\left(y_{i}, \hat{y}_{i}\right)=\frac{1}{n} \sum_{i=1}^{n} \delta\left(y_{i}, \operatorname{map}\left(\hat{y}_{i}\right)\right), 
\end{equation*}
where $n$ is the total amount of objects of database. $y_{i}$ and $\hat{y}_{i}$ respectively denotes the cluster label of the object and the true label of the object. $\operatorname{map}\left( \cdot \right)$ denotes the permutation mapping function, which is responsible for mapping each cluster label $y_{i}$ to the equivalent label of the database. $\delta\left(\right)$ is the delta function. The AC is used to measure how similar the true label and the experiment-got label . For example, if the object label $y_{i}$ and the real label $\hat{y}_{i}$ are equal, then $\left(y_{i}, \operatorname{map}\left(\hat{y}_{i}\right)\right)=1$, if not, then $\left(y_{i}, \operatorname{map}\left(\hat{y}_{i}\right)\right)=0$.

By combining with the information theory, the agreement between two cluster parts can be measured with mutual information (MI). The MI is widely used in clustering applications. MI between the two sets of cluster labels $C^{\prime}$ and true labels $C$ is defined by
\begin{equation*}
	\operatorname{MI}\left(C, C^{\prime}\right)=\sum_{c_{i} \in C, c_{i}^{\prime} \in C^{\prime}} p\left(c_{i}, c_{i}^{\prime}\right) \cdot \log _{2} \frac{p\left(c_{i}, c_{i}^{\prime}\right)}{p\left(c_{i}\right) \cdot p\left(c_{i}^{\prime}\right)}, 
\end{equation*}
in which $p\left(c_{i}\right)$ and $p\left(c_{i}^{\prime}\right)$ denote the probability that the object that is selected arbitrarily from the databases belongs to  of category $c_{i}$ and category $c_{i}^{\prime}$. And $p\left(c_{i}, c_{i}^{\prime}\right)$ is defined as the joint probability that object belongs to category $c_{i}$ and category $c_{i}^{\prime}$ at the same time. To force the score to have an upper bound, we use the NMI as one of evaluation measures and the definition of NMI is denoted as follows:
\begin{equation*}
	\operatorname{NMI}\left(C, C^{\prime}\right)=\frac{\operatorname{MI}\left(C, C^{\prime}\right)}{\max \left(H(C), H\left(C^{\prime}\right)\right)}, 
\end{equation*}
where $H\left(C\right)$ and $H\left(C^{\prime}\right)$ are defined as the entropies of the true label set $C$ and the cluster label set $C^{\prime}$. It is obvious to know the score ranges of $\operatorname{NMI}\left(C, C^{\prime}\right)$ is from $0$ to $1$. If the two label sets are the same, the $\operatorname{NMI}\left(C, C^{\prime}\right)=1$, otherwise $\operatorname{NMI}\left(C, C^{\prime}\right)=0$ .

The PUR measures the degree that how many each cluster contains data points are from single category. Given a  certain category $C_{i}$ which size is $n_{i}$, the definition of PUR is as follows \cite{huang2008similarity}:
\begin{equation*}
	\operatorname{PUR}(C_{i})=\frac{1}{n_{i}} \max _{h}\left(n_{i}^{h}\right),
\end{equation*}
where $\max _{h}\left(n_{i}^{h}\right)$ is the number that represents the number of categories from category $C_{i}$ belongs to category $h$.

\subsection{Databases}
\label{sec:4.2}
In the experiment, we use two databases to prove the effectiveness of our method. A brief introduction of the two databases is as follows:

\begin{itemize}
	\item ORL Database: The ORL face database contains 400 grey-scale images of 40 different individuals. For the image in each category, these images are collected in a different light, facial expressions, and facial details. All the images are taken against a dark, uniform background, and the front face. In this experiment, each image is adjusted to the size of $32\times27$. We construct a 3rd-order tensor $\mathcal{X} \in \mathbb{R}^{32 \times 27 \times 400}$.
	
	\item GT Database: The Georgia Tech face database contains images of 50 people. All people in the database are represented by 15 color images with a cluttered background that is taken at resolution $640\times480$ pixels. The average size of the faces in these images is 150x150 pixels. The pictures show frontal or tilted faces with different facial expressions, lighting conditions, and scales. We downsampled every image to $40\times30\times3$, finally we get the 4rd-order tensor $\mathcal{X} \in \mathbb{R}^{40 \times 30 \times 3\times 750}$. 
	
	\item Face95 Database: The Face95 dataset consists of color facial images, represented by 72 individuals, each with 20 exclusive images. Hence, there are 1440 face images in the dataset. The dataset contains frontal faces with mild expressions. We downsampled the database to $50\times40\times3$, finally we get the 4rd-order tensor $\mathcal{X} \in \mathbb{R}^{50 \times 40 \times 3\times 1000}$. 
	
	\item Face96 Database: The facial images in the Face96 dataset are full color and consist of 152 individuals, each with 20 images. Therefore, there are 3040 images in the dataset. All faces in this dataset have complicated backgrounds comparing to ORL and Face95. Furthermore, the images were taken under various lighting situations and the faces contain different ranges of expressions. We downsampled every image to $49\times49\times3$, finally we get the 4rd-order tensor $\mathcal{X} \in \mathbb{R}^{49 \times 49 \times 3\times 1200}$. 
	
\end{itemize}

%Since we make a certain permutation of database dimensions, the extracted feature from the tensor database by NTR, GNTR, HGNTR, and LraHGNTR  is contained in the last core tensor.  
\subsection{Comparison Algorithms}
\label{sec:4.3}
In order to ensure the fairness of the experiment, our algorithms all use Multiplicative Updating (MU) rules  to update the model. To verify the effectiveness of our proposed algorithms, we compare our algorithms with the following state-of-the-art algorithms in our tasks.
\begin{itemize}
	\item K-means: One of the most famous unsupervised clustering algorithms.
	\item NMF: Nonnegative matrix factorization aims to learn the parts-based basis of data objects \cite{lee1999learning}.
	\item GNMF: Graph regularized nonnegative matrix factorization (GNMF), the NMF considering the manifold geometric structure in the data \cite{cai2010graph}.
	\item HNMF: Hypergraph regularized nonnegative matrix factorization (HNMF), the NMF considering the high-dimensional manifold geometric structure in the data \cite{zeng2014image}.
	\item NTR-MU: Nonnegative tensor ring decomposition based on MU optimization rule to learn the parts-based basis of tensor data objects \cite{yu2020graph}.
	\item GNTR-MU: Graph regularized nonnegative tensor ring decomposition based on MU optimization rule, the NTR considering the manifold geometric structure in the data \cite{yu2020graph}.
\end{itemize}

\subsection{Clustering Tasks for Real-world Database}
\label{sec:4.4}
In this part of the experiment, we test the comprehensive clustering performance of extracted low-dimensional features of our proposed algorithms HGNTR and LraHGNTR on databases, including their subsets. The method of getting subsets is that we randomly select the partial $n$ samples from each database and merge $n$ samples as a subset of each database. Considering the number of categories, we divide the ORL into $k_{ORL} \in \{10, 15, 20, 25, 30, 35, 40\}$ categories. We divide the GT into $k_{GT} \in \{10, 20, 30, 40, 50\}$ categories. We divide the Face95 into $k_{Face95} \in \{10, 20, 30, 40, 50\}$ categories. We divide the Face96 into $k_{Face96} \in \{20, 30, 40, 50, 60\}$ categories. Meanwhile, while running the matrix methods such as NMF, GNMF, HNMF, we need to transform the tensor into the matrix by the Matlab reshape operation. All the experiments are performed on a personal computer with an i5-9500T at 2.20GHz, 16GB memory, Windows 10, and Matlab 2020b. 
%Before the LraHGNTR experiment on each database, we need to set the appropriate low-rank approximate Tucker rank for each dataset. 
For the ORL, 
%we set $\left[8, 8, 80\right]$, 
the maximum number of iterations is set to 10. For the other databases, 
%we set $\left[10, 10, 2, 100\right]$, and 
maximum number of iterations is set to 5. 
We have counted the ACC, NMI, PUR, and their STDs obtained from the experimental results in  Table {\ref{ORL_clustering}}, Table {\ref{GT_clustering}}, Table {\ref{Face95_clustering}}, and
Table {\ref{Face96_clustering}}. Through the experimental results, we can get the following conclusions:

\begin{itemize}
	
	\item [-] For the matrix method, the effectiveness of HNMF is usually better than that of GNMF and NMF, indicating that the manifold information obtained by our hypergraph regularization can better reflect the complex manifold geometry information of the data. And it also can help improve the recognizability of low-dimensional character. Meanwhile, we have also observed that with the growth of sample categories, in other words, the amount of data, HNMF shows limited ability to obtain hypergraph information. It proves the necessity that we need to develop the tensor method of manifold learning that is more suitable for high-dimensional data.  
	
	\item [-] GNTR is usually better than NTR, shows that the learned manifold geometry can effectively enhance the identifiability of low-dimensional features. HGNTR is usually better than HNMF, indicating that our algorithm HGNTR can learn low-dimensional features better by learning spatial structure information in the form of tensors.

	\item [-] HGNTR is usually better than general other algorithms, which illustrates the model in which tensor learning algorithm and  hypergraph are combined together, can learn the complex structure information, comparing with the state-of-the-art methods. Hypergraph is an effective method, it can find more than two connection relationships between data points. It can improve the accuracy of clustering.
	
	\item [-] When the data volume (number of categories) of the three datasets is small, we find that sometimes the performance metrics of tensor methods are a little weaker than matrix methods. The reason is that tensor methods are more effective in the condition of larger amounts of data.  It also reflects the advantages of the tensor approaches.
%	We also found that our algorithm model performed better than other algorithms in most cases, and there were also some cases of poor performance. The possible reason is that we use the random selection to divide the databases.
	
\end{itemize}

\begin{figure}[!tb] 
	\centering  
	\subfigure[ ]{
		\includegraphics[width=0.48\textwidth]{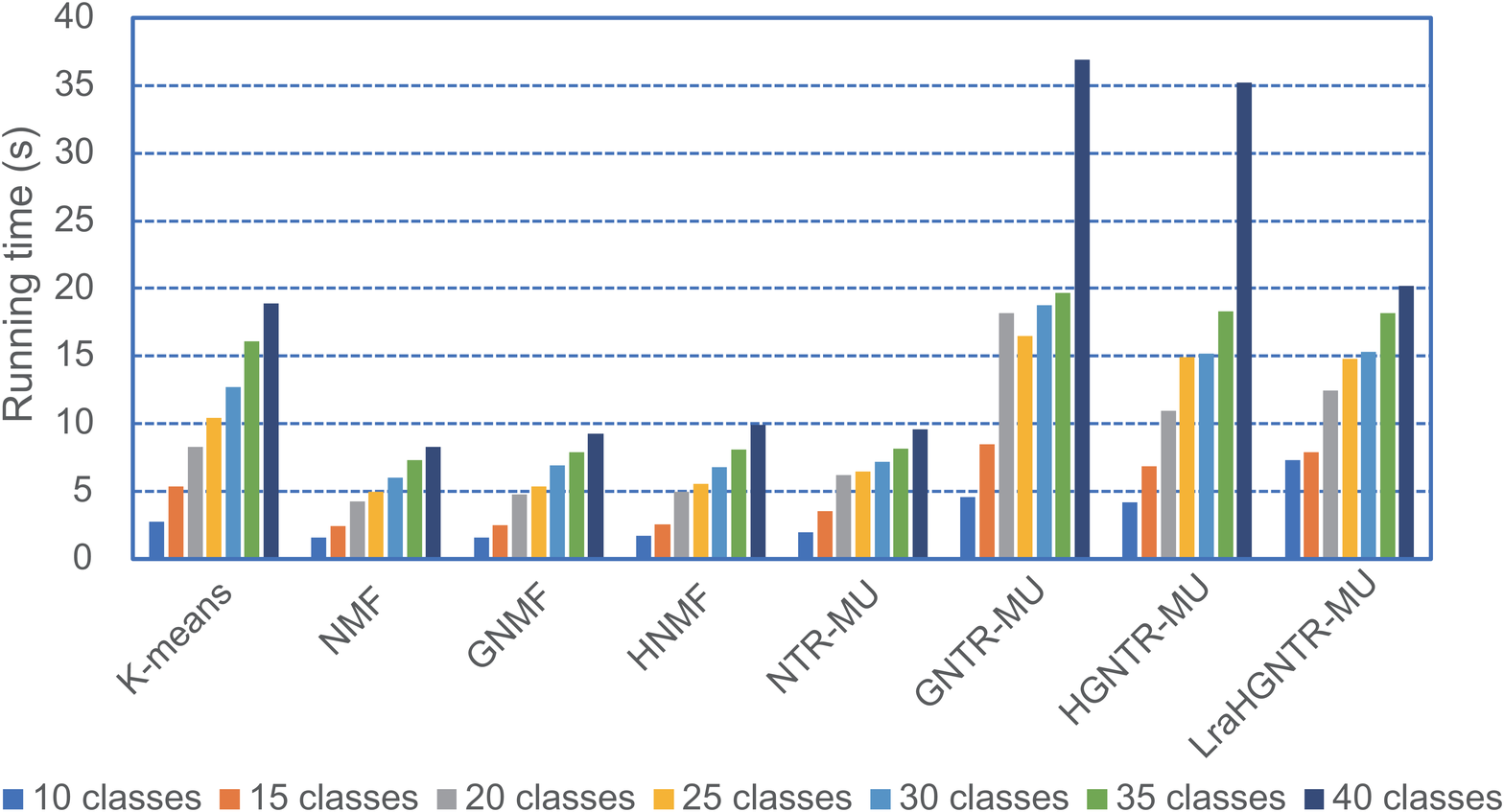}}
	\subfigure[ ]{
		\includegraphics[width=0.49\textwidth]{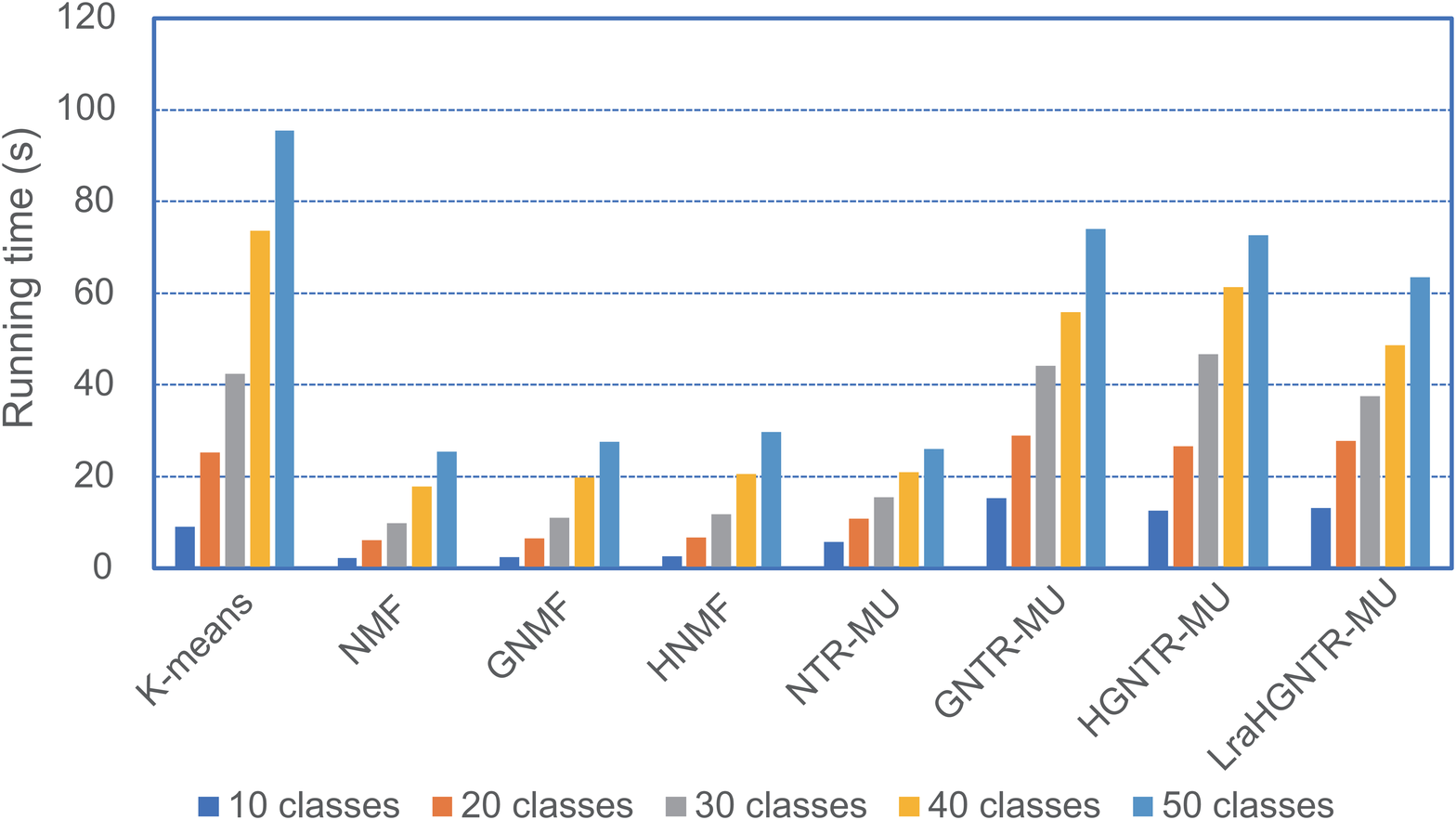}}
	\caption{The running time of various algorithms subset in different number of categories. (a) ORL. (b) GT.
	}
	\label{fig:runtime1} 
\end{figure}

We also count the running time of each database under different categories in $Fig.\ref{fig:runtime1}$. Through the experimental results, we can get the following conclusions: 
\begin{itemize}

	\item [-] Noting that the question of running time of HGNTR and the computational complexity, we consider using a low-rank Tucker approximation method to replace the greatly complex term in the iterative process. Experiments have shown that the performance of HGNTR and LraHGNTR are similar, but LraHGNTR can greatly reduce the running time. It verifies that the low-rank Tucker approximation can effectively reduce the computational cost while retaining the accuracy .This is a key point of our algorithm leading other algorithms. 
	
	\item [-] Through the comparison of subsets, we also found that as the category of subsets increases, in other words, as the number of data increases, compared with other algorithms, our algorithm HGNTR and LraHGNTR outperforms more obviously. Especially for LraHGNTR, as the amount of data increases, the performance of our algorithm LraHGNTR drops significantly in the running time.
	%	\item [-] By comparison, we find that our algorithm LraHGNTR can reduce the running time significantly while retaining the accuracy. This is a key point of our algorithm leading other algorithms. 
\end{itemize}	

By the experiment, we can conclude that the effect of rank $\bm{\tilde{R}}$ for LraHGNTR, if we choose a lower rank, our running time will be greatly reduced, but the accuracy will be lost. If we choose a little larger rank, our accuracy will increase. So if we choose a suitable rank, we can retain the accuracy, and get the ideal running time. 
% \usepackage{multirow}

\renewcommand\arraystretch{1.0}

%\newcolumntype{Y}{>{\centering\arraybackslash}c}
\begin{table}[!htb]
	 \setlength{\tabcolsep}{1.0mm}
	\begin{tabular}{lccccccccc}
		\hline
		& \multicolumn{1}{l}{$k_{ORL}$} & K-means         & NMF            & GNMF           & HNMF                 & NTR-MU         & GNTR-MU        & HGNTR-MU                & lraHGNTR-MU             \\ \hline
		\multicolumn{1}{r}{\multirow{7}{*}{ACC}} & 10                   & 75.80$\pm$3.35 & 77.70$\pm$3.26 & 74.60$\pm$2.87 & 75.30$\pm$3.40       & 71.70$\pm$2.75 & 78.00$\pm$2.70 & \textbf{78.30$\pm$3.49} & {\ul 78.10$\pm$3.14}    \\
		\multicolumn{1}{r}{}                     & 15                   & 69.93$\pm$2.97 & 75.06$\pm$5.57 & 73.80$\pm$4.48 & 76.33$\pm$4.01       & 68.53$\pm$4.80 & 75.40$\pm$2.40 & \textbf{77.06$\pm$3.37} & {\ul 76.66$\pm$2.96}    \\
		\multicolumn{1}{r}{}                     & 20                   & 69.60$\pm$3.31 & 72.55$\pm$2.99 & 75.15$\pm$2.69 & 78.85$\pm$3.85       & 70.00$\pm$2.69 & 78.00$\pm$4.18 & {\ul 80.05$\pm$2.87}    & \textbf{80.10$\pm$2.99} \\
		\multicolumn{1}{r}{}                     & 25                   & 68.64$\pm$4.24 & 72.80$\pm$4.58 & 76.12$\pm$2.34 & 75.90$\pm$3.05       & 68.60$\pm$3.08 & 77.64$\pm$2.66 & {\ul 80.68$\pm$3.69}    & \textbf{81.20$\pm$2.95} \\
		\multicolumn{1}{r}{}                     & 30                   & 73.36$\pm$2.66 & 69.36$\pm$2.66 & 73.80$\pm$2.41 & 72.96$\pm$3.39       & 66.16$\pm$4.91 & 73.60$\pm$2.66 & {\ul 74.00$\pm$2.51}    & \textbf{76.06$\pm$2.21} \\
		\multicolumn{1}{r}{}                     & 35                   & 66.54$\pm$4.69 & 69.08$\pm$2.43 & 72.85$\pm$2.30 & 72.32$\pm$2.15       & 63.71$\pm$2.97 & 73.65$\pm$2.54 & \textbf{74.08$\pm$1.60} & {\ul 73.91$\pm$2.87}    \\
		\multicolumn{1}{r}{}                     & 40                   & 66.95$\pm$3.54 & 67.93$\pm$2.01 & 73.28$\pm$2.77 & 71.47$\pm$2.73       & 68.55$\pm$3.43 & 74.70$\pm$2.85 & {\ul 75.65$\pm$2.09}    & \textbf{76.55$\pm$2.10} \\ \hline
		\multirow{7}{*}{NMI}                     & 10                   & 85.54$\pm$1.06 & 84.19$\pm$2.24 & 84.71$\pm$1.32 & 84.78$\pm$0.82       & 80.74$\pm$3.43 & 84.07$\pm$1.31 & {\ul 85.40$\pm$1.06}    & \textbf{85.22$\pm$0.78} \\
		& 15                   & 84.13$\pm$1.47 & 85.30$\pm$3.55 & 85.59$\pm$1.99 & {\ul 88.11$\pm$1.65} & 80.79$\pm$1.76 & 86.22$\pm$1.30 & 87.94$\pm$1.52          & \textbf{87.93$\pm$1.15} \\
		& 20                   & 82.99$\pm$2.08 & 84.03$\pm$2.21 & 85.60$\pm$1.86 & 88.12$\pm$2.46       & 83.05$\pm$2.04 & 87.36$\pm$1.62 & {\ul 89.15$\pm$0.95}    & \textbf{89.18$\pm$1.72} \\
		& 25                   & 84.62$\pm$1.71 & 85.40$\pm$2.75 & 87.38$\pm$1.67 & 87.81$\pm$1.05       & 82.79$\pm$1.78 & 87.35$\pm$0.80 & {\ul 89.98$\pm$1.20}    & \textbf{90.06$\pm$0.96} \\
		& 30                   & 84.13$\pm$1.77 & 84.46$\pm$1.55 & 87.20$\pm$0.76 & 87.02$\pm$1.35       & 82.78$\pm$2.37 & 87.36$\pm$0.99 & {\ul 88.20$\pm$1.25}    & \textbf{88.76$\pm$0.84} \\
		& 35                   & 83.87$\pm$2.17 & 85.13$\pm$1.18 & 86.55$\pm$1.42 & 87.13$\pm$1.21       & 80.89$\pm$1.61 & 87.02$\pm$1.15 & {\ul 88.06$\pm$1.01}    & \textbf{88.11$\pm$0.91} \\
		& 40                   & 84.59$\pm$1.43 & 84.08$\pm$0.68 & 87.78$\pm$1.07 & 87.44$\pm$0.98       & 84.78$\pm$1.65 & 88.51$\pm$0.77 & {\ul 89.81$\pm$1.07}    & \textbf{89.91$\pm$0.73} \\ \hline
		\multirow{7}{*}{PUR}                     & 10                   & 80.10$\pm$2.60 & 80.70$\pm$2.45 & 78.90$\pm$1.59 & 79.90$\pm$2.96       & 75.30$\pm$3.26 & 81.30$\pm$1.41 & {\ul 81.60$\pm$2.01}    & \textbf{81.70$\pm$2.05} \\
		& 15                   & 76.06$\pm$2.34 & 79.33$\pm$4.71 & 79.33$\pm$3.23 & 81.40$\pm$2.72       & 72.20$\pm$2.86 & 80.53$\pm$1.62 & {\ul 81.66$\pm$2.49}    & \textbf{81.73$\pm$1.87} \\
		& 20                   & 73.30$\pm$3.09 & 76.10$\pm$3.06 & 78.25$\pm$2.72 & 81.55$\pm$3.70       & 73.85$\pm$2.82 & 80.75$\pm$3.29 & {\ul 82.65$\pm$1.90}    & \textbf{82.50$\pm$2.83} \\
		& 25                   & 73.64$\pm$3.36 & 76.40$\pm$3.77 & 79.20$\pm$1.98 & 79.04$\pm$1.96       & 72.12$\pm$2.83 & 80.12$\pm$2.00 & {\ul 82.88$\pm$2.69}    & \textbf{83.28$\pm$1.70} \\
		& 30                   & 71.60$\pm$3.30 & 73.36$\pm$2.66 & 76.96$\pm$2.07 & 76.43$\pm$2.72       & 74.23$\pm$3.45 & 77.20$\pm$1.91 & {\ul 77.46$\pm$2.05}    & \textbf{78.96$\pm$1.95} \\
		& 35                   & 70.97$\pm$3.46 & 72.94$\pm$2.28 & 75.83$\pm$2.32 & 76.17$\pm$1.64       & 67.45$\pm$2.92 & 76.54$\pm$1.96 & {\ul 76.74$\pm$2.12}    & \textbf{77.45$\pm$2.07} \\
		& 40                   & 71.33$\pm$2.95 & 71.53$\pm$0.76 & 76.83$\pm$2.03 & 75.47$\pm$1.78       & 72.95$\pm$2.87 & 78.30$\pm$1.92 & {\ul 79.45$\pm$1.52}    & \textbf{79.65$\pm$2.00} \\ \hline
	\end{tabular}
\caption{Performance metrics comparisons on ORL database between K-means, NMF, GNMF, HNMF, NTR-MU, GNTR-MU, HGNTR-MU and LraHGNTR-MU methods. We divide the database into different subset. Each subset contains $k_{ORL}$ categories. }~\label{ORL_clustering}
\end{table}

\renewcommand\arraystretch{1.0}
\begin{table}[!htb]
	\setlength{\tabcolsep}{1.0mm}
	\centering
	\begin{tabular}{lccccccccc}
		\hline
		& $k_{GT}$     & K-means         & NMF            & GNMF           & HNMF           & NTR-MU         & GNTR-MU              & HGNTR-MU                & lraHGNTR-MU             \\ \hline
		\multirow{5}{*}{ACC} & 10.00 & 62.66$\pm$2.49 & 66.40$\pm$3.07 & 62.13$\pm$3.31 & 67.33$\pm$2.44 & 62.66$\pm$3.36 & {\ul 74.80$\pm$4.09} & 74.00$\pm$2.49          & \textbf{75.33$\pm$2.90} \\
		& 20.00 & 53.20$\pm$4.97 & 54.80$\pm$1.82 & 46.53$\pm$4.18 & 54.33$\pm$0.62 & 57.33$\pm$4.23 & 61.26$\pm$2.03       & {\ul 61.53$\pm$0.93}    & \textbf{62.66$\pm$3.30} \\
		& 30.00 & 46.48$\pm$2.67 & 48.27$\pm$1.70 & 43.91$\pm$2.24 & 49.06$\pm$0.92 & 48.53$\pm$6.14 & 53.78$\pm$2.29       & {\ul 58.22$\pm$2.48}    & \textbf{59.02$\pm$1.81} \\
		& 40.00 & 46.47$\pm$0.85 & 47.26$\pm$2.57 & 45.76$\pm$0.38 & 48.50$\pm$1.56 & 53.90$\pm$8.34 & 52.70$\pm$1.03       & {\ul 55.37$\pm$2.49}    & \textbf{55.80$\pm$1.55} \\
		& 50.00 & 45.71$\pm$0.76 & 45.72$\pm$1.16 & 43.33$\pm$1.80 & 49.49$\pm$1.20 & 43.47$\pm$6.21 & 50.80$\pm$2.86       & {\ul 52.00$\pm$2.37}    & \textbf{53.25$\pm$2.33} \\ \hline
		\multirow{5}{*}{NMI} & 10.00 & 72.76$\pm$2.74 & 72.97$\pm$2.30 & 68.06$\pm$2.74 & 69.00$\pm$3.09 & 66.45$\pm$1.77 & 74.25$\pm$3.79       & {\ul 74.69$\pm$2.70}    & \textbf{75.55$\pm$1.64} \\
		& 20.00 & 66.99$\pm$3.01 & 65.97$\pm$1.67 & 58.30$\pm$3.96 & 66.81$\pm$1.17 & 67.42$\pm$3.68 & 71.89$\pm$1.97       & {\ul 73.01$\pm$0.87}    & \textbf{73.04$\pm$2.10} \\
		& 30.00 & 62.21$\pm$0.92 & 63.50$\pm$2.17 & 59.33$\pm$1.63 & 63.57$\pm$0.80 & 64.45$\pm$5.05 & 67.70$\pm$0.62       & {\ul 70.85$\pm$1.32}    & \textbf{71.51$\pm$0.91} \\
		& 40.00 & 63.26$\pm$0.83 & 63.54$\pm$1.39 & 62.98$\pm$0.84 & 65.06$\pm$0.93 & 68.69$\pm$5.55 & 68.71$\pm$0.41       & \textbf{71.24$\pm$0.76} & {\ul 70.82$\pm$0.65}    \\
		& 50.00 & 64.37$\pm$0.58 & 63.64$\pm$0.48 & 62.76$\pm$1.58 & 66.12$\pm$0.86 & 63.30$\pm$5.52 & 69.25$\pm$2.31       & {\ul 70.64$\pm$2.23}    & \textbf{71.42$\pm$2.30} \\ \hline
		\multirow{5}{*}{PUR} & 10.00 & 66.26$\pm$3.60 & 69.60$\pm$2.64 & 64.93$\pm$2.38 & 67.46$\pm$2.68 & 64.66$\pm$3.62 & {\ul 74.80$\pm$4.09} & 74.53$\pm$2.23          & \textbf{76.13$\pm$1.52} \\
		& 20.00 & 56.87$\pm$3.73 & 57.47$\pm$1.72 & 49.33$\pm$3.74 & 57.73$\pm$1.48 & 60.67$\pm$3.66 & 64.60$\pm$2.01       & {\ul 64.86$\pm$0.98}    & \textbf{65.06$\pm$3.18} \\
		& 30.00 & 49.38$\pm$2.07 & 50.89$\pm$2.07 & 46.67$\pm$1.84 & 52.35$\pm$1.06 & 52.08$\pm$6.55 & 57.37$\pm$2.30       & {\ul 61.11$\pm$1.72}    & \textbf{61.86$\pm$1.09} \\
		& 40.00 & 49.20$\pm$0.43 & 50.00$\pm$2.57 & 48.03$\pm$0.55 & 51.33$\pm$1.75 & 56.73$\pm$7.16 & 55.93$\pm$0.75       & {\ul 58.36$\pm$2.07}    & \textbf{59.10$\pm$1.13} \\
		& 50.00 & 48.24$\pm$0.92 & 47.78$\pm$1.26 & 46.48$\pm$2.02 & 51.65$\pm$1.23 & 46.43$\pm$6.79 & 54.49$\pm$2.98       & {\ul 55.49$\pm$2.38} &\textbf { 56.16$\pm$2.98}    \\ \hline
	\end{tabular}
\caption{Performance metrics comparisons on GT database between K-means, NMF, GNMF, HNMF, NTR-MU, GNTR-MU, HGNTR-MU and LraHGNTR-MU methods. We divide the database into different subset. Each subset contains $k_{GT}$ categories. }~\label{GT_clustering}
\end{table}

\renewcommand\arraystretch{1.0}
\begin{table}[!htb]
	\setlength{\tabcolsep}{0.8mm}
	\begin{tabular}{lccccccccc}
		\hline
		& $k_{Face95}$   & K-means     & NMF        & GNMF             & HNMF                & NTR-MU     & GNTR-MU             & HGNTR-MU            & lraHGNTR-MU         \\ \hline
		\multirow{5}{*}{ACC} & 10 & 76.50$\pm$6.76 & 79.20$\pm$4.20 & 84.90$\pm$1.51       & 82.40$\pm$5.09          & 73.40$\pm$4.76 & 85.20$\pm$2.41          & {\ul 85.60$\pm$2.10}    & \textbf{86.10$\pm$2.56} \\
		& 20 & 59.05$\pm$1.67 & 58.10$\pm$2.82 & 65.50$\pm$1.74       & 60.40$\pm$3.92          & 56.45$\pm$1.68 & 64.05$\pm$4.66          & {\ul 64.80$\pm$4.42}    & \textbf{66.00$\pm$0.53} \\
		& 30 & 50.96$\pm$3.47 & 49.53$\pm$1.70 & 54.73$\pm$1.69       & 55.60$\pm$1.42          & 51.66$\pm$1.53 & 55.06$\pm$1.78          & \textbf{56.50$\pm$1.87} & {\ul 56.30$\pm$2.62}    \\
		& 40 & 47.90$\pm$3.10 & 45.70$\pm$1.90 & 50.00$\pm$1.40       & 49.25$\pm$2.06          & 46.02$\pm$1.61 & 50.17$\pm$1.80          & {\ul 50.58$\pm$1.96}    & \textbf{51.43$\pm$1.83} \\
		& 50 & 44.82$\pm$1.99 & 44.18$\pm$1.38 & 49.24$\pm$1.20       & 46.20$\pm$2.65          & 45.74$\pm$1.54 & 49.68$\pm$1.57          & \textbf{51.02$\pm$2.11} & {\ul 50.48$\pm$1.89}    \\ \hline
		\multirow{5}{*}{NMI} & 10 & 79.67$\pm$2.88 & 78.58$\pm$3.59 & 83.55$\pm$1.62       & 84.07$\pm$1.74          & 74.78$\pm$1.69 & 83.48$\pm$1.89          & {\ul 84.20$\pm$2.37}    & \textbf{85.04$\pm$2.45} \\
		& 20 & 70.64$\pm$1.34 & 69.52$\pm$0.94 & 73.71$\pm$1.00       & 71.66$\pm$1.78          & 67.95$\pm$0.93 & \textbf{74.85$\pm$2.77} & 74.00$\pm$1.88          & {\ul 74.39$\pm$0.53}    \\
		& 30 & 68.53$\pm$2.26 & 67.18$\pm$0.79 & 70.48$\pm$0.78       & 71.12$\pm$0.59          & 68.15$\pm$0.82 & 70.25$\pm$0.98          & {\ul 71.35$\pm$0.60}    & \textbf{71.49$\pm$1.85} \\
		& 40 & 67.39$\pm$1.46 & 65.60$\pm$0.98 & 68.57$\pm$0.51       & 68.65$\pm$0.67          & 66.64$\pm$1.14 & 68.56$\pm$1.10          & {\ul 69.39$\pm$0.82}    & \textbf{69.45$\pm$0.89} \\
		& 50 & 67.53$\pm$0.76 & 66.20$\pm$0.57 & 69.06$\pm$0.32       & 67.88$\pm$1.37          & 67.24$\pm$1.31 & 69.83$\pm$0.80          & {\ul 70.26$\pm$0.95}    & \textbf{70.65$\pm$0.66} \\ \hline
		\multirow{5}{*}{PUR} & 10 & 78.60$\pm$4.90 & 79.40$\pm$3.87 & 84.90$\pm$1.51       & 84.10$\pm$4.15          & 74.70$\pm$3.03 & 85.20$\pm$2.41          & {\ul 85.70$\pm$2.07}    & \textbf{86.20$\pm$2.51} \\
		& 20 & 61.90$\pm$1.82 & 60.80$\pm$1.67 & {\ul 67.15$\pm$1.65} & 63.10$\pm$2.58          & 59.35$\pm$1.64 & 67.10$\pm$4.10          & 66.80$\pm$2.95          & \textbf{67.75$\pm$0.50} \\
		& 30 & 54.13$\pm$3.00 & 52.53$\pm$1.32 & 57.70$\pm$1.50       & \textbf{59.23$\pm$1.53} & 54.53$\pm$1.74 & 57.53$\pm$1.27          & {\ul 59.17$\pm$1.10}    & 58.93$\pm$2.03          \\
		& 40 & 50.92$\pm$2.41 & 49.00$\pm$2.07 & 53.27$\pm$1.44       & 52.47$\pm$1.54          & 49.70$\pm$1.23 & 53.02$\pm$1.30          & {\ul 53.75$\pm$2.16}    & \textbf{54.78$\pm$1.60} \\
		& 50 & 48.50$\pm$1.57 & 47.60$\pm$0.83 & 52.32$\pm$0.61       & 50.22$\pm$2.23          & 48.86$\pm$1.68 & 52.66$\pm$1.32          & \textbf{53.86$\pm$1.54} & {\ul 53.44$\pm$1.64}    \\ \hline
	\end{tabular}
\caption{Performance metrics comparisons on Face95 database between K-means, NMF, GNMF, HNMF, NTR-MU, GNTR-MU, HGNTR-MU and LraHGNTR-MU methods. We divide the database into different subset. Each subset contains $k_{Face95}$ categories. }~\label{Face95_clustering}
\end{table}

\renewcommand\arraystretch{1.0}
\begin{table}[!htb]
	\setlength{\tabcolsep}{0.8mm}
	\begin{tabular}{lccccccccc}
		\hline
		& \multicolumn{1}{l}{$k_{Face96}$}    & K-means     & NMF                 & GNMF       & HNMF       & NTR-MU     & GNTR-MU    & HGNTR-MU            & lraHGNTR-MU         \\ \hline
		\multirow{5}{*}{ACC} & 20                      & 76.80$\pm$6.96 & 77.00$\pm$3.82          & 77.65$\pm$5.70 & 79.05$\pm$2.04 & 66.35$\pm$3.89 & 75.90$\pm$4.13 & {\ul 80.05$\pm$4.73}    & \textbf{81.00$\pm$6.14} \\
		& 30                      & 74.40$\pm$2.48 & {\ul 76.83$\pm$2.69}    & 72.76$\pm$3.87 & 72.70$\pm$3.93 & 71.63$\pm$2.45 & 71.66$\pm$2.72 & 76.67$\pm$4.79          & \textbf{78.87$\pm$3.87} \\
		& 40                      & 69.85$\pm$4.97 & \textbf{74.60$\pm$2.35} & 66.97$\pm$3.66 & 69.17$\pm$4.10 & 66.27$\pm$2.21 & 71.57$\pm$2.71 & 72.25$\pm$2.24          & {\ul 72.47$\pm$4.03}    \\
		& 50                      & 68.46$\pm$1.46 & 68.76$\pm$1.33          & 69.02$\pm$3.10 & 71.38$\pm$2.49 & 66.64$\pm$1.98 & 71.32$\pm$4.15 & {\ul 74.20$\pm$3.00}    & \textbf{74.32$\pm$2.06} \\
		& 60                      & 67.50$\pm$2.42 & 70.06$\pm$2.06          & 70.78$\pm$2.78 & 69.56$\pm$1.45 & 64.76$\pm$3.37 & 71.95$\pm$3.03 & {\ul 72.43$\pm$4.10}    & \textbf{72.50$\pm$2.35} \\ \hline
		\multirow{5}{*}{NMI} & 20                      & 85.55$\pm$2.77 & 85.85$\pm$1.56          & 89.25$\pm$1.97 & 89.96$\pm$0.70 & 78.44$\pm$2.90 & 88.37$\pm$1.74 & \textbf{89.98$\pm$2.52} & {\ul 89.40$\pm$2.94}    \\
		& 30                      & 86.75$\pm$1.73 & 88.23$\pm$1.11          & 88.81$\pm$1.47 & 89.58$\pm$1.80 & 85.57$\pm$1.88 & 89.87$\pm$1.06 & \textbf{91.20$\pm$1.10} & {\ul 91.08$\pm$1.31}    \\
		& 40                      & 86.77$\pm$1.40 & 88.02$\pm$0.56          & 87.52$\pm$0.52 & 88.05$\pm$1.50 & 84.37$\pm$0.95 & 89.37$\pm$0.57 & \textbf{89.57$\pm$0.81} & {\ul 89.55$\pm$0.85}    \\
		& 50                      & 86.52$\pm$1.01 & 87.20$\pm$0.40          & 88.04$\pm$0.95 & 88.10$\pm$1.08 & 85.08$\pm$0.72 & 89.62=1.08 & \textbf{90.37$\pm$0.85} & {\ul 89.96$\pm$0.53}    \\
		& 60                      & 86.88$\pm$1.03 & 88.57$\pm$0.47          & 88.55$\pm$0.81 & 88.29$\pm$0.70 & 85.10$\pm$1.87 & 90.29$\pm$0.83 & {\ul 90.86$\pm$0.52}    & \textbf{91.02$\pm$0.43} \\ \hline
		\multirow{5}{*}{PUR} & 20                      & 79.60$\pm$5.19 & 80.80$\pm$2.60          & 82.50$\pm$4.34 & 83.40$\pm$1.77 & 70.30$\pm$4.00 & 81.20$\pm$2.55 & 83.10$\pm$4.28          & \textbf{84.10$\pm$4.97} \\
		& 30                      & 77.73$\pm$2.03 & 80.36$\pm$1.76          & 78.36$\pm$2.83 & 78.70$\pm$3.18 & 75.10$\pm$1.73 & 77.83=1.92 & 81.30$\pm$3.47          & \textbf{82.93$\pm$2.36} \\
		& 40                      & 74.55$\pm$3.39 & 78.57$\pm$1.38          & 73.32$\pm$2.32 & 75.32$\pm$3.02 & 70.35$\pm$1.73 & 77.20$\pm$1.79 & {\ul 77.50$\pm$2.03}    & 76.75$\pm$2.60          \\
		& 50                      & 72.70$\pm$1.23 & 73.48$\pm$1.01          & 74.26$\pm$2.15 & 76.04$\pm$1.81 & 69.98$\pm$1.94 & 76.78$\pm$3.18 & \textbf{78.48$\pm$2.41} & {\ul 78.34$\pm$1.54}    \\
		& \multicolumn{1}{c}{60} & 72.15$\pm$1.86 & 75.13$\pm$1.35          & 75.30$\pm$1.89 & 74.41$\pm$0.62 & 68.81$\pm$2.96 & 76.65$\pm$2.40 & \textbf{77.73$\pm$2.02} & {\ul 77.00$\pm$0.82}    \\ \hline
	\end{tabular}
\caption{Performance metrics comparisons on Face96 database between K-means, NMF, GNMF, HNMF, NTR-MU, GNTR-MU, HGNTR-MU and LraHGNTR-MU methods. We divide the database into different subset. Each subset contains $k_{Face96}$ categories. }~\label{Face96_clustering}
\end{table}

\begin{figure}[!tb] 
	\centering  
	\subfigure[ ]{
		\includegraphics[width=0.48\textwidth]{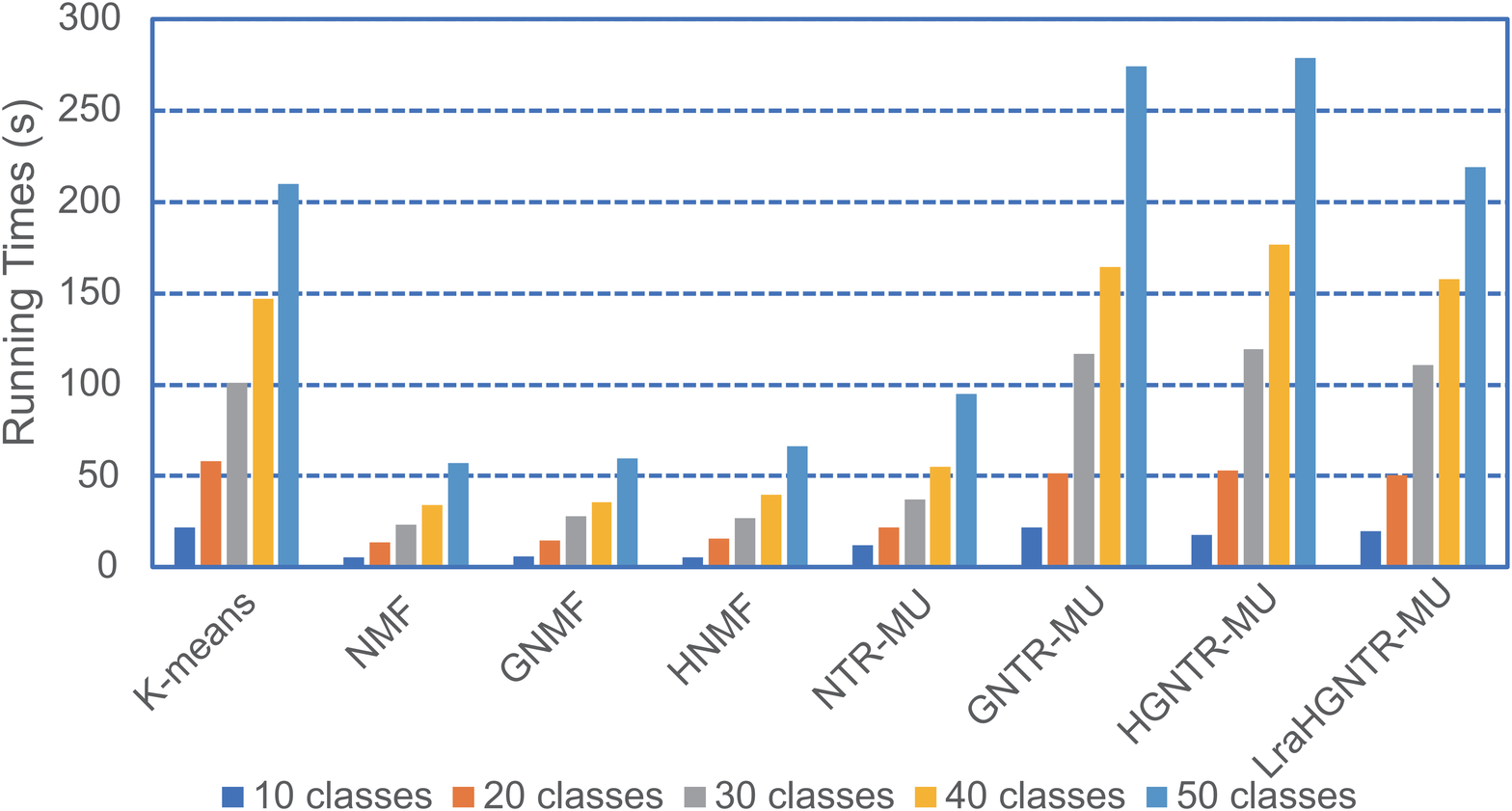}}
	\subfigure[ ]{
		\includegraphics[width=0.49\textwidth]{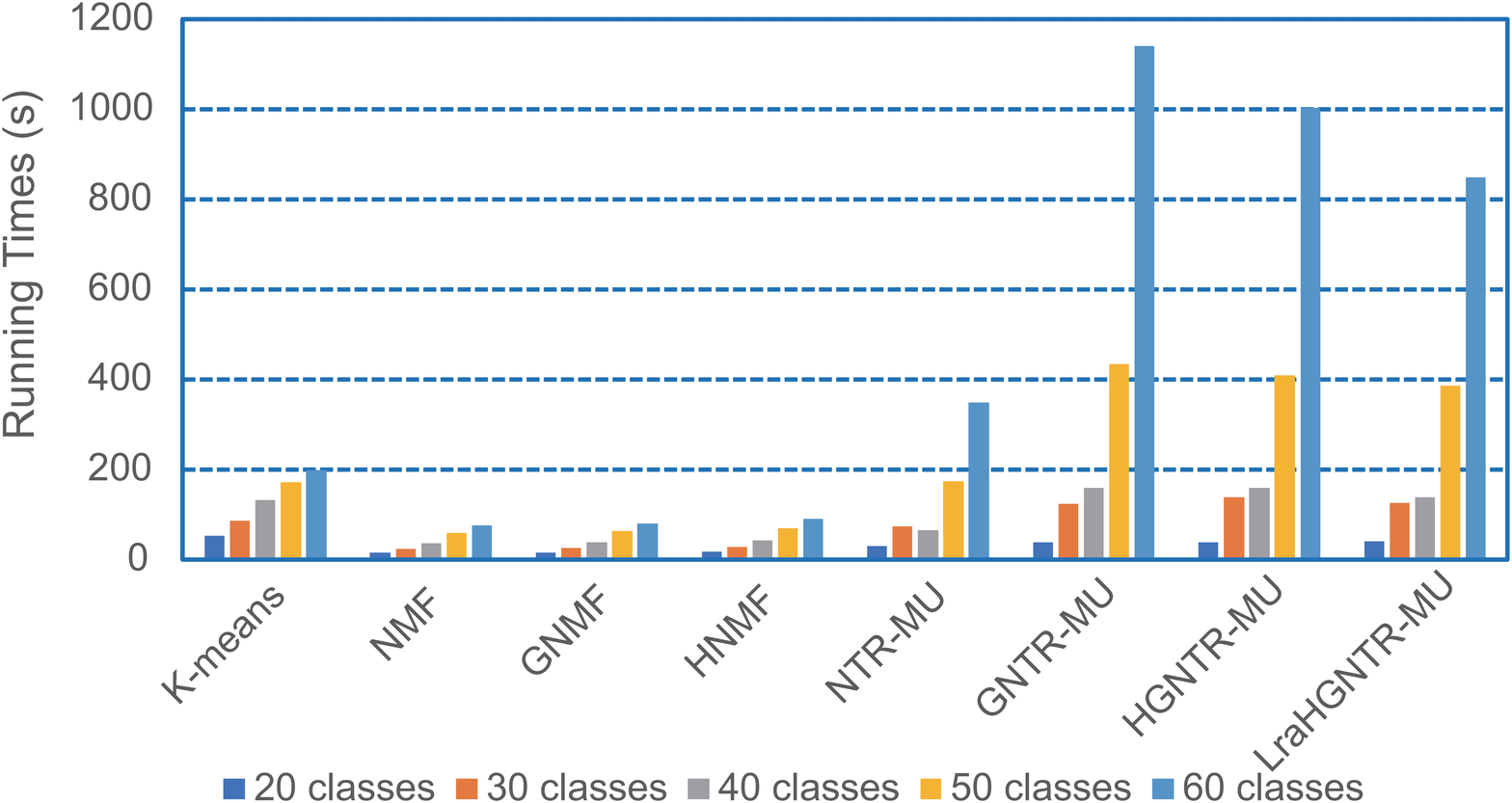}}
	\caption{The running time of various algorithms subset in different number of categories. (a) ORL. (b) GT.
	}
	\label{fig:runtime2} 
\end{figure}

\subsection{Clustering Tasks for Real-world Database with Gaussian Noise}
\label{sec:4.5}

\begin{figure}[!htb]  
	\centering 	
	\subfigure[ ]{		\includegraphics[width=0.32\textwidth]{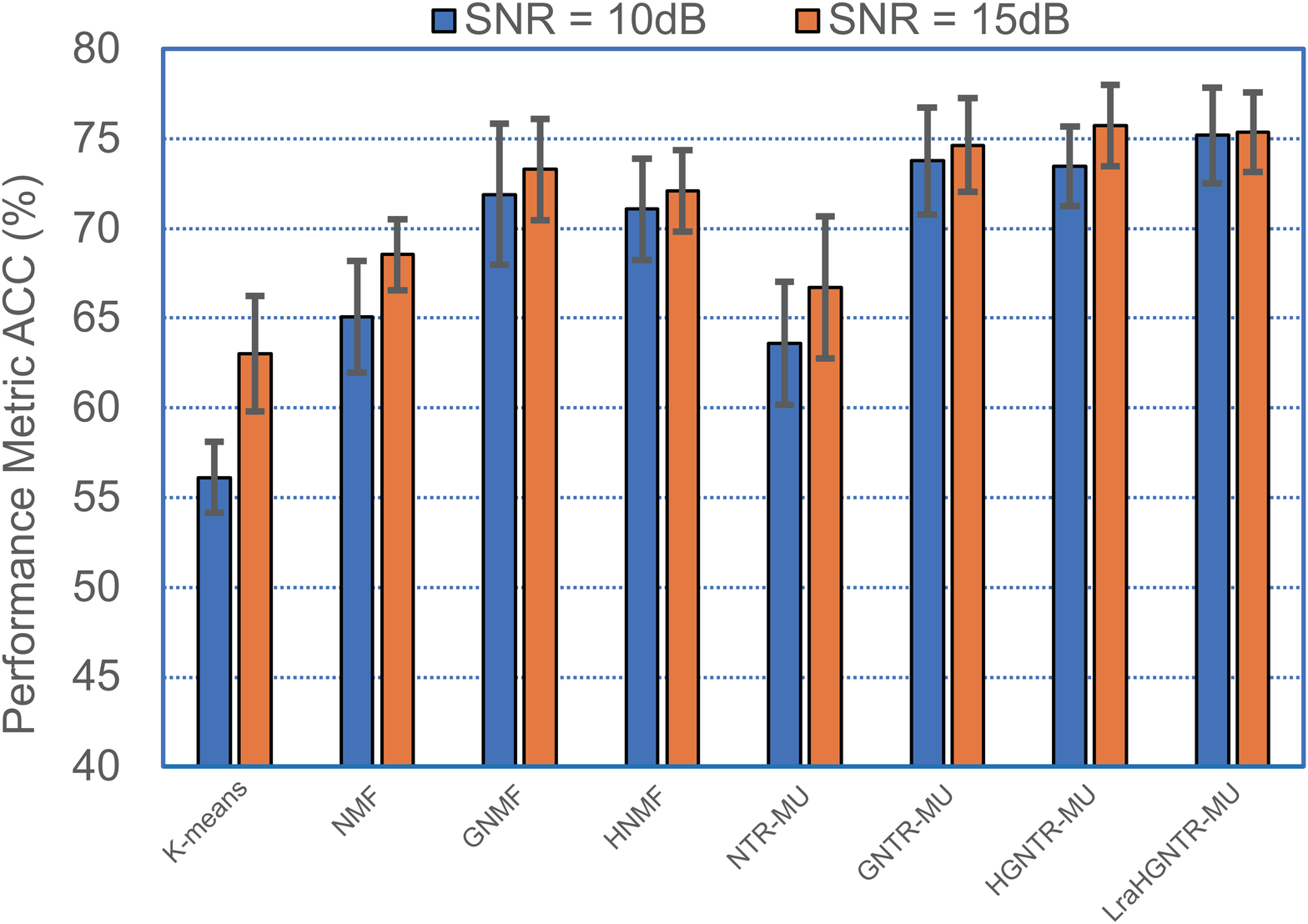}}
	\subfigure[ ]{
		\includegraphics[width=0.32\textwidth]{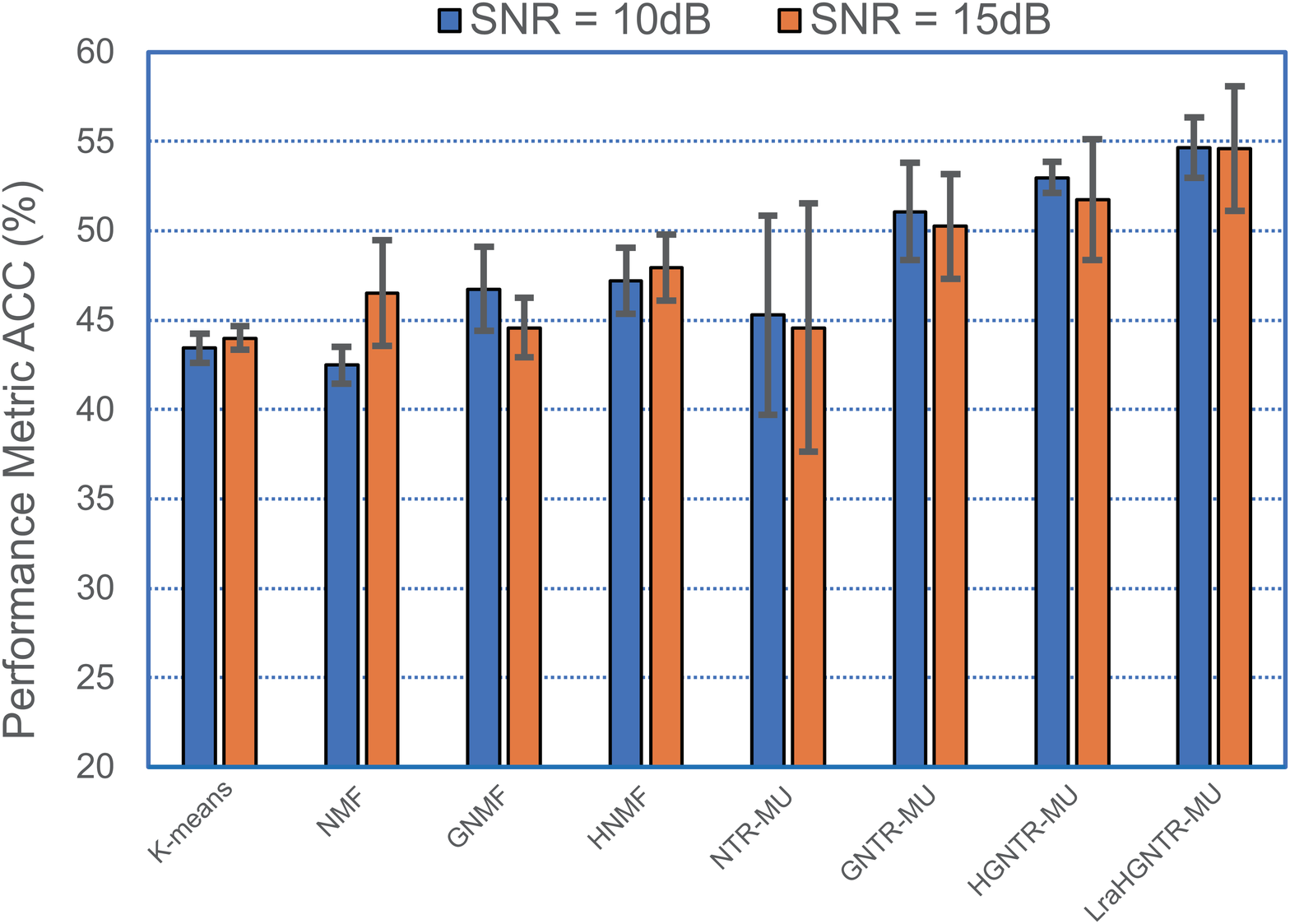}}
	\subfigure[ ]{
		\includegraphics[width=0.32\textwidth]{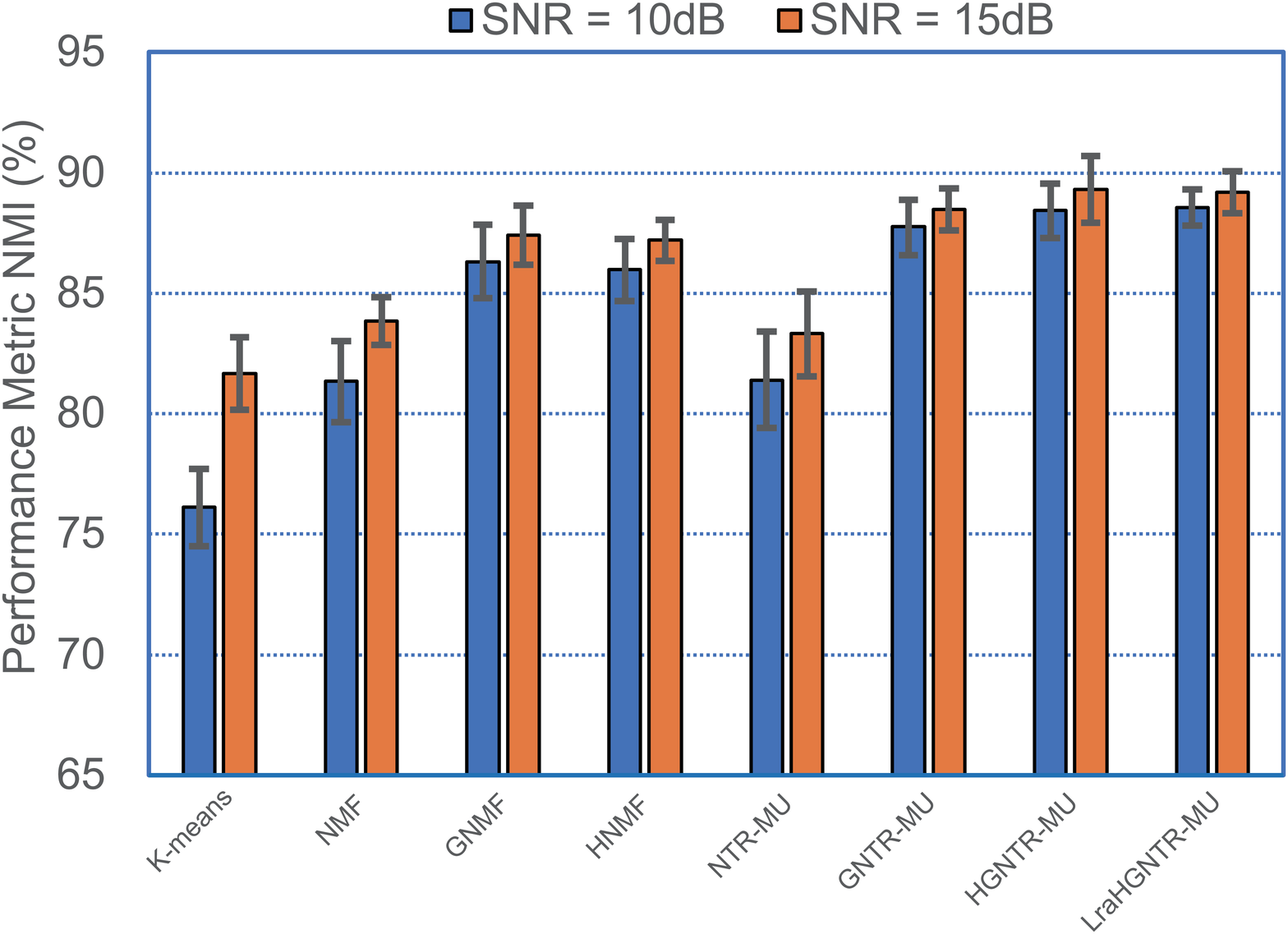}}
	\caption{Performance metrics $\left(\%\right)$ comparisons on between K-means, NMF, GNMF, HNMF, NTR-MU, GNTR-MU, HGNTR-MU and LraHGNTR-MU methods for noise-added ORL and GT database. (a) ACC on ORL. (b) ACC on GT. (c) NMI on ORL.
	}
	\label{noise_1} 
\end{figure}

\begin{figure}[!htb]  
	\centering  
	\subfigure[ ]{
		\includegraphics[width=0.32\textwidth]{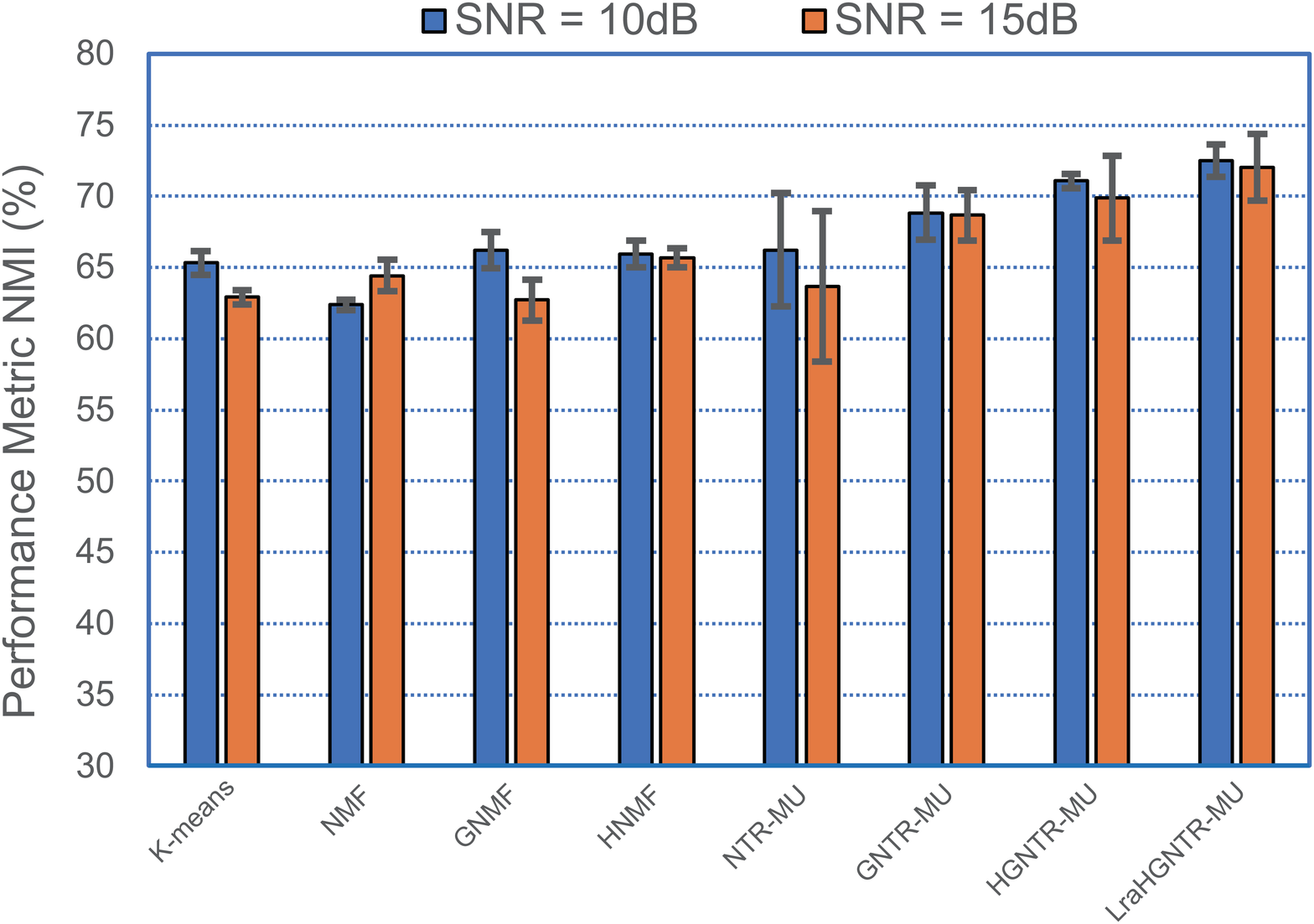}}
	\subfigure[ ]{
		\includegraphics[width=0.32\textwidth]{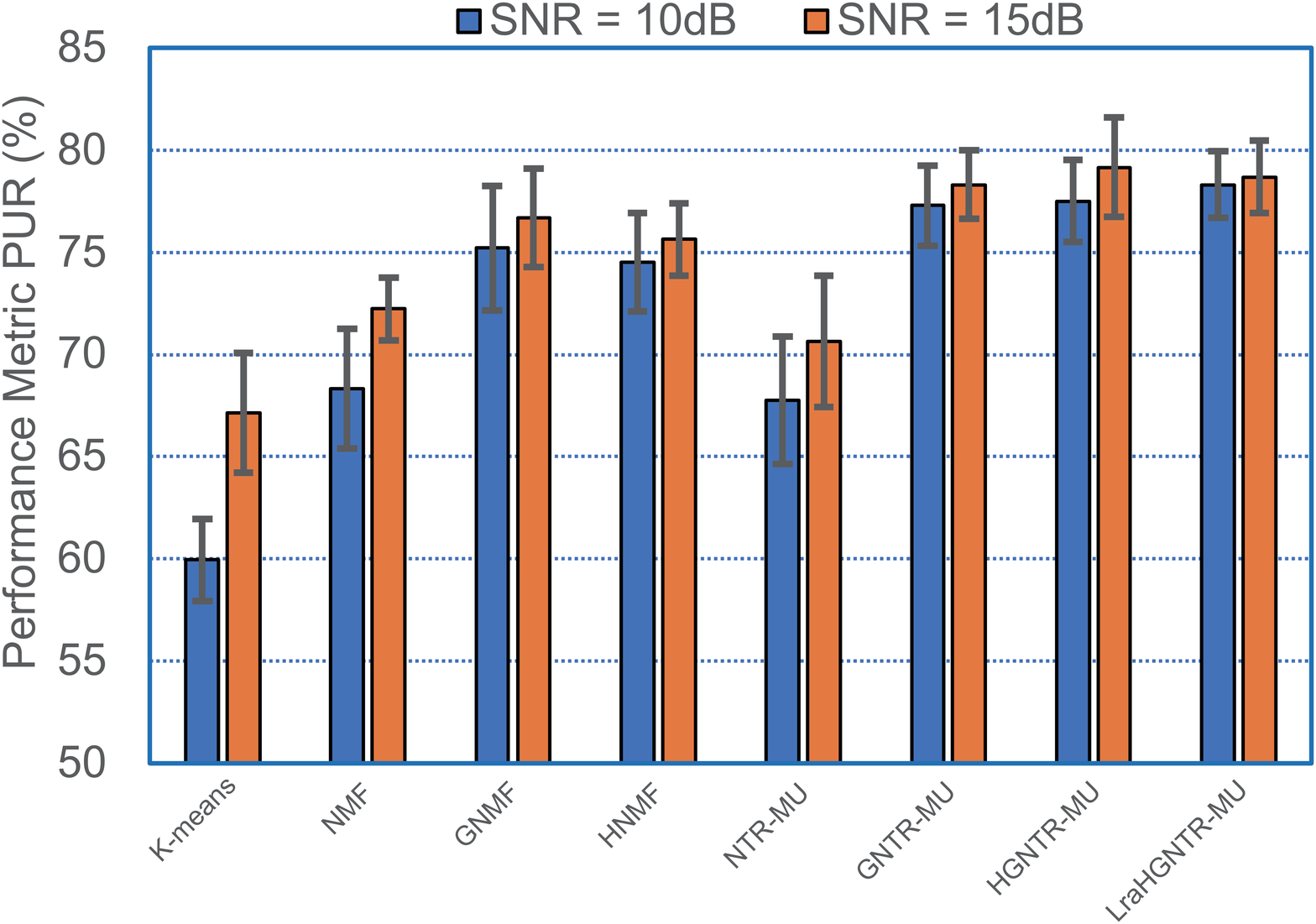}}
	\subfigure[ ]{
		\includegraphics[width=0.32\textwidth]{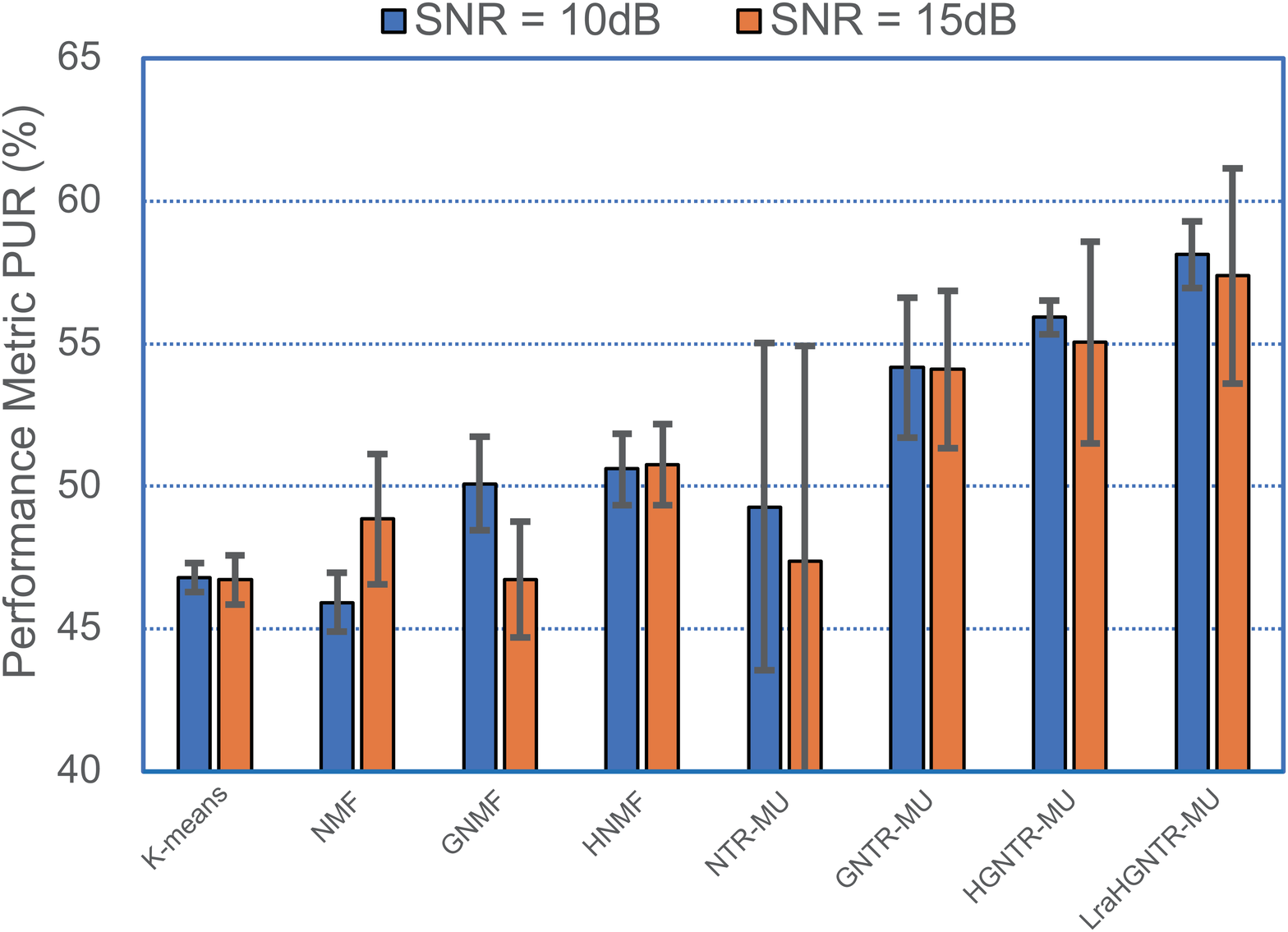}}
	\caption{Performance metrics $\left(\%\right)$ comparisons on between K-means, NMF, GNMF, HNMF, NTR-MU, GNTR-MU, HGNTR-MU and LraHGNTR-MU methods for noisy-added ORL and GT database. (a) ACC on GT. (b) ACC on ORL. (c) PUR on GT.
	}
	\label{noise_2} 
\end{figure}

\begin{figure}[!htb]  
	\centering  
	\subfigure[ ]{
		\includegraphics[width=0.32\textwidth]{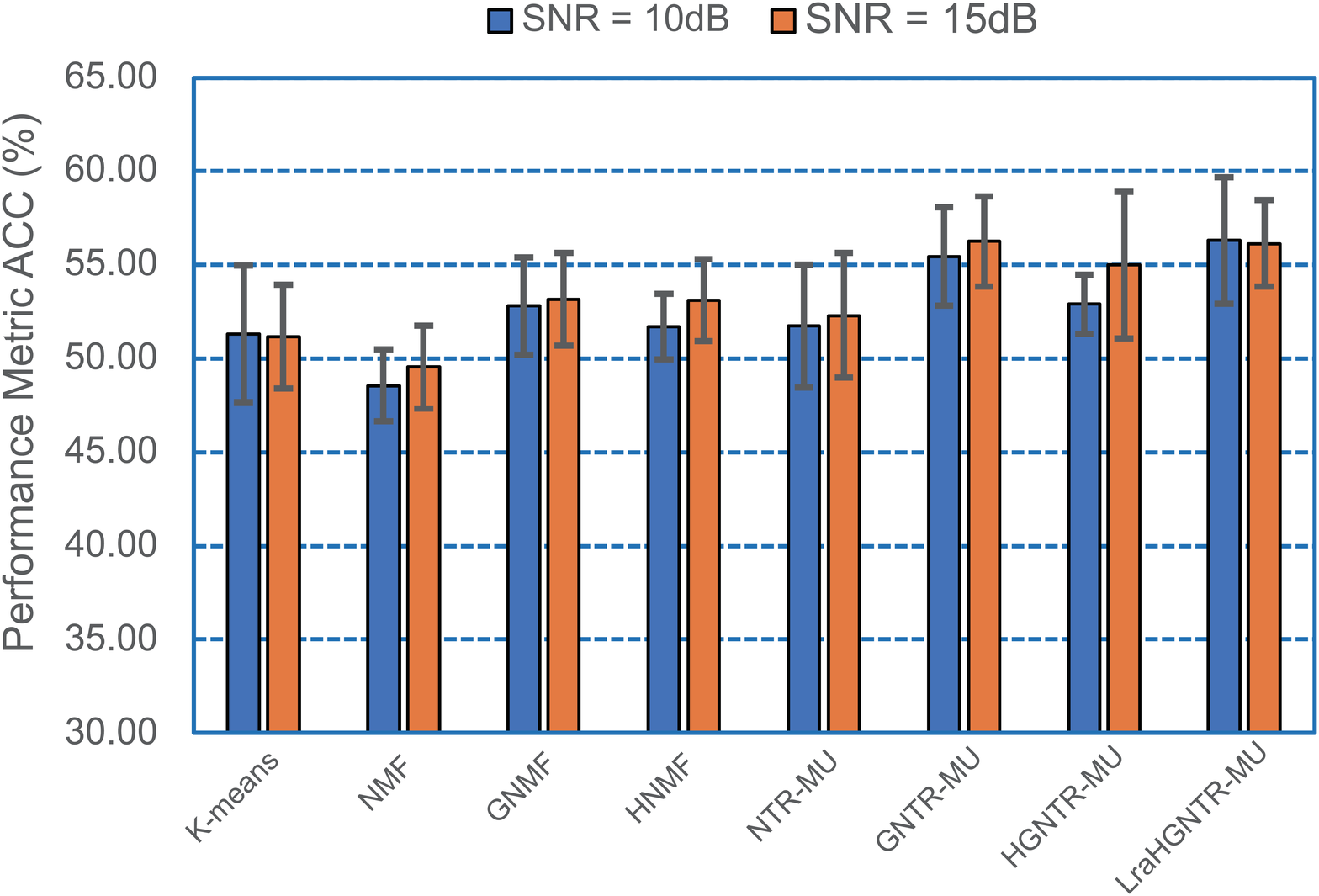}}
	\subfigure[ ]{
		\includegraphics[width=0.322\textwidth]{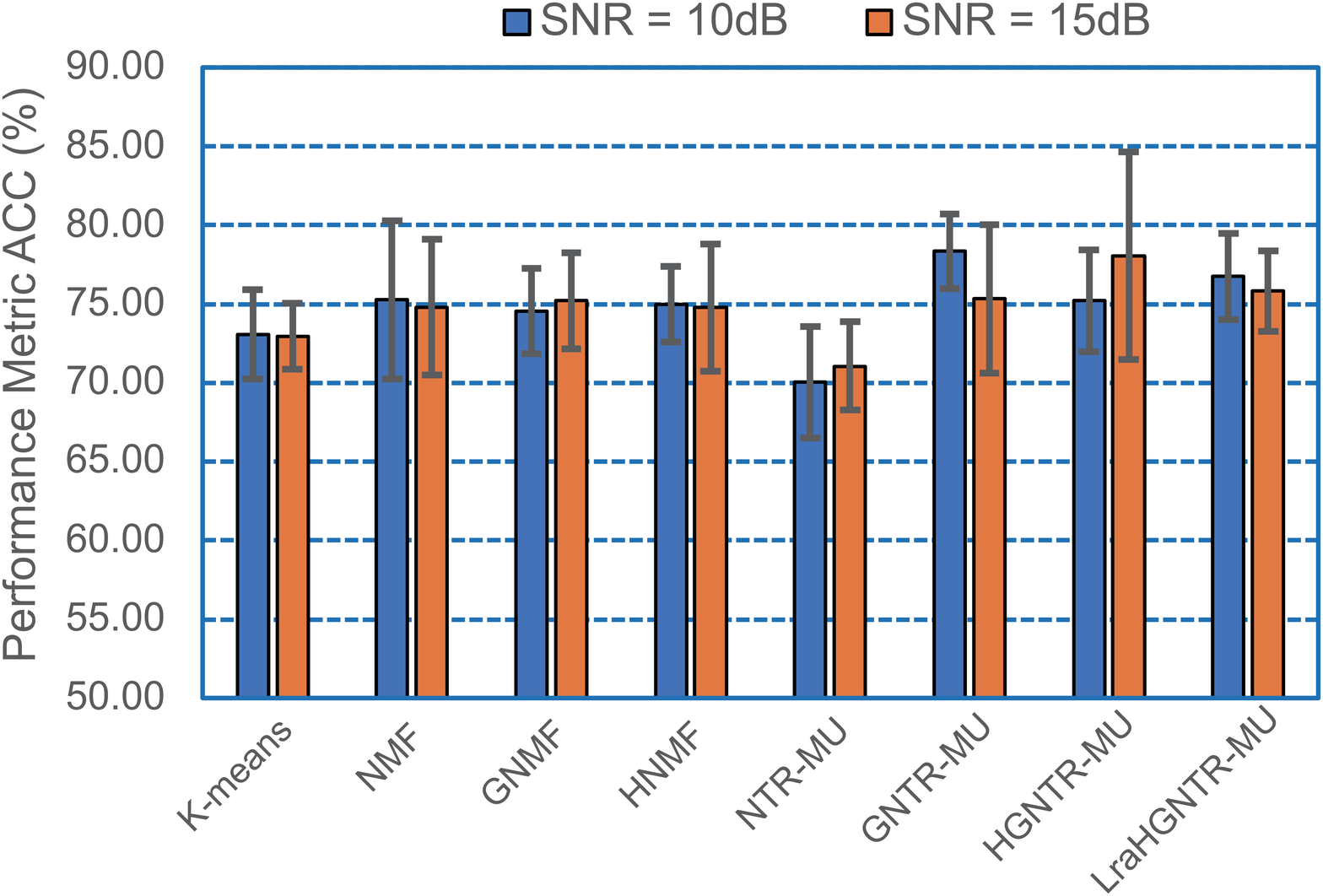}}
	\subfigure[ ]{
		\includegraphics[width=0.32\textwidth]{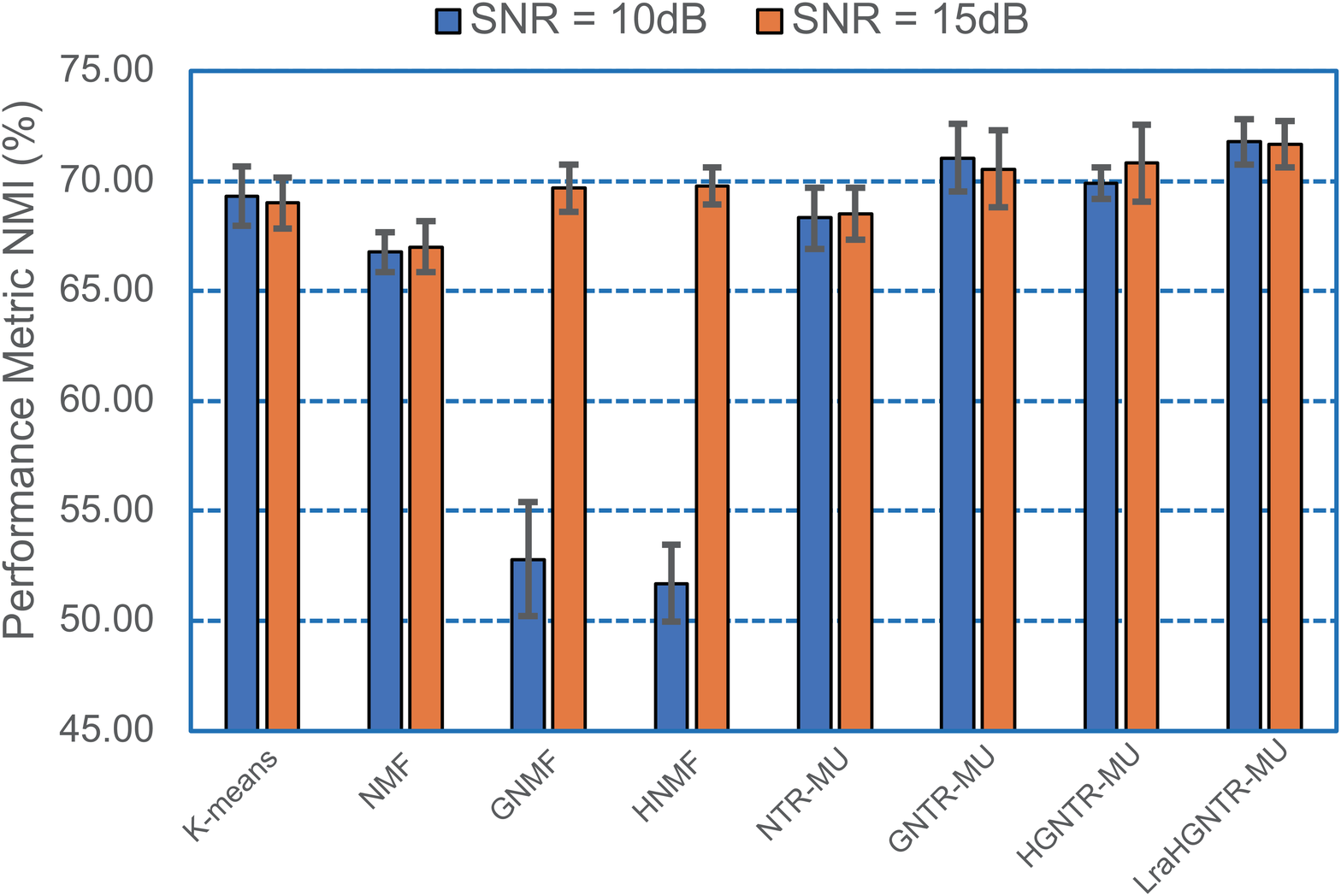}}
	\caption{Performance metrics $\left(\%\right)$ comparisons on between K-means, NMF, GNMF, HNMF, NTR-MU, GNTR-MU, HGNTR-MU and LraHGNTR-MU methods for noisy-added ORL and GT database. (a) NMI on Face95. (b) PUR on Face96. (c) PUR on Face95.
	}
	\label{noise_3} 
\end{figure}

\begin{figure}[!htb]  
	\centering  
	\subfigure[ ]{
		\includegraphics[width=0.32\textwidth]{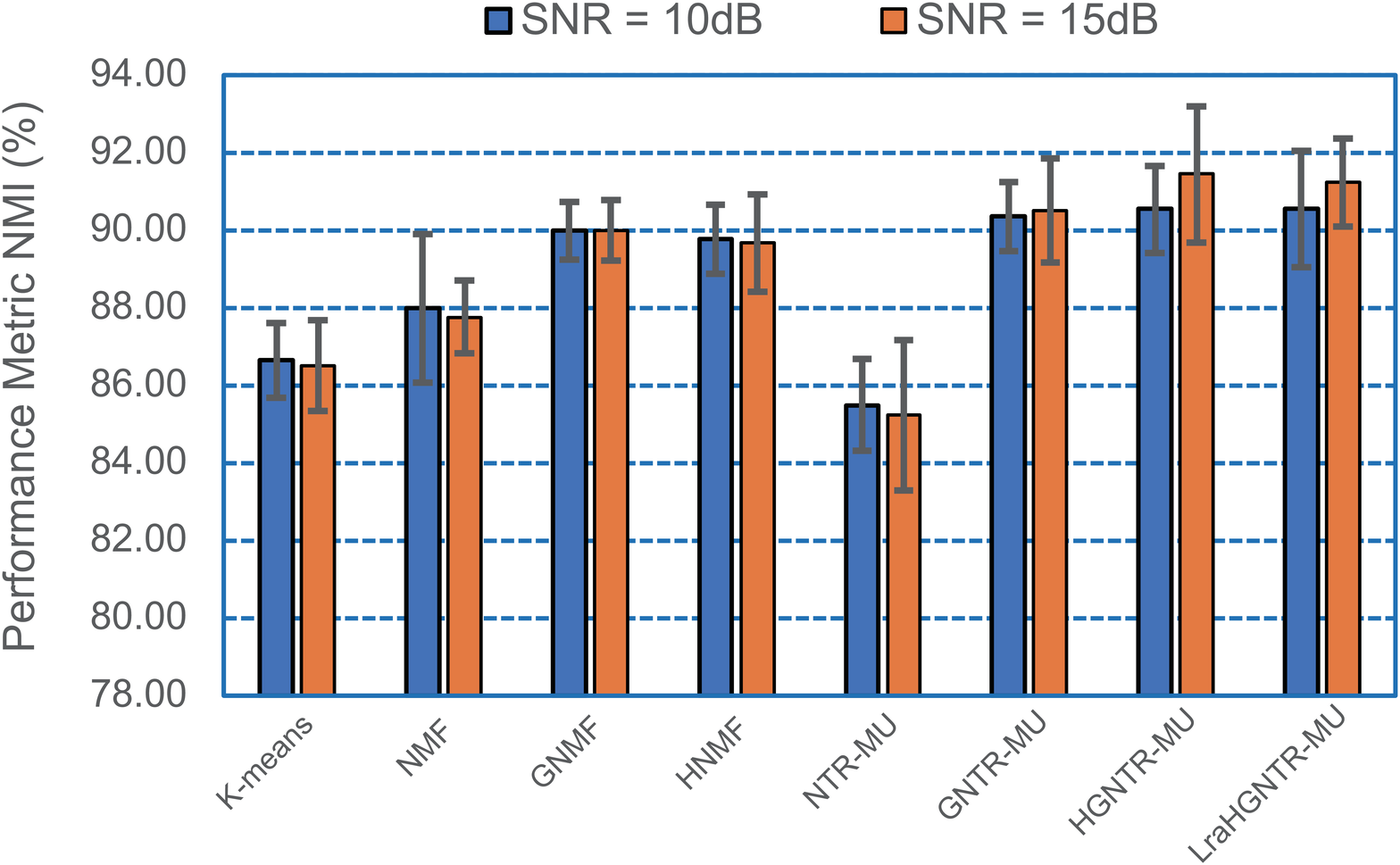}}
	\subfigure[ ]{
		\includegraphics[width=0.32\textwidth]{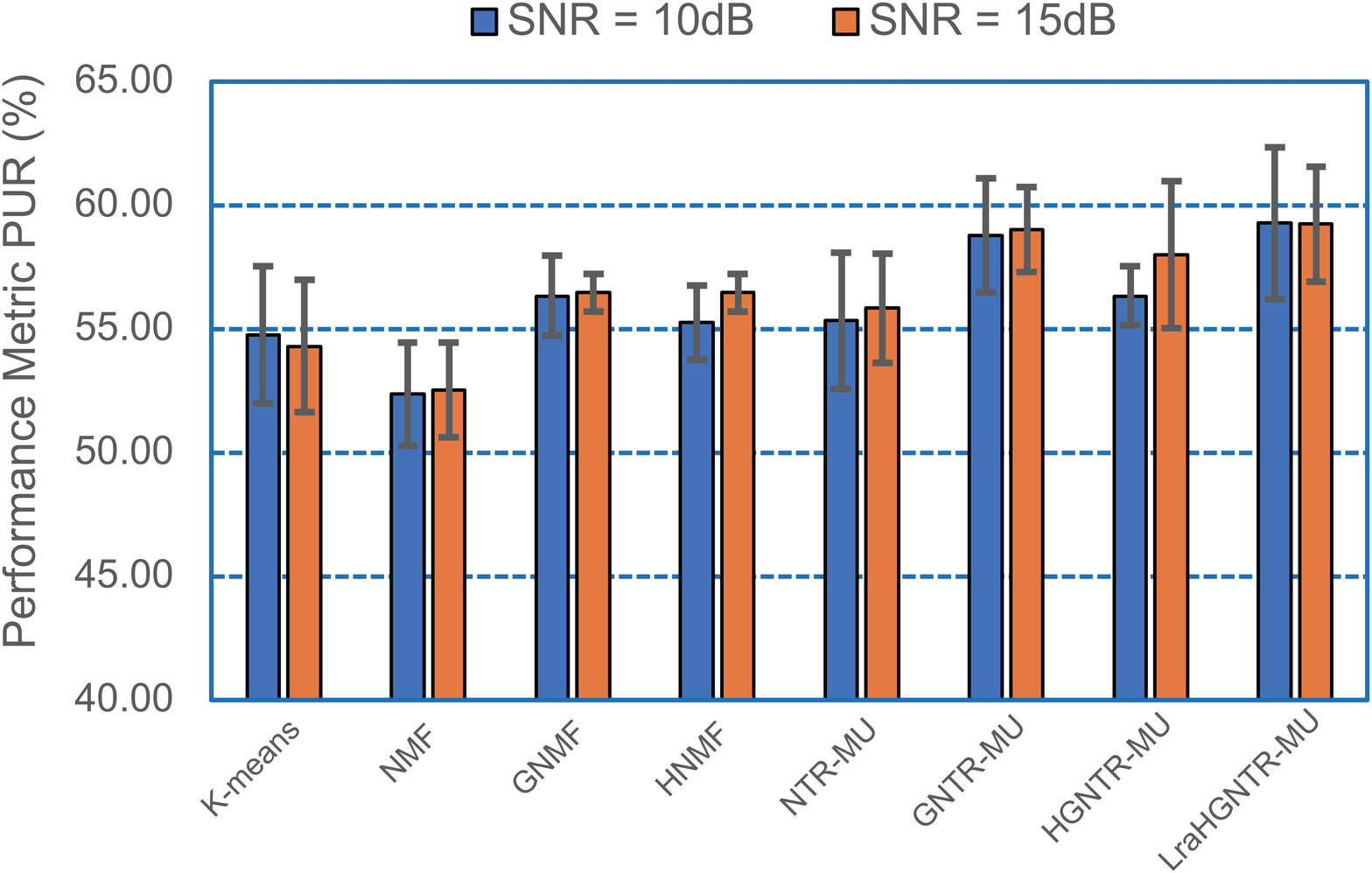}}
	\subfigure[ ]{
		\includegraphics[width=0.32\textwidth]{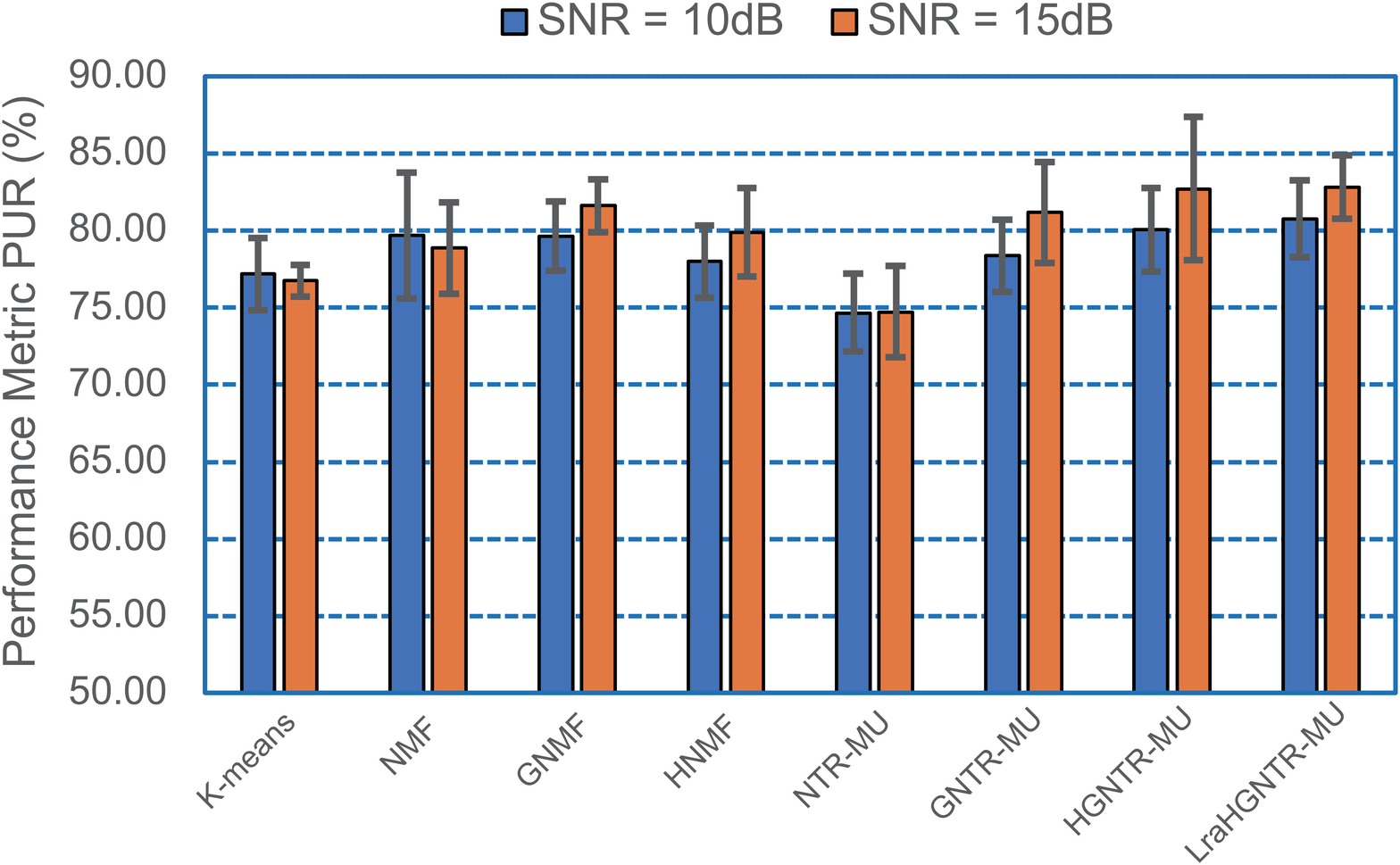}}
	\caption{Performance metrics $\left(\%\right)$ comparisons on between K-means, NMF, GNMF, HNMF, NTR-MU, GNTR-MU, HGNTR-MU and LraHGNTR-MU methods for noisy-added ORL and GT database. (a) NMI on Face96. (b) PUR on Face95. (c) PUR on Face96.
	}
	\label{noise_4} 
\end{figure}

In this part, we count the performance of our algorithm on the noisy dataset. In the case of noiseless, we can easily find that LraHGNTR is similar to HGNTR in accuracy, but LraHGNTR is faster. We compare various algorithm performances by adding the Gaussian noise on ORL and GT database. The degree of noise is measured by the signal-to-noise ratio (SNR). Before the experiment, as a result of the negative value may appear in datasets after adding very heavy Gaussian noise, we make the negative value in each database are truncated at zero to ensure the negativity for each algorithm except LraHGNTR. The reason is that LraHGNTR can directly deal with non-negative elements.  The  $Fig .\ref{noise_1}$, $Fig .\ref{noise_2}$, $Fig .\ref{noise_3}$ and $Fig.\ref{noise_4}$ show the comparsion with noise. Through experiment results, we can get  the following conclusions:

\begin{itemize}
	
	\item [-] For the datasets with adding Gaussian noise, our algorithms can still achieve good results compared with other algorithms, and the running time of LraHGNTR is also reduced. It proves the effectiveness of our algorithms again.
	
	\item [-] In the case of mild noise, our approximate algorithm LraHGNTR can achieve almost the same performance as the HGNTR. And in the case of high noise, such as SNR = $10\,dB$, the LraHGNTR is obviously better than HGNTR, which demonstrates that the low-rank approximation method is beneficial to filter noise. 
	
%	\item [-] For the tensor model, the STDs of HGNTR is smaller than NTR. It demonstrates that the graph regulation including normal-graph and hypergraph can reduce the sensitivity of the algorithm to the noise. And the STDs of LraHGNTR are the smallest among them. It proves the low-rank approximation method can make the model more stable, and has certain robustness to noise.
	
	\item [-] Low-rank approximation methods can eliminate the noise effectively. The reason is that in most datasets with noise, noise usually does not have the low rank property, so replacing the original tensor $\mathcal{X}$ with an approximate tensor $\tilde{\mathcal{X}}$ can ignore the noise and thus avoid the influence of noise. 
	
\end{itemize}


\begin{figure}[ht]  
	\centering  
	\subfigure[ ]{
		\includegraphics[width=0.49\textwidth]{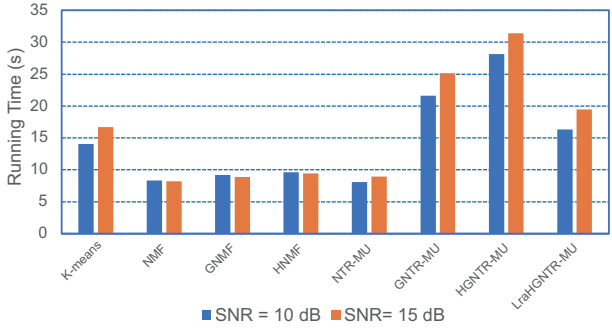}}
	\subfigure[ ]{
		\includegraphics[width=0.49\textwidth]{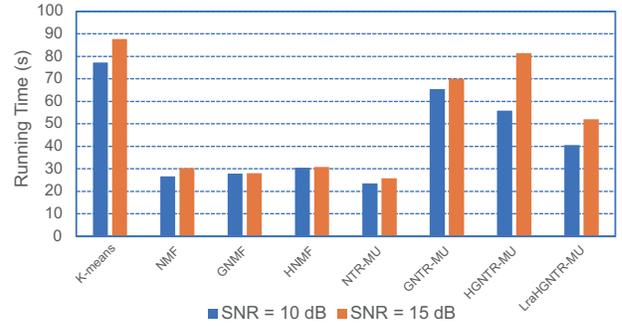}}
	\caption{Running time comparisons on between K-means, NMF, GNMF, HNMF, NTR-MU, GNTR-MU, HGNTR-MU and LraHGNTR-MU methods for noise-added ORL and GT database. (a) ORL. (b) GT.
	}
	\label{runtime3 4} 
\end{figure}
\begin{figure}[ht]  
	\centering  
	\subfigure[ ]{
		\includegraphics[width=0.48\textwidth]{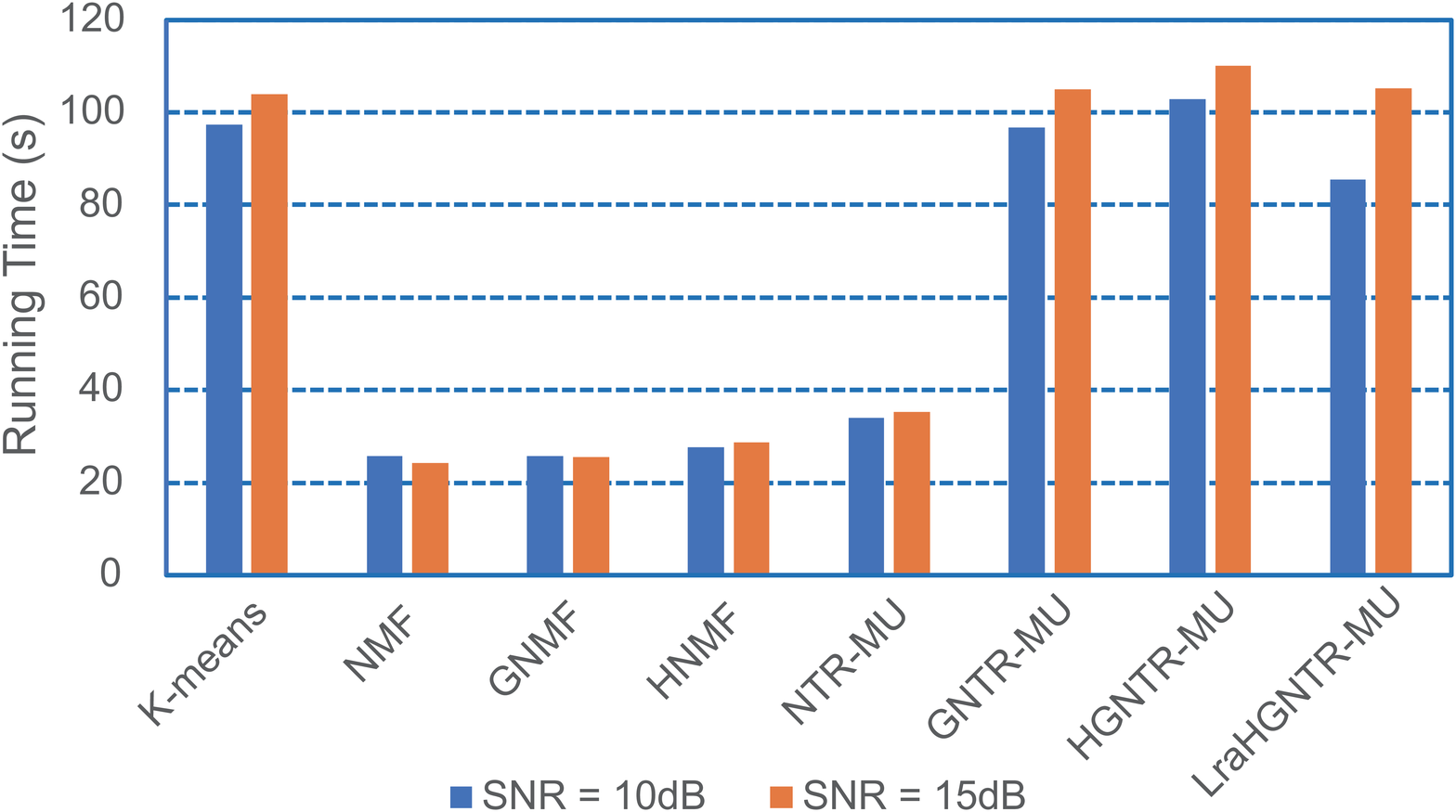}}
	\subfigure[ ]{
		\includegraphics[width=0.49\textwidth]{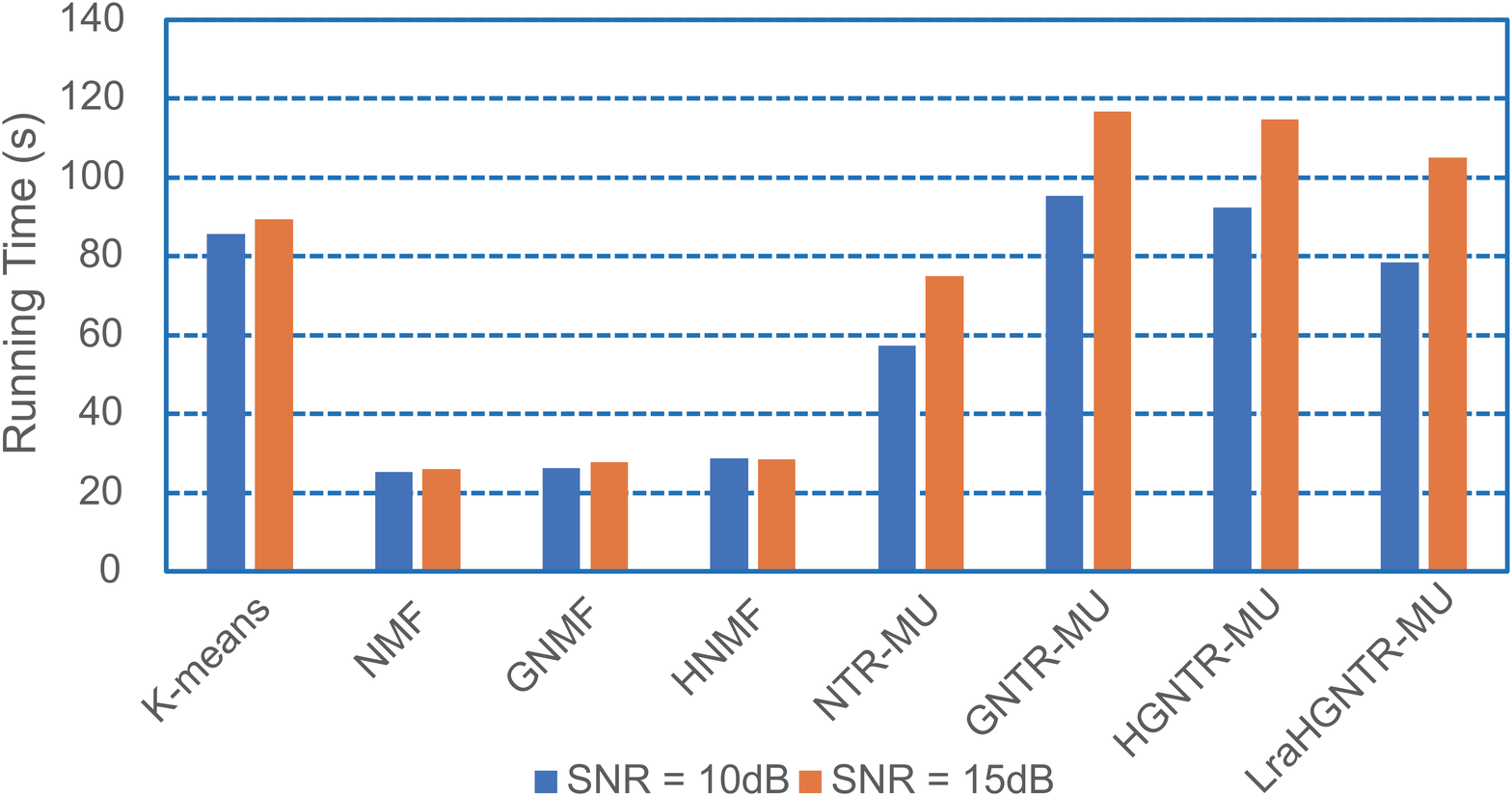}}
	\caption{Running time comparisons on between K-means, NMF, GNMF, HNMF, NTR-MU, GNTR-MU, HGNTR-MU and LraHGNTR-MU methods for noise-added Face95 and Face96 database. (a) Face95. (b) Face96.
	}
	\label{runtime5 6} 
\end{figure}
\subsection{Parameter Sensitivity}
\label{sec:4.6}
In order to investigate the effect for our model  on the situation of parameter changing, we need to carry out parameter sensitivity test. Considering that our algorithms has three parameters that can be adjusted, which are the number of inner iterations $t$, the number of nearest neighbours $k$ and the graph regularization parameter $\beta$, we use the method of controlling variables. We divide the experiment into three cases: $\left(1\right)$ fix $k=5$ and $\beta = 0.1$, and choose $t \in \{20,40,60,80\}$; $\left(2\right)$ fix $t=20$ and $\beta =0.1$, and choose $k$ to vary from $3$ to $6$; $\left(3\right)$ fix $t=20$ and $k=5$, and choose $\beta \in \{0.1,0.2,0.3,0.4,0.5,0.6\}$.

\begin{figure*}[ht]
	\subfigure[ ]{
		\includegraphics[width=0.32\textwidth]{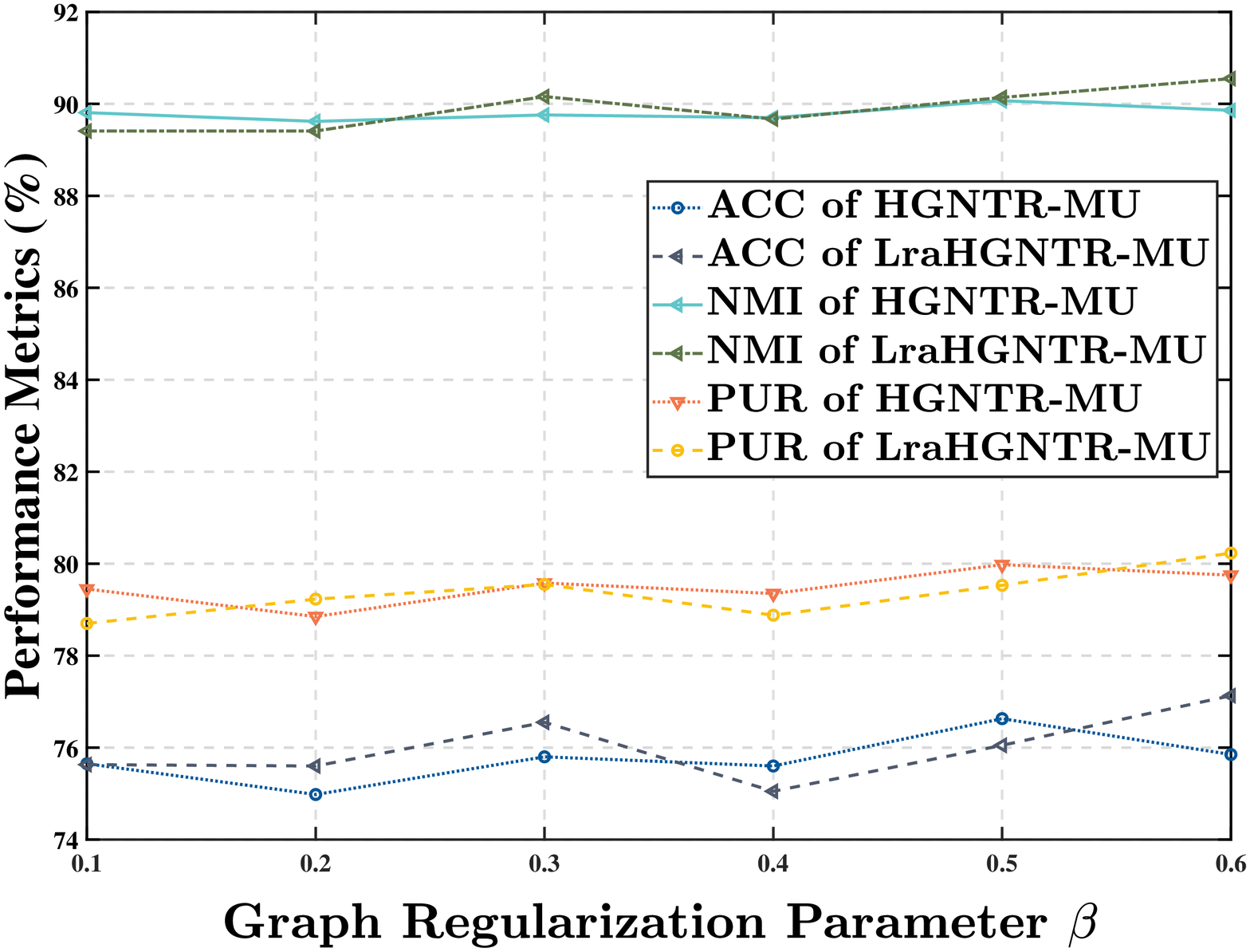}}
	\subfigure[ ]{
		\includegraphics[width=0.32\textwidth]{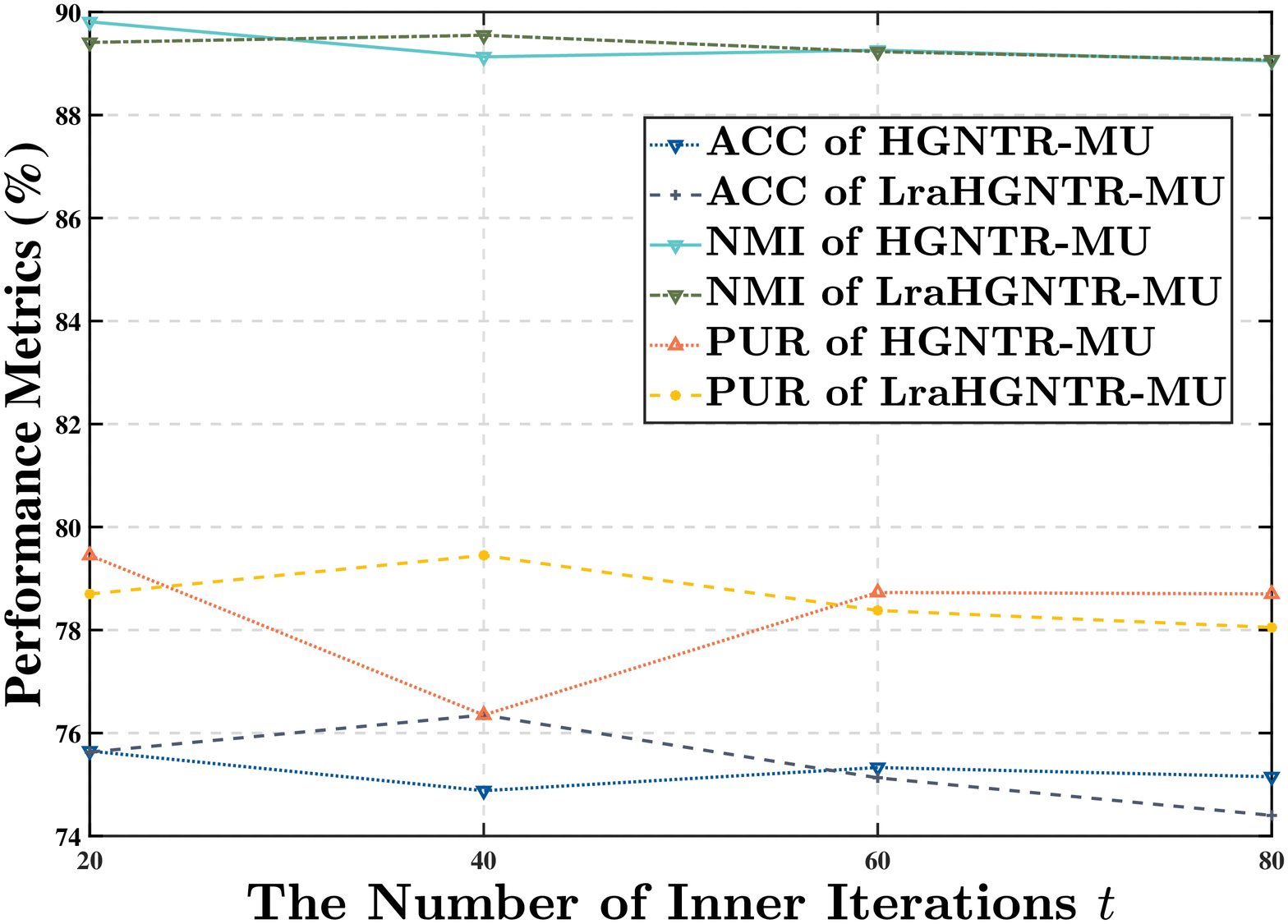}}
	\subfigure[ ]{
		\includegraphics[width=0.32\textwidth]{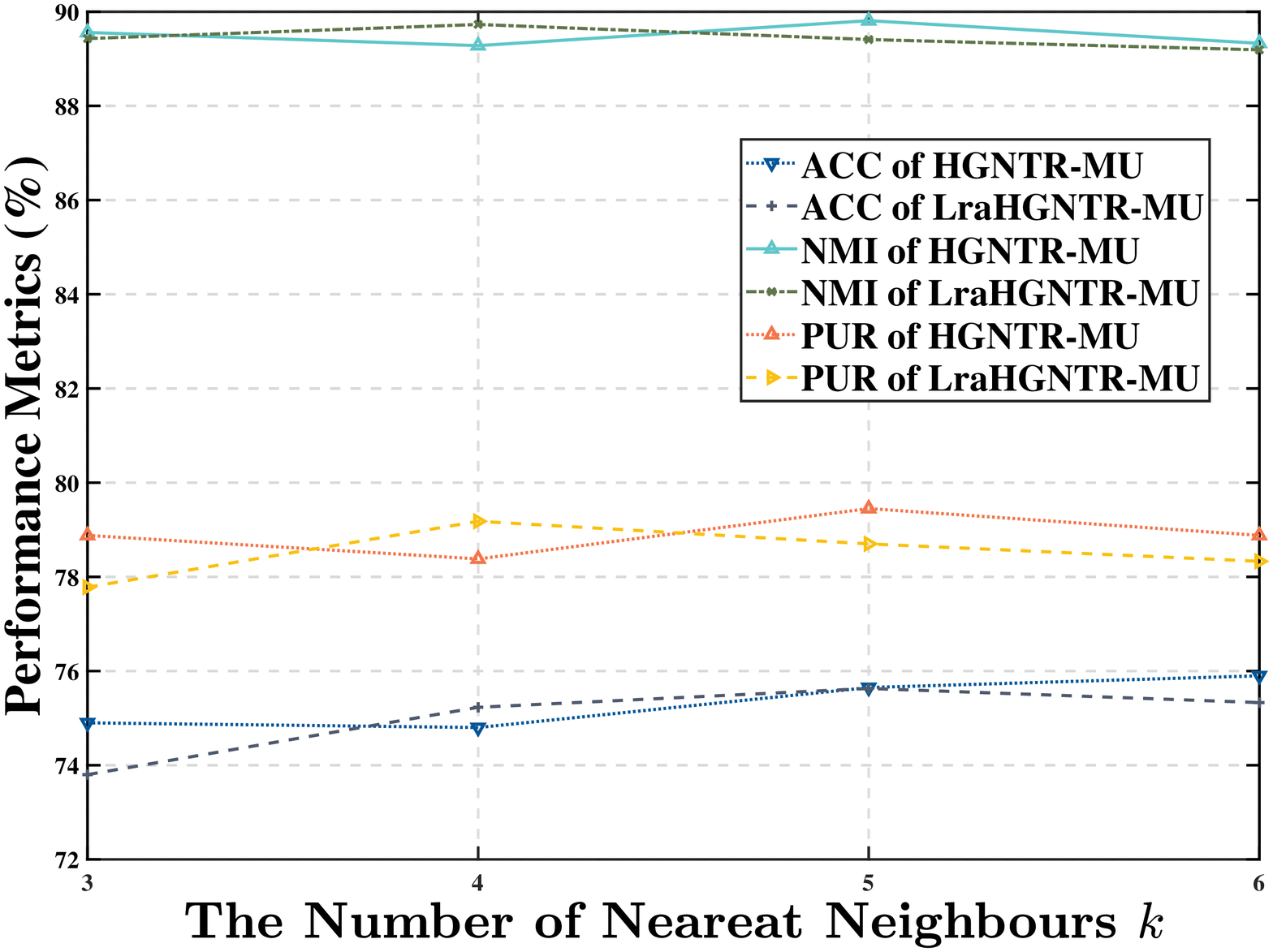}}

	\caption{The algorithm performance curve on ORL database in different parameters. (a) The graph regularization parameter $\beta$. (b) The number of inner iterations $t$. (c) The the number of nearest neighbours $k$. }
	\label{beta change sensitivity orl}
\end{figure*}

\begin{figure*}[ht]
	\subfigure[ ]{
		\includegraphics[width=0.32\textwidth]{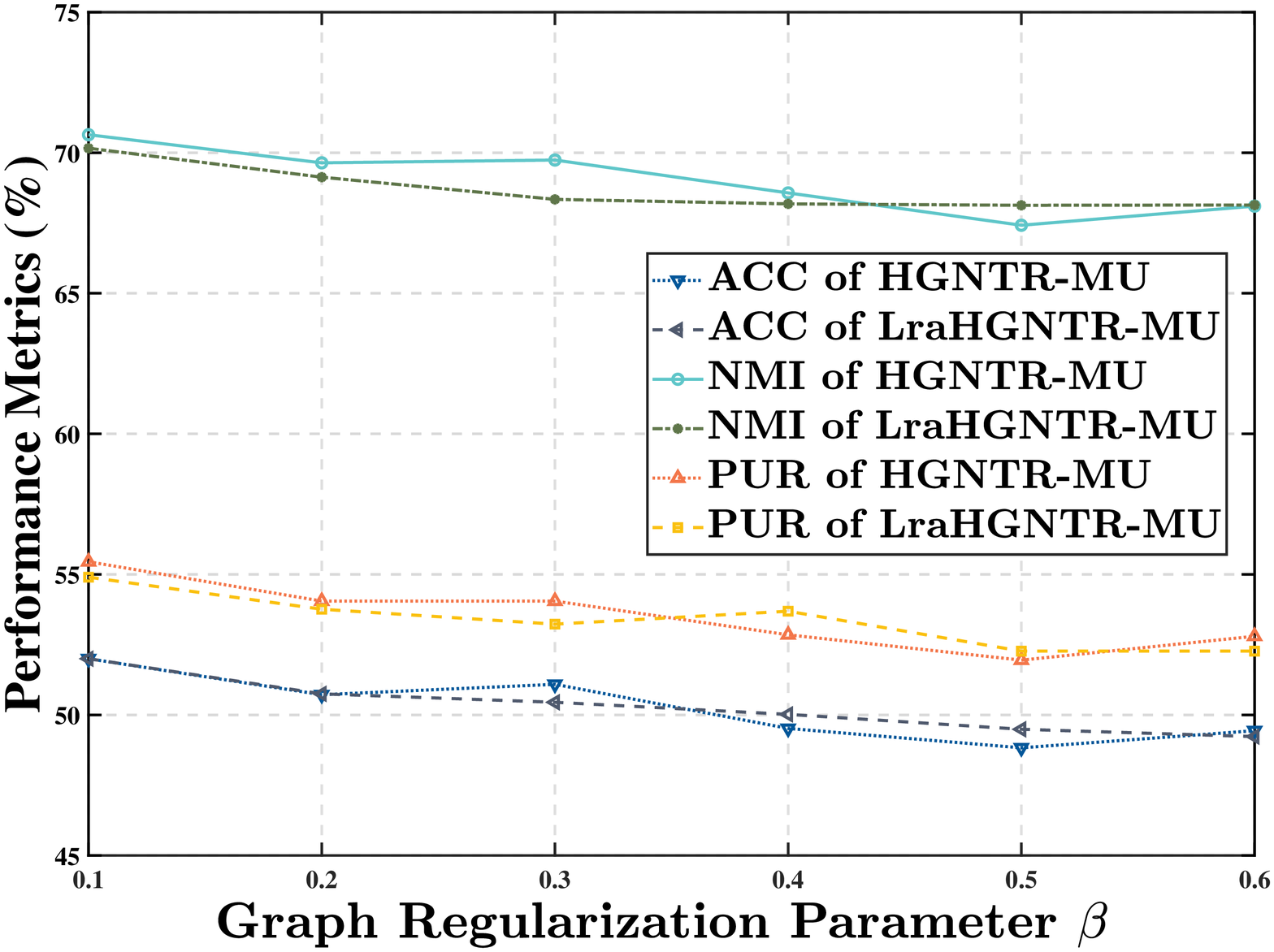}}
	\subfigure[ ]{
		\includegraphics[width=0.32\textwidth]{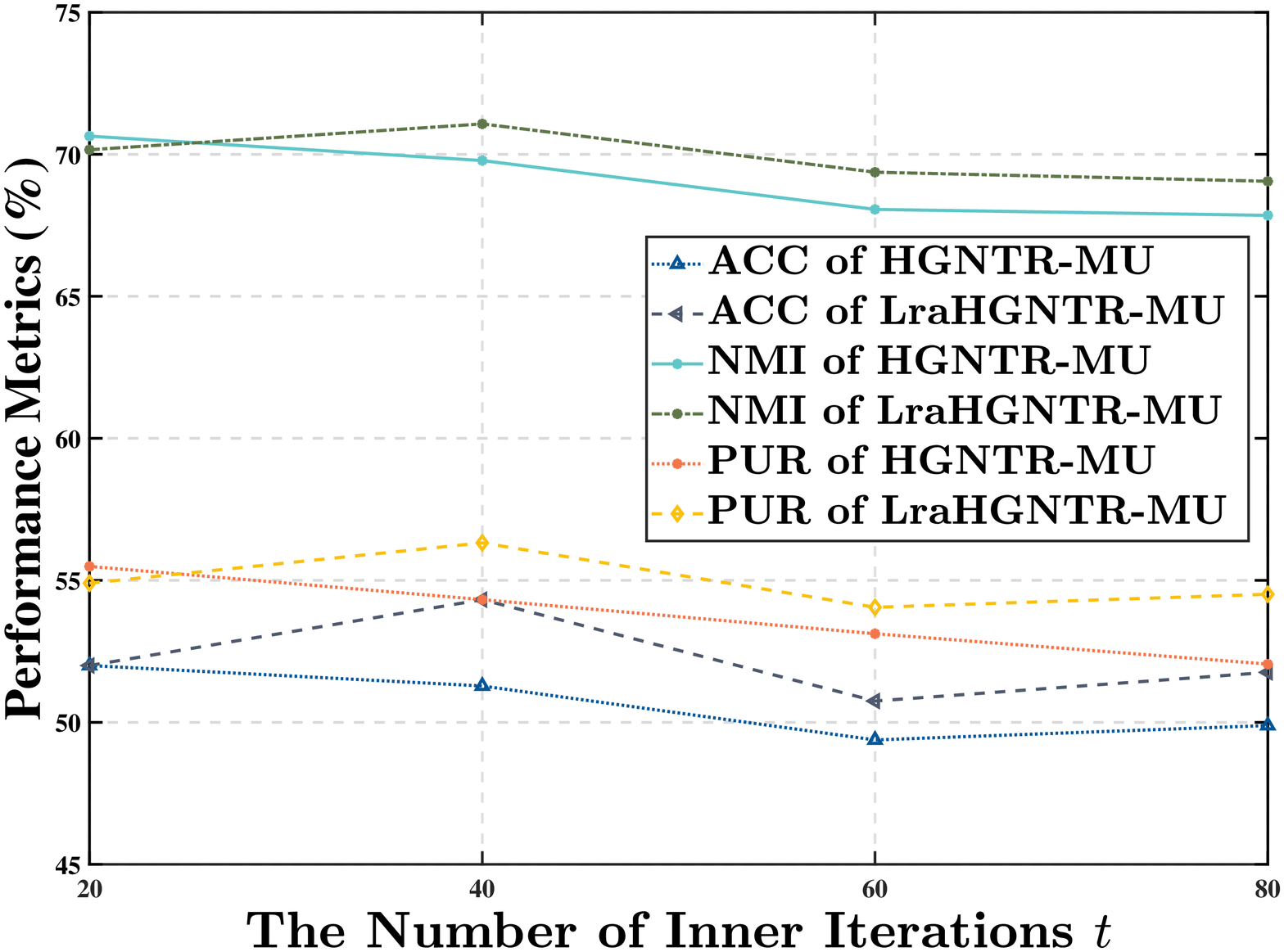}}
	\subfigure[ ]{
		\includegraphics[width=0.32\textwidth]{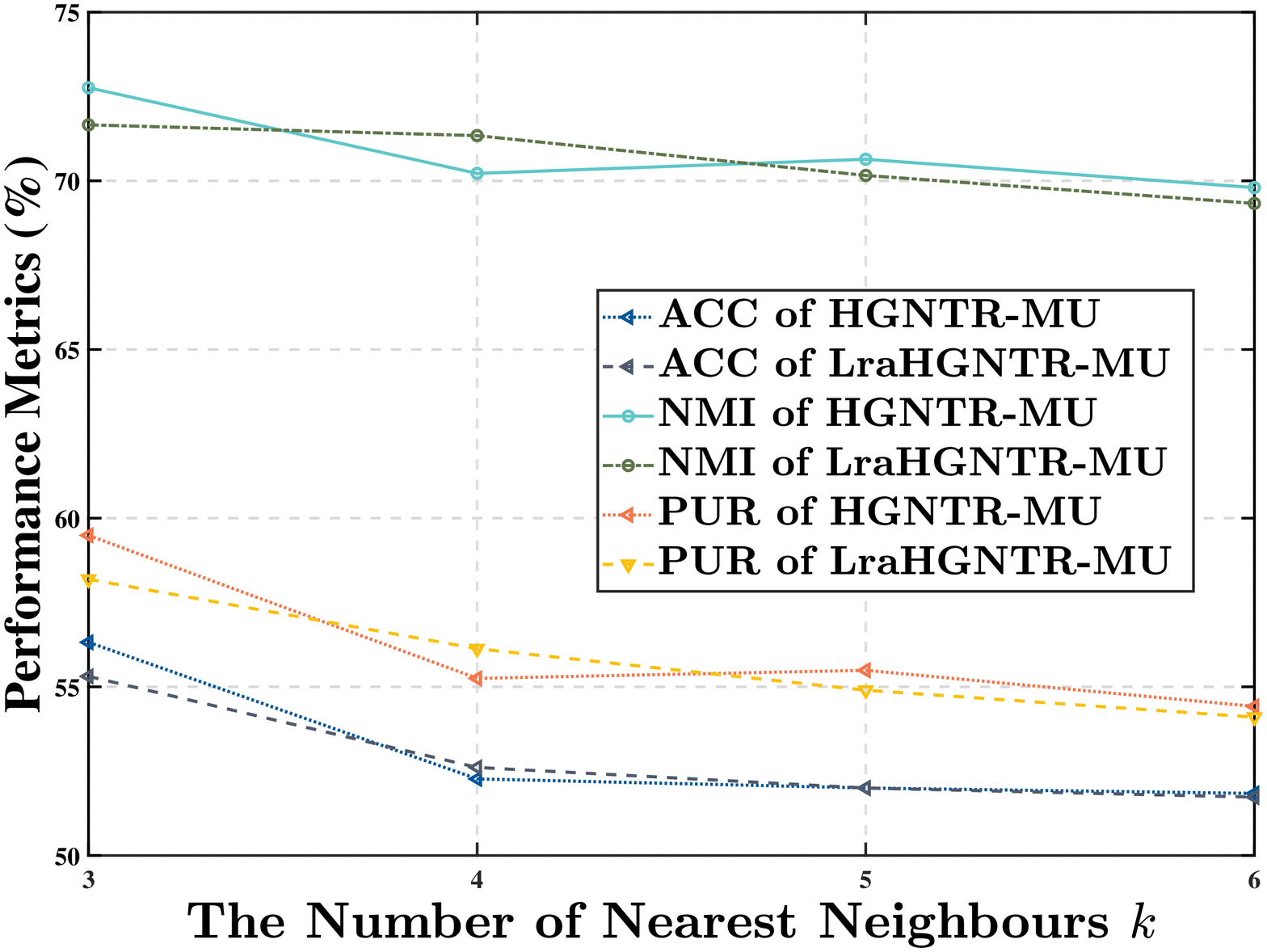}}

	\caption{The algorithm performance curve on GT database in different parameters. (a) The graph regularization parameter $\beta$. (b) The number of inner iterations $t$. (c) The  number of nearest neighbours $k$.}
	\label{t change sensitivity orl}
\end{figure*}

$Fig.\ref{beta change sensitivity orl}$ and $Fig.\ref{t change sensitivity orl}$ present the clustering performance of the HGNTR and LraHGNTR across different parameters on the two public databases. It can be observed that the performances of HGNTR and LraHGNTR algorithms across different parameters are quite stable. The ACC, NMI, and PUR change little when the number of inner iterations $t$ rises from $20$ to $80$, so $t$ can be selected around $20$ to balance the calculation cost and fitting error. It can be observed that the ACC, NMI, and PUR are stable when $k$ in the range of $3-6$. It can be noted that the hypergraph regularization parameter $\beta$ also has a relatively weak effect on the performance of our both algorithms.
\subsection{Convergence Analysis}
\label{sec:4.7}
\begin{figure*}[!ht]
	\subfigure[ ]{
		\includegraphics[width=0.30\textwidth]{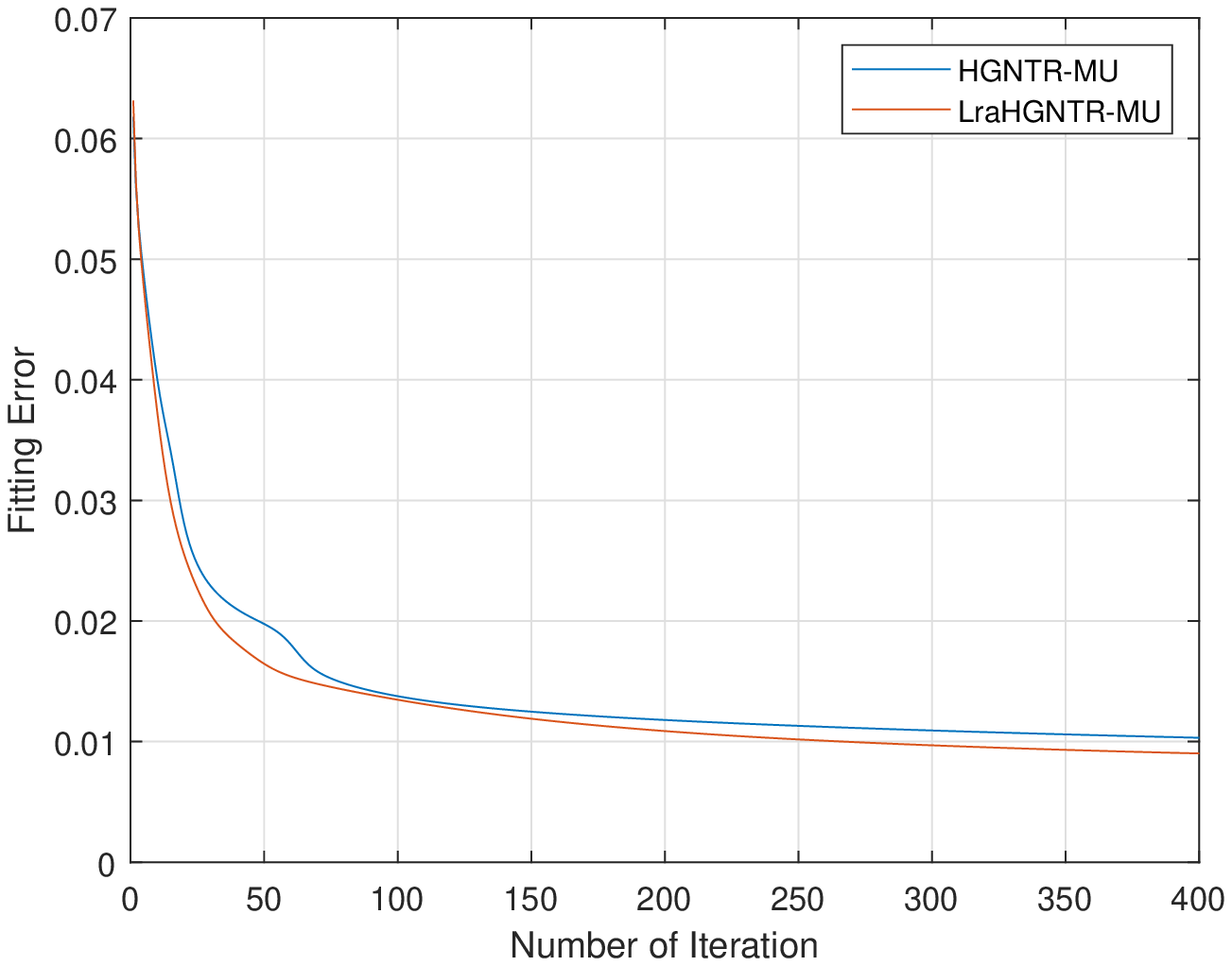}}
	\subfigure[ ]{
		\includegraphics[width=0.30\textwidth]{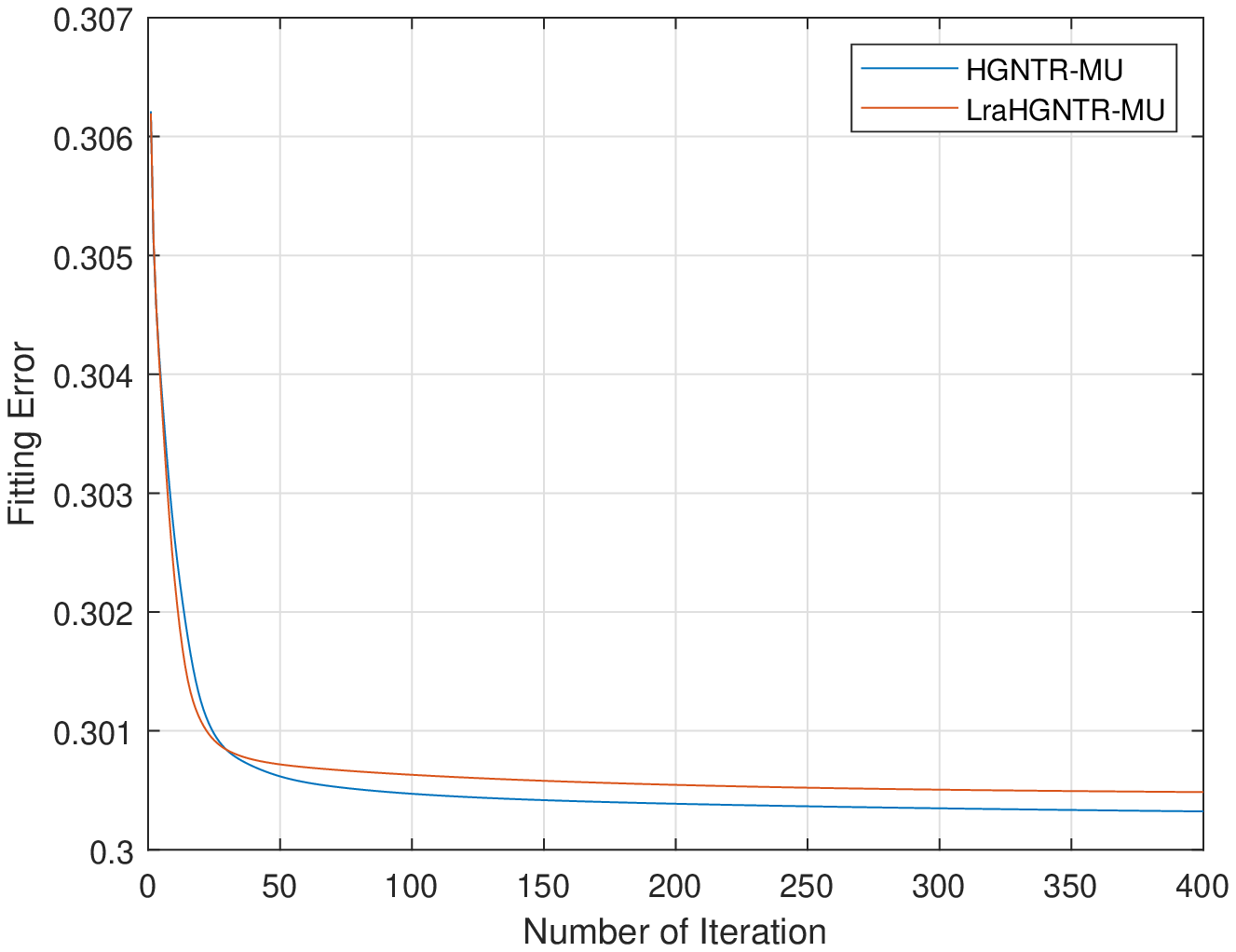}}
	\subfigure[ ]{
		\includegraphics[width=0.30\textwidth]{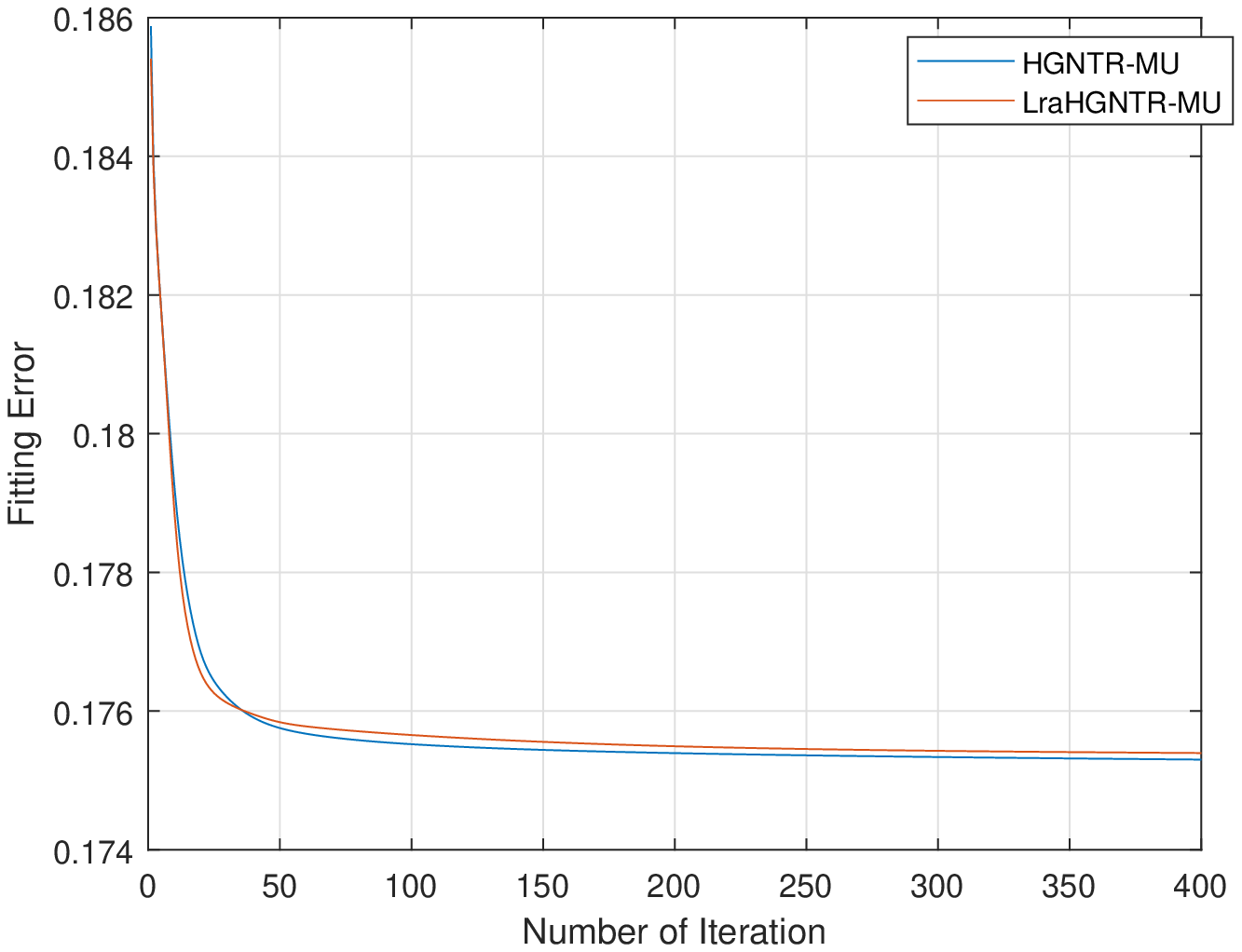}}

	\caption{The convergence curve on tensor $\mathcal{X}$ and noisy-added $\mathcal{X}$.The horizontal axis is the number of iterations. The vertical axis is the fitting error. (a) Original tensor. (b) Noisy-added tensor with SNR = $10\,dB$. (c) Noisy-added tensor with SNR = $15\,dB$. }
	\label{convergence curve}
\end{figure*}

We construct a tensor $\mathcal{X}\in\mathbb{R}^{40\times40\times40\times40}$ by random selection obeying uniform distribution between 0 and 1, and add independent Gaussian noisy with SNR = $10\,dB$ and SNR = $15\,dB$ on this tensor. Then we decomposition it by HGNTR and LraHGNTR. We count the convergence evolution of the error respectively in $Fig.\ref{convergence curve}$. 

\begin{figure}[!htb]  %?????????????????????
	\centering  %??
	\includegraphics[width=0.45\linewidth]{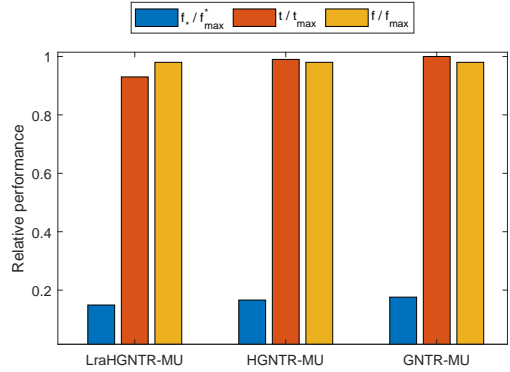} %photo???????????????????????????photo.png????????
	\caption{The comparsion of performance in terms of runtime and the final fitting accuracy on different algorithms. $f^{*}$ and $f^{*}_{max}$ denote the noiseless fitting error and max noiseless fitting error. $t$ and $t_{max}$ denote the each running time and max running time among them. $f$ and $f_{max}$ denote the noisy-added fitting error and max noise-added fitting error.
	}
	\label{convergence_relative} %?????????
\end{figure}

Our algorithms can reach a stable value after one hundred iterations. It can be proved that both algorithms are convergent. Particularly, we can see the LraHGNTR has a faster convergence rate in the first half part of the iteration process. It can verify that the convergence of the low-rank approximate method is faster. And our low-rank approximation method always reaches the stable convergence value firstly. Regretfully, in the condition of noisy-added tensor data, when the convergence is about to complete, our convergence accuracy of LraHGNTR is a little weaker than the HGNTR, there is a slight difference, this may is related to the heavy Gaussian noise added on the certain synthetic manual tensor data, it has completely random properties in the statistic. And $fig.\ref{convergence_relative}$ shows the relative performance in terms of running time and final fitting error. A similar conclusion has been shown in \cite{zhou2012fast}.
\subsection{Basis Visualization}
\label{sec:4.8}
\begin{figure*}[!htb]
	\subfigure[ ]{
		\includegraphics[width=0.132\textwidth]{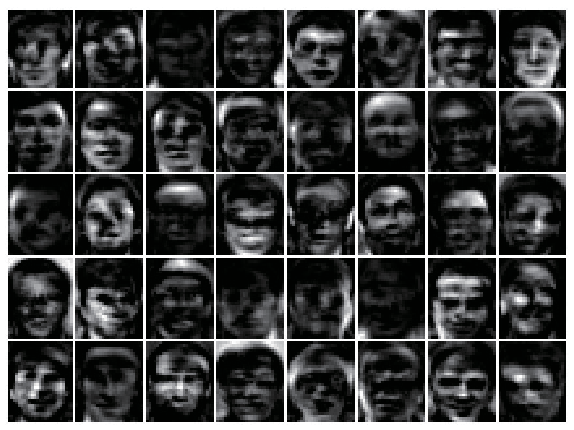}}
	\subfigure[ ]{
		\includegraphics[width=0.13\textwidth]{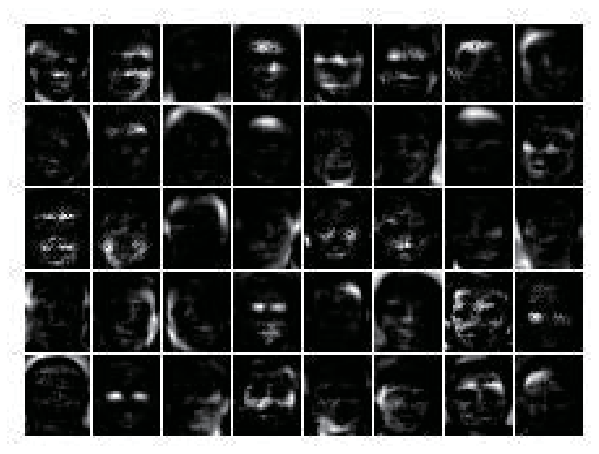}}
	\subfigure[ ]{
		\includegraphics[width=0.13\textwidth]{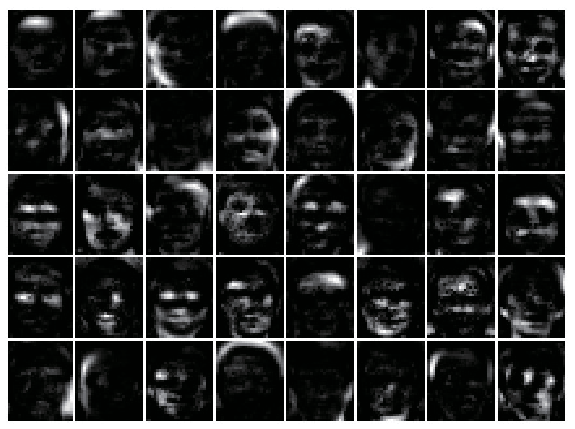}}
	\subfigure[ ]{
		\includegraphics[width=0.132\textwidth]{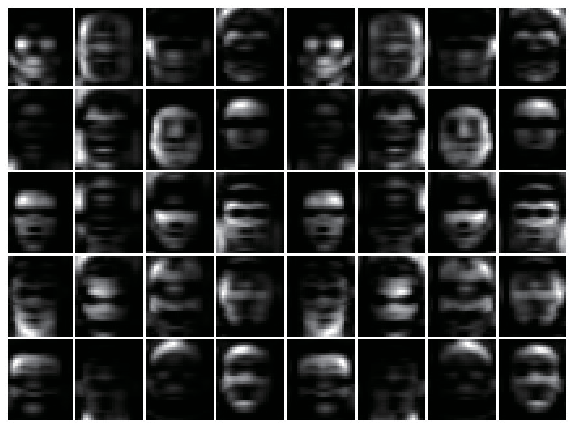}}
	\subfigure[ ]{
		\includegraphics[width=0.131\textwidth]{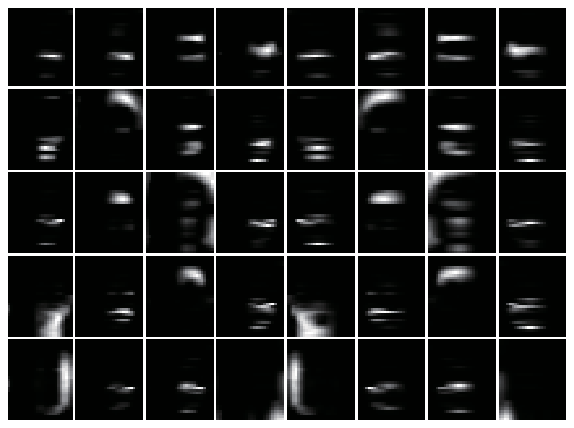}}
	\subfigure[ ]{
		\includegraphics[width=0.13\textwidth]{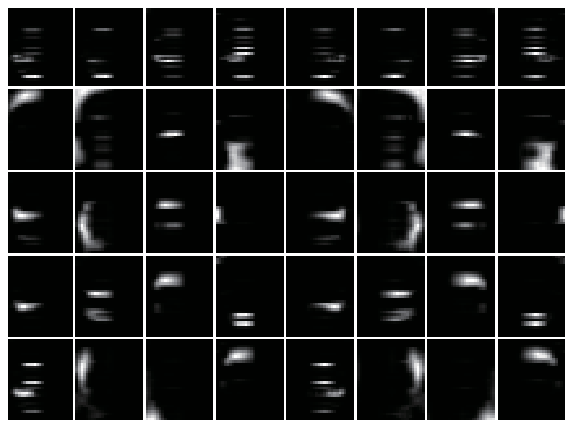}}
	\subfigure[ ]{
		\includegraphics[width=0.132\textwidth]{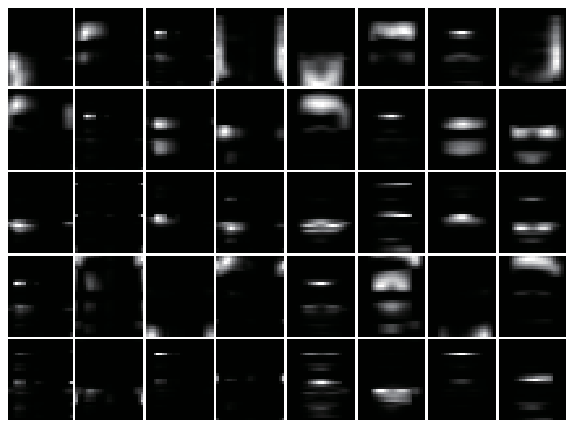}}
	
	\caption{Visual images of base-matrix based on different algorithms for ORL database. (a) NMF. (b) GNMF. (c) HNMF. (d) NTR-MU. (e) GNTR-MU. (f) HGNTR-MU. (g) LraHGNTR-MU.}.
	\label{fig:visual_orl}
\end{figure*}

\begin{figure*}[!htb]
	\subfigure[ ]{
		\includegraphics[width=0.13\textwidth]{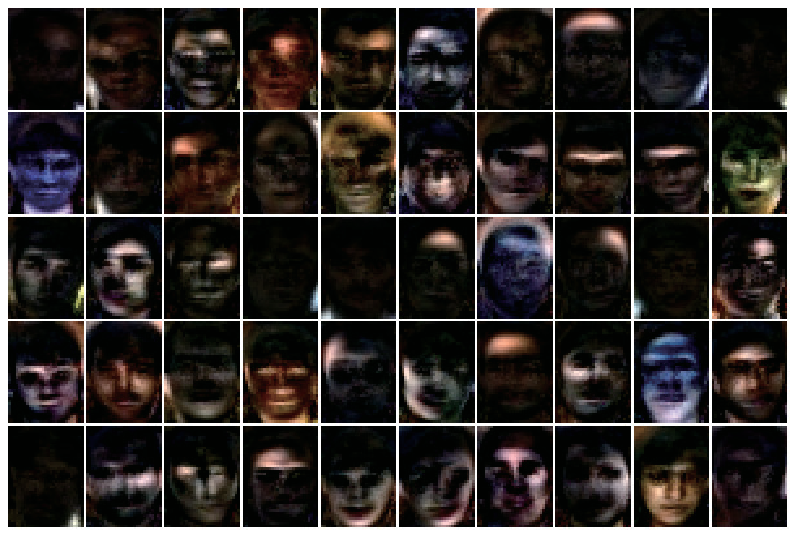}}
	\subfigure[ ]{
		\includegraphics[width=0.13\textwidth]{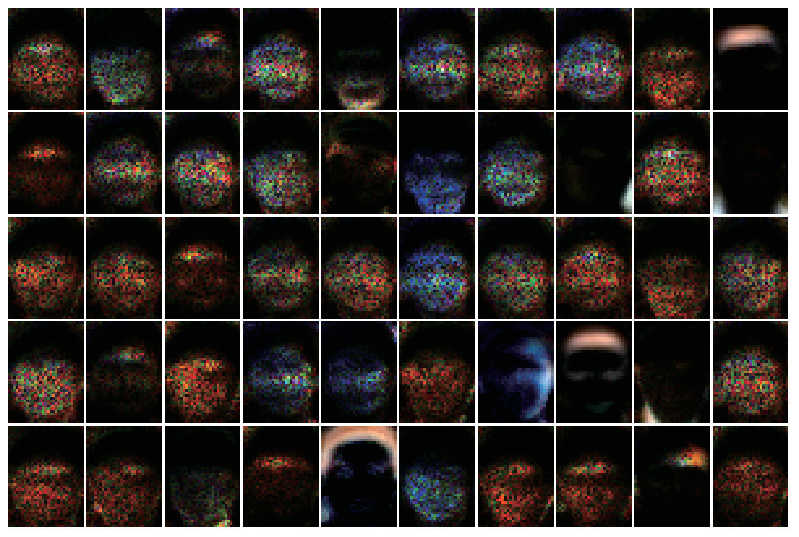}}
	\subfigure[ ]{
		\includegraphics[width=0.132\textwidth]{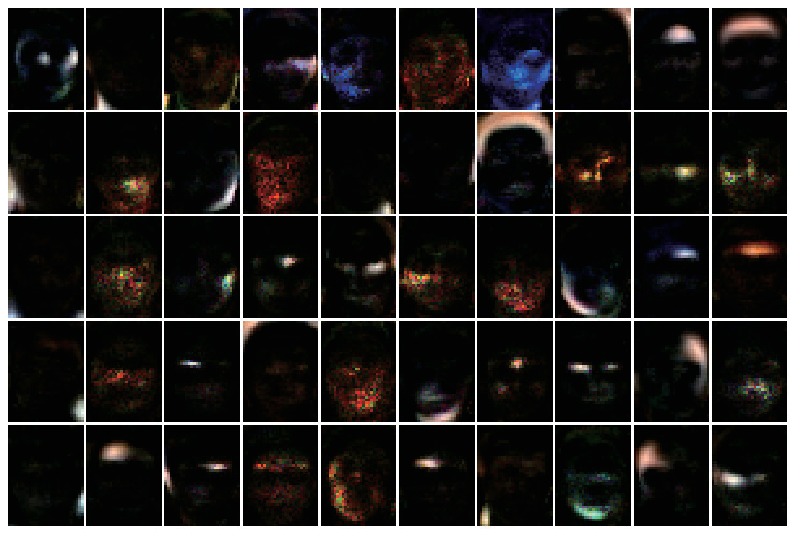}}
	\subfigure[ ]{
		\includegraphics[width=0.13\textwidth]{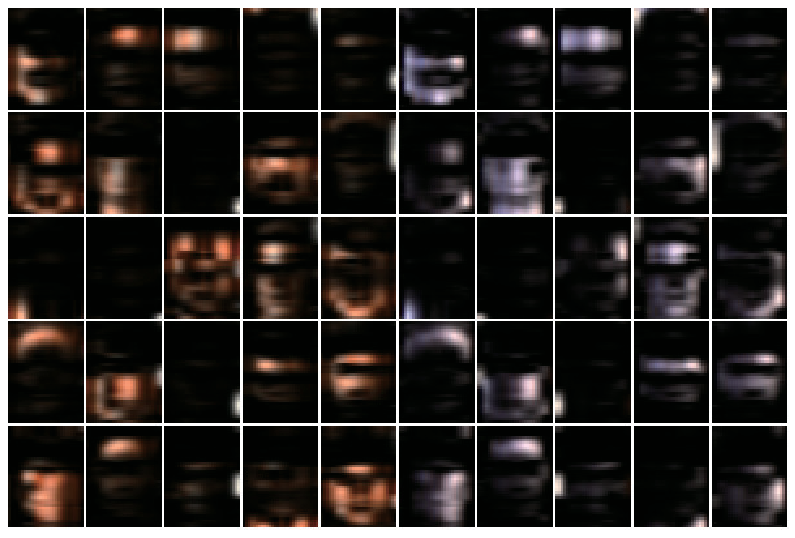}}
	\subfigure[ ]{
		\includegraphics[width=0.13\textwidth]{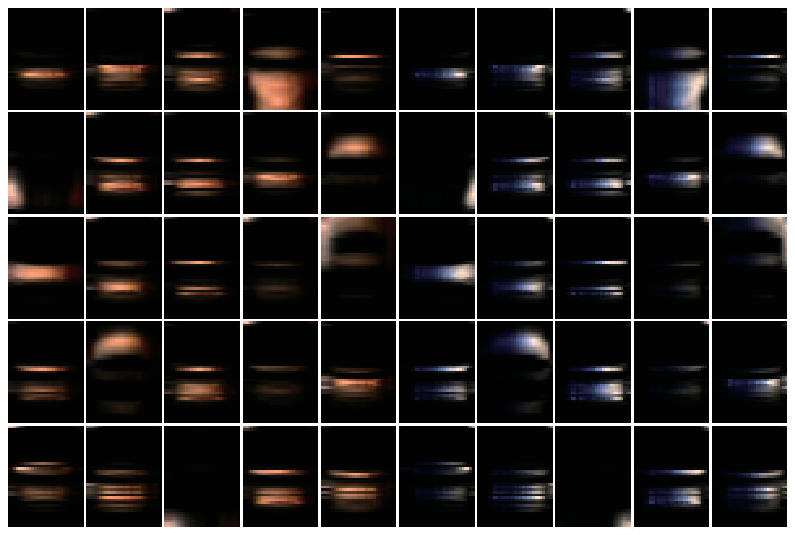}}
	\subfigure[ ]{
		\includegraphics[width=0.13\textwidth]{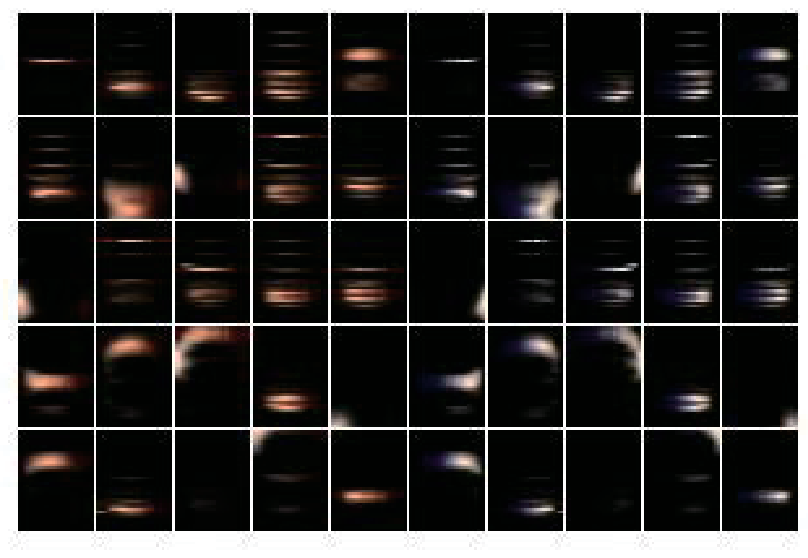}}
	\subfigure[ ]{
		\includegraphics[width=0.13\textwidth]{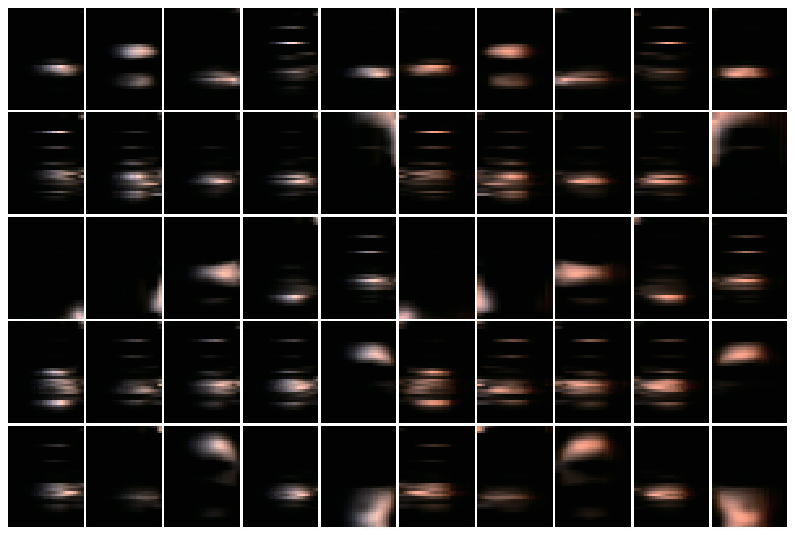}}
	
	\caption{Visual images of base-matrix based on different algorithms for GT database. (a) NMF. (b) GNMF. (c) HNMF. (d) NTR-MU. (e) GNTR-MU. (f) HGNTR-MU. (g) LraHGNTR-MU.}.
	\label{fig:visual_gt}    
\end{figure*}

In order to compare the ability that extracts the parts-based features from database objects by each algorithm, we visual the basis matrices that are extracted from the ORL and GT database in $Fig.\ref{fig:visual_orl}$ and $Fig.\ref{fig:visual_gt}$. For the  high-dimensional databases, such as ORL and GT, the matrix methods show very limited capabilities, so that they produce ghost faces, while the tensor methods can produce the more sparse representation. The nonnegativity of the tensor decomposition produces about two main advantages: 
(1) it can represent the certain potential structure of the data and be physically interpretable. (2) it can reduce the dimensionality of the data and memory storage costs. By comparing all methods, our algorithms HGNTR and LraHGNTR have a higher degree of sparsity. And at the same time, they retain more physically meaningful features.

\section{Conclusion}
\label{sec:5}
In this paper, we propose the method of HGNTR. Each dimension information of the third-order tensor is expressed as the corresponding three-order core tensor. At the same time, compared with ordinary manifold graph learning, hypergraph manifold learning can explore more high-order manifold structure information.
An efficient optimization method based on multiplicative updating rules (MUR) is developed. Considering the computational complexity of the algorithm, we use the low-rank approximation method to reduce the running time. The proposed HGNTR and LraHGNTR can achieve better performance than state-of-the-art (SOTA) algorithms.
\section{Acknowledgement}
\label{sec:6}
The authors would like to thank...

\bibliographystyle{cas-model2-names.bst}
\bibliographystyle{unsrt}
\bibliography{document.bib}

\end{document}